\documentclass{article}

\usepackage{microtype}
\usepackage{graphicx}
\usepackage{subfigure}
\usepackage{booktabs} 

\usepackage{paralist}
\usepackage{multicol}
\usepackage{multirow}
\usepackage{tabularx}
\usepackage{float}

\usepackage{xcolor}        
\usepackage{graphicx}
\usepackage{subfigure}
\usepackage{bigstrut}
\usepackage{wrapfig}
\usepackage[font=footnotesize]{caption}
\usepackage[thinc]{esdiff}
\usepackage[ruled]{algorithm2e}
\usepackage[firstpage]{draftwatermark}

\usepackage{dsfont}

\usepackage{amsmath}
\usepackage{amssymb}
\usepackage{mathtools}

\usepackage{colortbl}

\usepackage{hyperref}

\newcommand{\eg}{\textit{e.g., }}
\newcommand{\ie}{\textit{i.e., }}
\newcommand{\pufferfish}{\textsc{Pufferfish}}
\newcommand{\cuttlefish}{\textsc{Cuttlefish}}

\definecolor{Gray}{gray}{0.95}
\definecolor{DarkGray}{gray}{0.4}
\definecolor{LightCyan}{rgb}{0.88,1,1}

\usepackage[accepted]{mlsys2023}

\mlsystitlerunning{\cuttlefish{}: Low-rank Model Training without All The Tuning}

\begin{document}
\SetWatermarkText{
 \hspace*{4.8in}
 \raisebox{9.5in}{
  \includegraphics[height=0.9in]{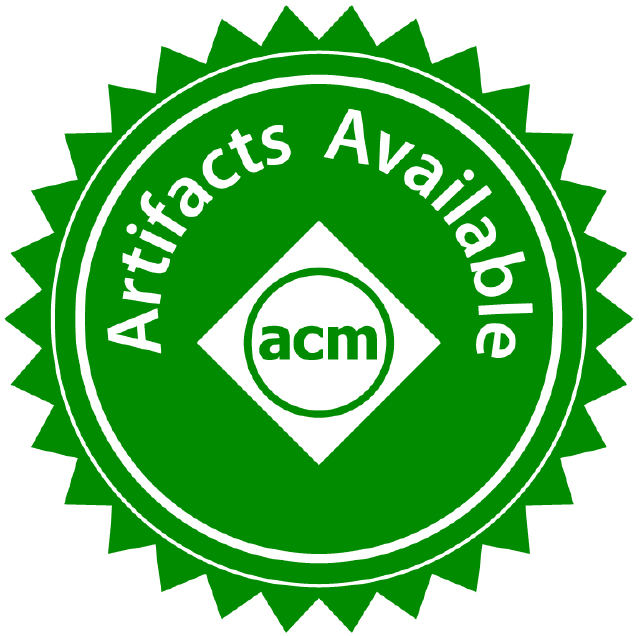}\hspace{-1.3cm}
  \includegraphics[height=0.9in]{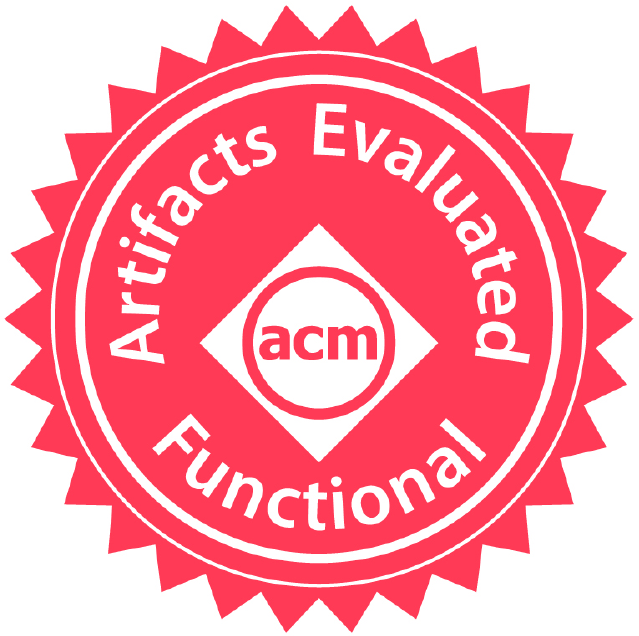}
 }
 \vspace{1.2cm}
}
\SetWatermarkAngle{0}

\twocolumn[
\mlsystitle{\cuttlefish{}: Low-rank Model Training without All The Tuning}




\begin{mlsysauthorlist}
\mlsysauthor{Hongyi Wang}{cmu}
\mlsysauthor{Saurabh Agarwal}{uwcs}
\mlsysauthor{Pongsakorn U-chupala}{sony}
\mlsysauthor{Yoshiki Tanaka}{sony}\\
\mlsysauthor{Eric P. Xing}{mbz,cmu,petuum}
\mlsysauthor{Dimitris Papailiopoulos}{uwece}
\end{mlsysauthorlist}

\mlsysaffiliation{cmu}{Machine Learning Department, Carnegie Mellon University}
\mlsysaffiliation{uwcs}{Department of Computer Sciences, University of Wisconsin-Madison}
\mlsysaffiliation{sony}{Sony Group Corporation}
\mlsysaffiliation{mbz}{Mohamed bin Zayed University of Artificial Intelligence}
\mlsysaffiliation{petuum}{Petuum, Inc.}
\mlsysaffiliation{uwece}{Department of Electrical and Computer Engineering, University of Wisconsin-Madison}

\mlsyscorrespondingauthor{Hongyi Wang}{hongyiwa@andrew.cmu.edu}

\mlsyskeywords{Machine Learning, MLSys}

\vskip 0.3in

\begin{abstract}
    Recent research has shown that training low-rank neural networks can effectively reduce the total number of trainable parameters without sacrificing predictive accuracy, resulting in end-to-end speedups. However, low-rank model training necessitates adjusting several additional factorization hyperparameters, such as the rank of the factorization at each layer. In this paper, we tackle this challenge by introducing \cuttlefish{}, an automated low-rank training approach that eliminates the need for tuning factorization hyperparameters. \cuttlefish{} leverages the observation that after a few epochs of full-rank training, the {\it stable rank} (\ie an approximation of the true rank) of each layer stabilizes at a constant value. \cuttlefish{} switches from full-rank to low-rank training once the stable ranks of all layers have converged, setting the dimension of each factorization to its corresponding stable rank. Our results show that \cuttlefish{} generates models up to 5.6$\times$ smaller than full-rank models, and attains up to a 1.2$\times$ faster end-to-end training process while preserving comparable accuracy. Moreover, \cuttlefish{} outperforms state-of-the-art low-rank model training methods and other prominent baselines. The source code for our implementation can be found at: \url{https://github.com/hwang595/Cuttlefish}.
\end{abstract}
]



\printAffiliationsAndNotice{}  

\section{Introduction}\label{sec:intro}
\vspace{-1.5mm}
As neural network-based models have experienced exponential growth in the number of parameters, ranging from 23 million in ResNet-50 (2015) to 175 billion in GPT-3 (2020) and OPT-175B (2022)~\cite{devlin2018bert,gpt-neurips-2020,fedus2021switch,zhang2022opt}, training these models has become increasingly challenging, even with the assistance of state-of-the-art accelerators like GPUs and TPUs. This problem is particularly pronounced in resource-limited settings, such as cross-device federated learning~\cite{kairouz2019advances,wang2020federated,wang2021field}. In response to this challenge, researchers have explored the reduction of trainable parameters during the early stages of training~\cite{frankle2018lottery,waleffe2020principal,khodak2020initialization,wang2021pufferfish} as a strategy for speeding up the training process.

Previous research efforts have focused on developing several approaches to reduce the number of trainable parameters during training. One method involves designing compact neural architectures, such as MobileNets~\cite{howard2017mobilenets} and EfficientNets~\cite{tan2019efficientnet}, which demand fewer FLOPs. However, this may potentially compromise model performance. Another alternative is weight pruning, which reduces the number of parameters in neural networks~\cite{han2015deep,han2015learning,frankle2018lottery,renda2020comparing,sreenivasan2022rare}. While unstructured sparsity pruning methods can result in low hardware resource utilization, recent advancements have proposed structured pruning based on low-rank weight matrices to tackle this issue~\cite{waleffe2020principal,khodak2020initialization,wang2021pufferfish,hu2021lora,chen2021drone,vodrahalli2022algorithms}. However, training low-rank models necessitates tuning additional hyperparameters for factorization, such as the width/rank of the factorization per layer, in order to achieve both compact model sizes, as measured by the number of parameters, and high accuracy.

\begin{figure}[ht]
\vspace{-2mm}
\centering
\includegraphics[width=0.35\textwidth]{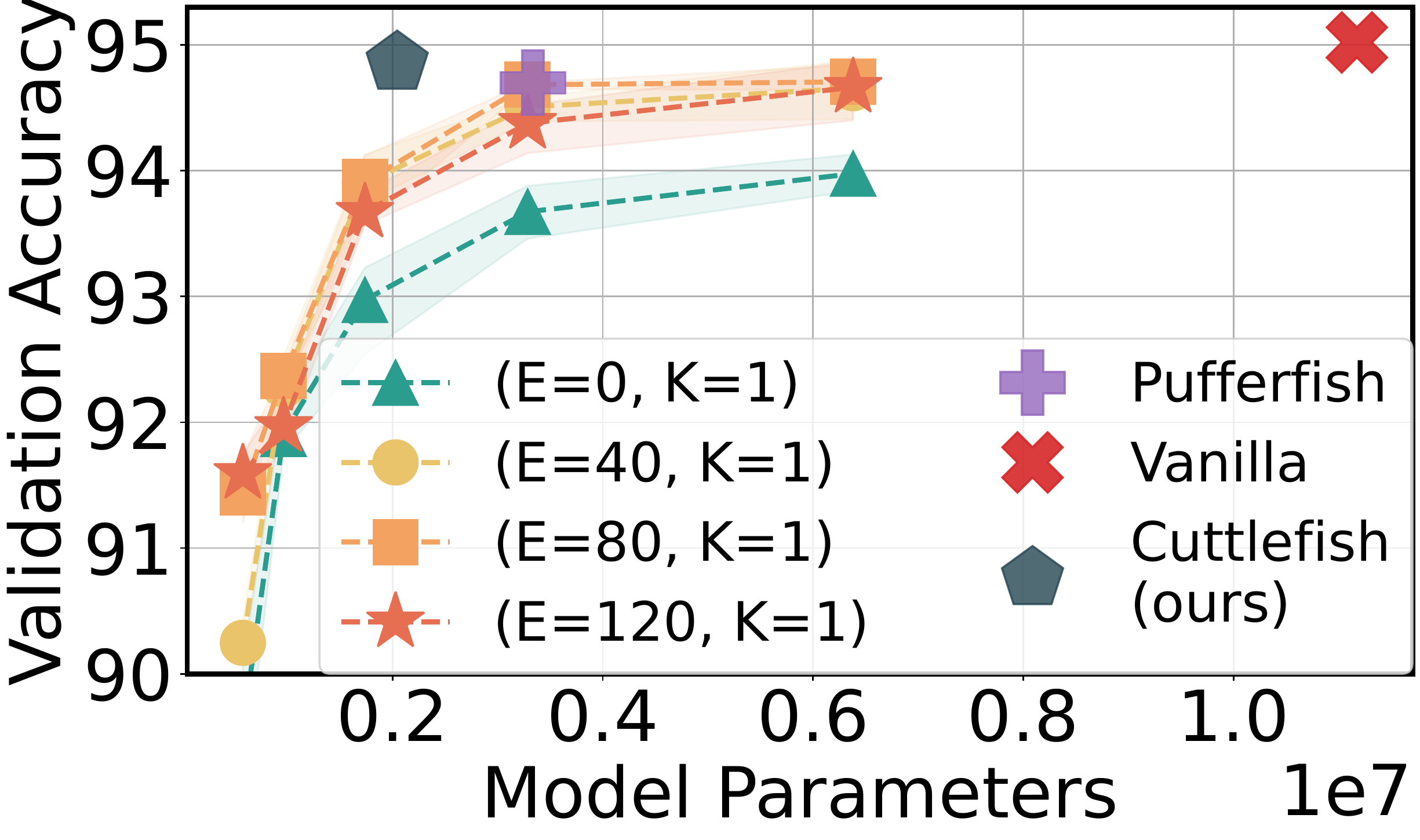}
\vspace{-1mm}
\includegraphics[width=0.35\textwidth]{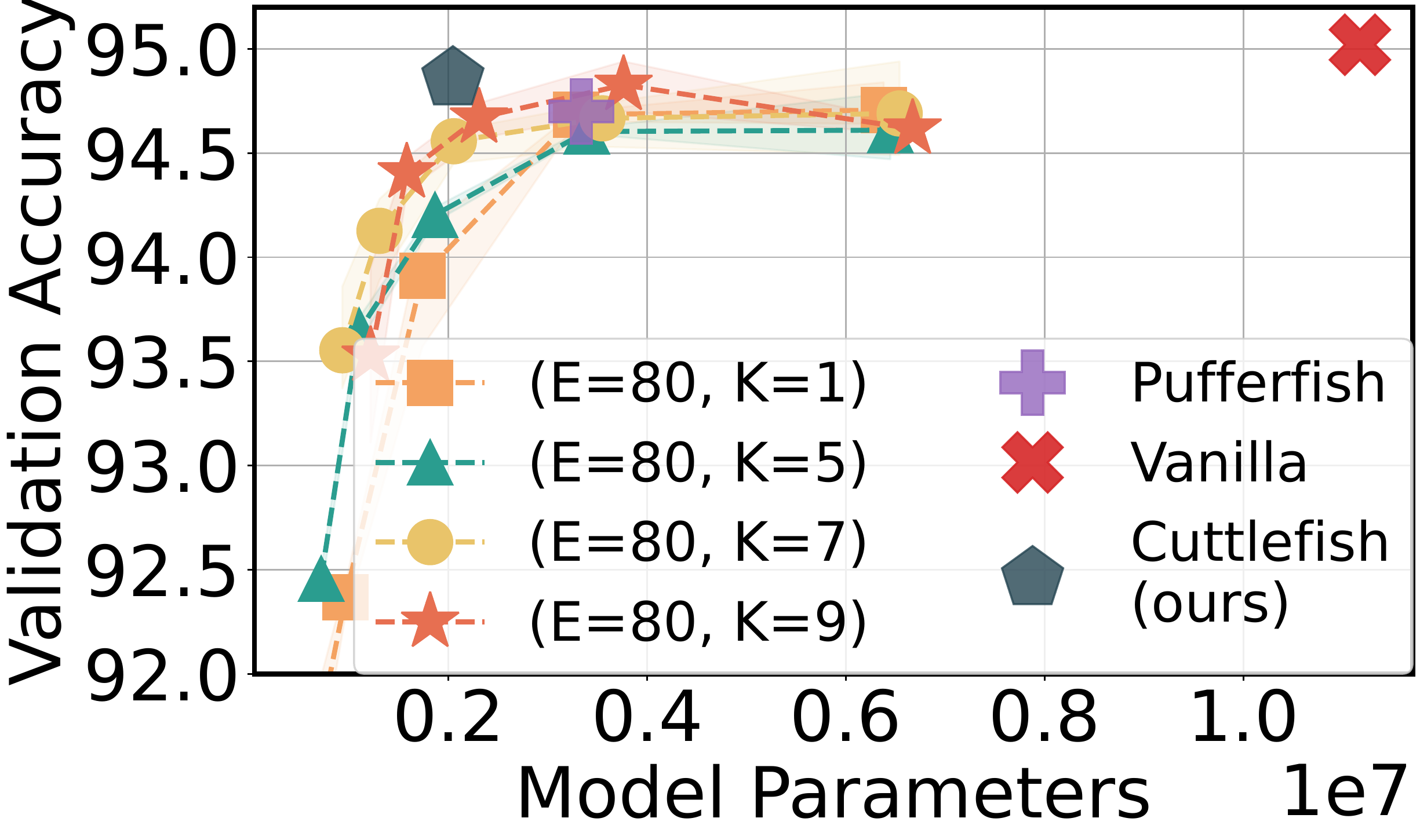}
\vspace{-3mm}
\caption{Comparison between \cuttlefish{} and grid search tuning results: ({\bf top}): fixing $K=1$ (the very first convolution layer is always not factorized) and varying $E \in \{0, 40, 80, 120\}$ and varying the selection of $\mathcal{R}$ by choosing various fixed rank ratios. ({\bf bottom}): fixing a good choice of $E$, \eg $E=80$ and varying $K$ and the rank ratio. The rank ratio varies among $\{ \frac{1}{32}, \frac{1}{16}, \frac{1}{8}, \frac{1}{4}, \frac{1}{2}\}$. Experiments ran on ResNet-18 trained over CIFAR-10.}
\label{fig:motivating-intro}
\vspace{-4mm}
\end{figure}

Striking the right balance between the size of a low-rank model and its accuracy is crucial, and depends on accurately tuning the rank of the factorized layers. As demonstrated in Figure~\ref{fig:motivating-intro}, improper tuning of factorization ranks can lead to either large models or diminished predictive accuracy. Training low-rank networks from scratch may cause significant accuracy loss~\cite{waleffe2020principal,wang2021pufferfish}. To address this, previous studies have suggested starting with full-rank model training for a specific number of epochs, $E$, before transitioning to low-rank model training~\cite{waleffe2020principal,wang2021pufferfish}. However, varying the number of full-rank training epochs can influence the final model accuracy, as illustrated in Figure~\ref{fig:motivating-intro}. Thus, selecting the appropriate number of full-rank training epochs, $E$, is essential (\eg neither $E=0$ nor $E=120$ yield the optimal model accuracy). Furthermore, to attain satisfactory accuracy, some earlier work~\cite{wang2021pufferfish} has proposed excluding the factorization from the first $K$ layers, resulting in a ``hybrid network" that balances model size and accuracy through the choice of $K$.

In this paper, we introduce a novel method for automatically determining the hyperparameters associated with low-rank training, ensuring that the resulting factorized model achieves both a compact size and high final accuracy.

\vspace{-4mm}
\paragraph{Challenges.} We would like to emphasize several reasons why this problem presents considerable challenges. Firstly, the search space $\mathcal{S}$ is vast. For a two hidden layer fully connected (FC) neural network with $100$ neurons in each layer (assuming the rank for each layer is $100$) and training with $100$ epochs, the cardinality of the search space is $|\mathcal{S}| = 100 \times 100 \times 100 \times 2 = 2 \times 10^6$. Furthermore, our objective of automatically optimizing low-rank training factorization hyperparameters while maintaining the advantages of end-to-end training speedups renders traditional neural architecture search (NAS) methods impractical. NAS necessitates concurrent training of both network architecture and network weights, resulting in computational requirements that substantially exceed those of standard model training.

\begin{figure}[ht]
    \vspace{-2mm}
    \centering
    \includegraphics[width=0.4\textwidth]{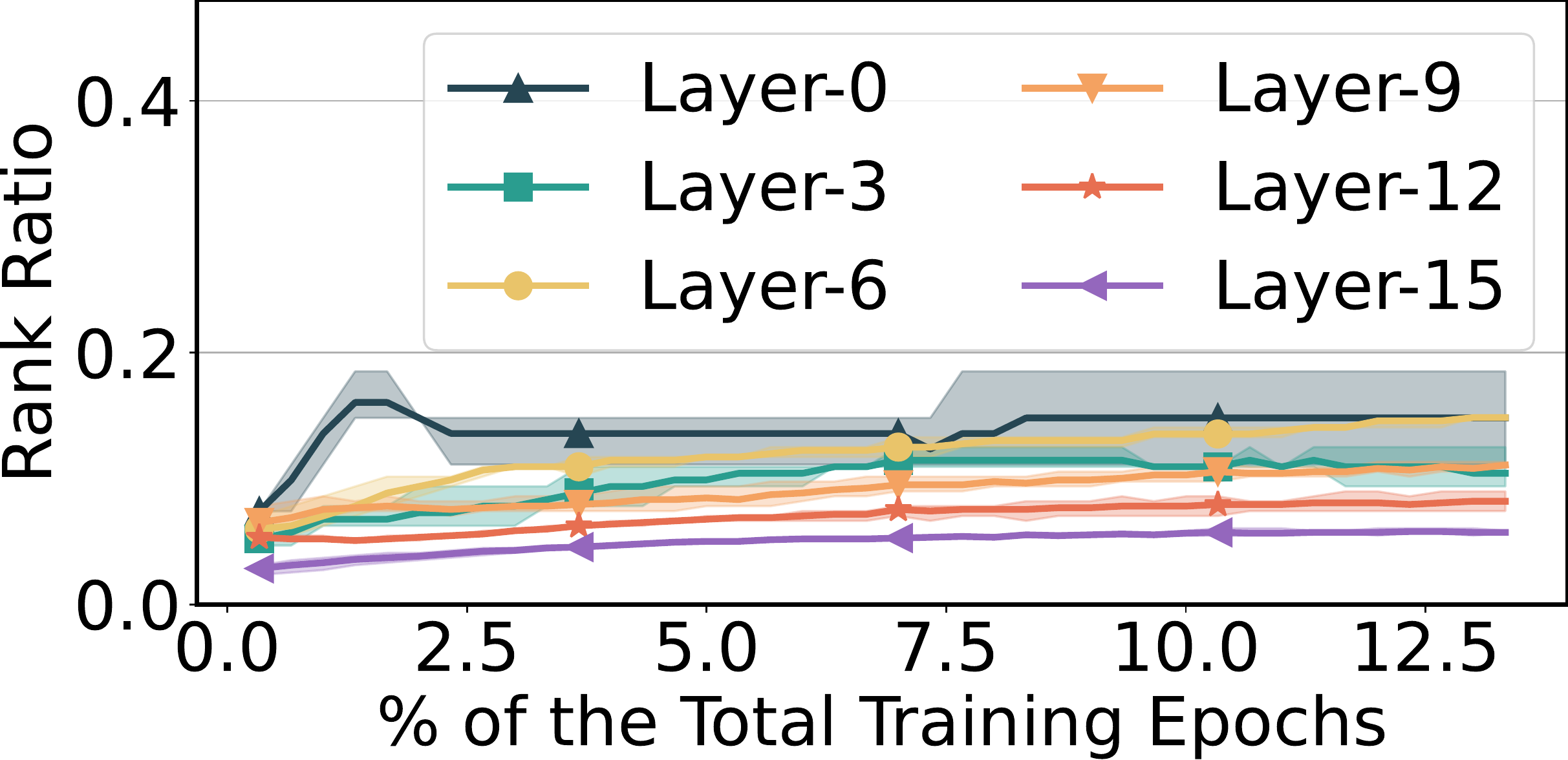}
    \vspace{-3mm}
    \caption{The estimated ranks for various layers in ResNet-18 trained on CIFAR-10, using {\it stable rank} (which will be discussed in detail later), can be found in our results. The results for other tasks are available in the appendix.}
    \label{fig:rank-est-resnet18-cifar10-intro}
    \vspace{-4mm}
\end{figure}

In this work, we present \cuttlefish{}, an automated low-rank factorized training method that eliminates the need for tuning factorization hyperparameters. We observe a key pattern in which the estimated rank of each layer changes rapidly during the initial stages of training and then stabilizes around a constant value (as depicted in Figure~\ref{fig:rank-est-resnet18-cifar10-intro}). We exploit this observation to develop a simple heuristic for selecting the layer ranks $\mathcal{R}$ and the full-rank training duration $E$: (i) transition from full-rank model training to low-rank model training when all the layer's stable ranks have converged to a constant, and (ii) use these constants as the rank of the factorization.

In addition to determining the factorization ranks and full-rank training duration, \cuttlefish{} also addresses the issue of deciding which layers to factorize. For convolutional neural networks (CNNs), \cuttlefish{} observes that factorizing early layers does not lead to considerable speedups, as elaborated in Section~\ref{sec:determine-k}. This observation, along with insights from prior research~\cite{wang2021pufferfish}, implies that factorizing early layers may negatively affect the final model accuracy without offering significant performance gains. To tackle this challenge, \cuttlefish{} performs lightweight profiling to identify the layers to factorize, ensuring that factorization occurs only in layers that can effectively enhance the training speed.

\vspace{-4mm}
\paragraph{Our contributions.} We observe a stabilizing effect in the stable ranks of neural network (NN) layers during training, where their stable ranks initially change rapidly and then converge to a constant. Based on this observation, we devise a heuristic to adaptively select the rank of each layer and the duration of full-rank warm-up training. We implement our technique, called \cuttlefish{}, and assess it in large-scale training environments for both language and computer vision tasks. Our comprehensive experimental results demonstrate that \cuttlefish{} automatically selects all factorization hyperparameters during training on-the-fly, eliminating the need for multiple experimental trials for factorization hyperparameter tuning. \cuttlefish{} strikes a balance between model size and final predictive accuracy, excelling in at least one dimension of producing smaller, more accurate models and achieving considerable training speedups compared to state-of-the-art low-rank training, structured pruning, sparse training, quantized training, and learnable factorization methods~\cite{rastegari2016xnor,frankle2019stabilizing,wang2020picking,yu2019universally,you2019drawing,idelbayev2020low,khodak2020initialization,wang2021pufferfish}.

\vspace{-2.5mm}
\subsection{Related work.}
\vspace{-1mm}
Several methods have been developed in the literature to eliminate redundancy in the parameters of modern NNs. Model compression strives to eliminate redundancy in the parameters of trained NNs~\cite{han2015deep}. Over time, numerous methods have been devised to remove redundant weights in NNs, encompassing pruning~\cite{li2016pruning,wen2016learning,hu2016network,zhu2017prune,he2017channel,yang2017designing,liu2018rethinking,yu2018nisp,yu2018slimmable,sreenivasanrare2020raregems}, quantization~\cite{rastegari2016xnor,zhu2016trained,hubara2016binarized,wu2016quantized,hubara2017quantized,zhou2017incremental}, low-rank factorization~\cite{xue2013restructuring,sainath2013low,jaderberg2014speeding,wiesler2014mean,konevcny2016federated}, and knowledge distillation~\cite{hinton2015distilling,yu2018slimmable,sanh2019distilbert,jiao2020tinybert,li2023mpcformer}.

\vspace{1.1mm}
The {\it Lottery Ticket Hypothesis} (LTH) suggests that smaller, randomly initialized subnetworks can be trained to attain accuracy levels comparable to those of the full network, although pinpointing these subnetworks can be computationally challenging~\cite{frankle2018lottery}. {\it Iterative Magnitude Pruning} (IMP) was devised to stabilize LTH while reducing computational costs through warm-up steps~\cite{frankle2019stabilizing}. Other efforts have sought to eliminate the need for model weight rewinding~\cite{renda2020comparing} and to identify winning tickets at initialization~\cite{wang2020picking,sreenivasan2022rare}. Moreover, researchers have explored sparsifying NNs during training~\cite{evci2020rigging}. However, these sparsification methods focus on unstructured sparsity, which does not yield actual speedups on current hardware. In contrast, low-rank training can lead to tangible acceleration. 

\vspace{1.1mm}
Low-rank factorized training, as well as other structured pruning methods, aim to achieve NNs with structured sparsity during training, allowing for tangible speedups to be obtained~\cite{waleffe2020principal,khodak2020initialization,you2020drawing,wang2021pufferfish,chen2021drone}. Low-rank factorized training has also been employed in federated learning methods to improve communication efficiency and hardware heterogeneity awareness~\cite{hyeon-woo2022fedpara,yao2021fedhm}. Low-rank factorization techniques have been shown to be combinable and used for training low-rank networks from scratch~\cite{ioannou2015training}. However, this method results in a noticeable loss of accuracy, as demonstrated in~\cite{wang2021pufferfish}. To tackle this problem, \cite{khodak2020initialization} introduces spectral initialization and Frobenius decay, while~\cite{vodrahalli2022algorithms} proposes the {\it Nonlinear Kernel Projection} method as an alternative to SVD. Low-rank training has also been investigated for fine-tuning large-scale pre-trained models~\cite{hu2021lora}. These techniques all necessitate additional hyperparameters, which can be tedious to fine-tune. The {\it LC compression} method attempts to resolve this issue by explicitly learning $\mathcal{R}$ during model training through alternating optimization~\cite{idelbayev2020low}. However, this approach is computationally demanding. Our proposed \cuttlefish{} method automatically determines all factorization hyperparameters during training on-the-fly, eliminating the heavy computation overhead and the need for multiple experimental trials for factorization hyperparameter tuning.

Alternative transformations have also been investigated, including Butterfly matrices~\cite{chen2022pixelated}, the fusion of low-rank factorization and sparsification~\cite{chen2021scatterbrain}, and block-diagonal matrices~\cite{dao2022monarch}. Furthermore, novel architectures have been developed for enhanced training or inference efficiency, such as SqueezeNet~\cite{iandola2016squeezenet}, Eyeriss~\cite{chen2016eyeriss}, ShuffleNet~\cite{zhang2018shufflenet}, EfficientNet~\cite{tan2019efficientnet}, MobileNets~\cite{howard2017mobilenets}, Xception~\cite{chollet2017xception}, ALBERT~\cite{lan2019albert}, and Reformer~\cite{kitaev2020reformer}.
\vspace{-3mm}
\section{Preliminary}
\vspace{-1.5mm}
In this section, we present an overview of the core concepts of low-rank factorization for various NN layers, along with a selection of specialized training methods specifically designed for low-rank factorized training.
\vspace{-3mm}
\subsection{Low-rank factorization of NN layers}
\vspace{-1.5mm}
\paragraph{FC/MLP Mixer layer.} A 2-layer fully connected (FC) neural network can be represented as $f(x) = \sigma(\sigma(x\mathbf{W}_1)\mathbf{W}_2)$, where $\mathbf{W}$s are weight matrices, $\sigma(\cdot)$ is an arbitrary activation function, and $x$ is the input data point. The weight matrix $\mathbf{W}$ can be factorized as $\mathbf{U} \mathbf{V}^\top$. A similar approach can be applied to ResMLP/MLP mixer layers, where each learnable weight can be factorized in the same manner~\cite{touvron2021resmlp,tolstikhin2021mlp}. 

\vspace{-4mm}
\paragraph{Convolution layer.} For a convolutional layer with dimensions $(m, n, k, k)$, where $m$ and $n$ are the number of input and output channels and $k$ represents the size of a convolution filter, a common approach involves factorizing the unrolled 2D matrix. We will discuss a popular method for factorizing a convolutional layer. Initially, the 4D tensor $\mathbf{W}$ is unrolled to obtain a 2D matrix of shape $(mk^2, n)$, where each column represents the weight of a vectorized convolution filter. The rank of the unrolled matrix is determined by $\min\{
mk^2,n\}$. Factorizing the unrolled matrix results in $\mathbf{U} \in \mathbb{R}^{mk^2\times r}$ and $\mathbf{V}^\top \in \mathbb{R}^{r \times n}$. Reshaping the factorized $\mathbf{U}, \mathbf{V}^\top$ matrices back to 4D yields $\mathbf{U} \in \mathbb{R}^{m \times r \times k \times k}$ and $\mathbf{V}^\top \in \mathbb{R}^{r \times n}$. Consequently, factorizing a convolutional layer produces a thinner convolutional layer $\mathbf{U}$ with $r$ convolution filters and a linear projection layer $\mathbf{V}^\top$. The $\mathbf{V}^\top$s can also be represented by a $1\times 1$ convolutional layer, such as $\mathbf{V}^\top \in \mathbb{R}^{r \times n \times 1 \times 1}$, which is more suited for computer vision tasks since it operates directly in the spatial domain~\cite{lin2013network,wang2021pufferfish}. 

\vspace{-3mm}
\paragraph{Multi-head attention (MHA) layer.}\label{sec:low-rank-transformer}
A $p$-head attention layer learns $p$ attention mechanisms on the key, value, and query ($\mathbf{K}, \mathbf{V}, \mathbf{Q}$) of each input token:
\vspace{-1.5mm}
{\small
\begin{align*}
\texttt{MHA}(\mathbf{Q}, \mathbf{K}, \mathbf{V}) = \texttt{Concat}(\text{head}_1, \dots, \text{head}_p) \mathbf{W}^O.
\end{align*}\\[-0.45cm]
}
Each head performs the computation of: 
\vspace{-1.5mm}
{\small
\begin{align*}
\texttt{head}_i &= \texttt{Attention}(\mathbf{Q} \mathbf{W}^{(i)}_Q, \mathbf{K}\mathbf{W}^{(i)}_K, \mathbf{V}\mathbf{W}^{(i)}_V)\\
&= \texttt{softmax}\left(\frac{\mathbf{Q} \mathbf{W}^{(i)}_Q \mathbf{W}^{(i)\top}_K \mathbf{K}^\top}{\sqrt{d/p}}\right) \mathbf{V}\mathbf{W}^{(i)}_V.
\end{align*}\\[-0.45cm]
}
where $d$ is the hidden dimension. The trainable weights $\mathbf{W}^{(i)}_Q, \mathbf{W}^{(i)}_K, \mathbf{W}^{(i)}_V, i \in \{1, 2, \dots, p\}$ can be factorized by simply decomposing all learnable weights $\mathbf{W}{\cdot}$ in an attention layer and obtaining $\mathbf{U}{\cdot} \mathbf{V}^\top{\cdot}$~\cite{vaswani2017attention}.
\vspace{-3mm}
\subsection{Training methods for low-rank networks}
\vspace{-1.5mm}
\paragraph{Hybrid NN architecture.} It has been noted that factorizing the initial layers may negatively impact a model's accuracy~\cite{konevcny2016federated,waleffe2020principal,wang2021pufferfish}. One possible explanation is that NN layers can be viewed as feature extractors, and poor features extracted by the early layers can accumulate and propagate throughout the NN. To address this issue, the \textit{hybrid NN architecture} was proposed, which only factorizes the lower layers while keeping the initial layers full-rank~\cite{wang2021pufferfish}. The weights of a full-rank $L$-layer NN can be represented as $\mathcal{W}=\{\mathbf{W}_i| 1 \leq i \leq L\}$. The corresponding hybrid model's weights can be represented as $\mathcal{H}=\{\mathbf{W}_1, \mathbf{W}_2, \dots, \mathbf{W}_K, \mathbf{U}_{K+1},\mathbf{V}^\top_{K+1}, \dots,\mathbf{U}_{L-1},$ $\mathbf{V}^\top_{L-1}, \mathbf{W}_L \}$, where $K$ is the number of layers that are not factorized and is treated as a hyperparameter to be tuned manually~\cite{wang2021pufferfish}. It is important to note that the last classification layer, \ie $\mathbf{W}_L$, is usually not factorized~\cite{khodak2020initialization,wang2021pufferfish}. 

\vspace{-4mm}
\paragraph{Full-rank to low-rank training.} 
Training low-rank factorized models from scratch often results in a decrease in accuracy~\cite{waleffe2020principal,khodak2020initialization,wang2021pufferfish}. To mitigate this drop, it is common to train the full-rank model for $E$ epochs before factorizing it~\cite{waleffe2020principal,khodak2020initialization,wang2021pufferfish}. However, determining the appropriate number of full-rank training epochs is treated as a hyperparameter and typically tuned manually in experiments~\cite{wang2021pufferfish}. Observations indicate that finding the right number of full-rank training epochs is crucial for achieving optimal final model accuracy in low-rank factorized NNs. 

\vspace{-4mm}
\paragraph{Initialization and weight decay.} Factorized low-rank networks can benefit from tailored initialization methods~\cite{ioannou2015training,khodak2020initialization}. One such method, called \textit{spectral initialization}, aims to approximate the behavior of existing initialization methods~\cite{khodak2020initialization}. Spectral initialization represents a special case of transitioning from full-rank to low-rank training with $E=0$. Additionally, specific regularization techniques have been proposed to enhance the accuracy of low-rank networks. For example, \textit{Frobenius decay} applies weight decay on $\|\mathbf{U}\mathbf{V}^\top\|^2_F$ instead of $\|\mathbf{U}\|^2_F+\|\mathbf{V}^\top\|^2_F$ during factorized low-rank training. 
\vspace{-3mm}
\section{\cuttlefish{}: Automated low-rank factorized training}
\vspace{-2mm}
In this section, we outline the problem formulation of \cuttlefish{}, elaborate on each factorization hyperparameter, and describe \cuttlefish{}'s heuristics for determining all of these factorization hyperparameters.
\vspace{-3mm}
\subsection{Problem formulation.} 
\vspace{-1.5mm}
The search space for adaptive factorized tuning is defined by three sets of hyperparameters, namely $\mathcal{S} = (E, K, \mathcal{R})$ (full-rank training epochs, the number of initial layers that remain unfactorized, and layer factorization ranks). The objective of \cuttlefish{} is to find an optimal $\hat s \in \mathcal{S}$ on-the-fly, with minimal computational overhead during training, such that the resulting low-rank factorized models are both compact and maintain high accuracy, comparable to their full-rank counterparts.

\vspace{-3mm}
\subsection{Components in the search space and the trade-offs among hyperparameter selections.}
\vspace{-1.5mm}
\paragraph{Full-rank training epochs $E$.} The value of $E$ can range from $0$ to $T-1$. Neither too small (\eg $E=0$) nor too large (\eg $E=120$) values of $E$ result in the best accuracy (Figure~\ref{fig:motivating-intro}), highlighting the necessity of tuning $E$.

\vspace{-4mm}
\paragraph{The number of full-rank layers $K$.} As previously mentioned, the weights of a hybrid NN architecture can be represented by $\mathcal{H}=\{\mathbf{W}_1, \dots, \mathbf{W}_K, \mathbf{U}_{K+1},\mathbf{V}^\top_{K+1}, \dots,\mathbf{U}_{L-1}, \mathbf{V}^\top_{L-1}, \mathbf{W}_L \}$, where $K$ is a hyperparameter to be tuned. The selection of $K$ can range from 1 to $L-1$, meaning the very first and very last layers are always not factorized. Factorizing additional layers results in increased accuracy loss but also reduces the model size and computational complexity. Thus, an optimal choice for $K$ should balance the trade-off between accuracy loss and model compression rate.

\vspace{-4mm}
\paragraph{Rank selections for factorized layers $\mathcal{R}$.} $\mathcal{R}$ specifies the ranks used when factorizing the $(L-K-1)$ layers in a hybrid NN architecture, \ie $\mathcal{R}=\{r_i|K+1 \leq i \leq L-1\}$. For a layer weight $\mathbf{W}_i \in \mathbb{R}^{m\times n}$, the full rank of the layer is $\text{rank}(\mathbf{W}_i) = \min\{m, n\}$. Thus, $1\leq r_i \leq \text{rank}(\mathbf{W}_i), \forall i \in \{K+1,\dots,L-1\}$. Using a too small $r$ for factorizing a layer may result in a decrease in accuracy. However, employing a relatively large $r$ to factorize the layer could negatively impact the model compression rate.

In this paper, we develop heuristics that automatically identify optimal choices for each component within the search space, \ie $\hat s \in \mathcal{S}$, in order to strike a balance between the final accuracy and model size.

\vspace{-4mm}
\paragraph{Why is finding an appropriate $s \in \mathcal{S}$ challenging?}
Firstly, the search space's cardinality, $|\mathcal{S}|$, is vast. Furthermore, the primary goal of low-rank factorized training is to accelerate model training. Therefore, it is crucial to identify $\hat s$ without introducing significant computational overhead. While Neural Architecture Search (NAS) style methods could potentially be employed to search for $s \in \mathcal{S}$, they result in high computational overhead. Consequently, adopting NAS-based algorithms is not suitable for our scenario, as our objective is to achieve faster training.

\vspace{-3mm}
\subsection{Determining factorization ranks ($\mathcal{R}$) for NN layers.}
\vspace{-1.5mm}
In previous work, ranks of the factorized layers have typically been treated as a hyperparameter, with a fixed global rank ratio $\rho$ often employed, for example, $\mathcal{R}=\{\rho \cdot \text{rank}(\mathbf{W}_i)|1\leq i \leq L\}$~\cite{khodak2020initialization,wang2021pufferfish}. However, a crucial question to consider is \textit{do all layers converge to the same $\rho$ during training?}

\vspace{-4mm}
\paragraph{Rank estimation metric.} One might wonder why we cannot simply use the normal rank for estimating the layer ranks of NNs. The answer is that the normal rank is always full for layer weights of NNs. However, NN weight matrices are ``nearly" low-rank when they exhibit a rapid spectral decay. Therefore, we require a metric to estimate layer ranks. In \cuttlefish{}, we utilize the {\it stable rank}, which serves as a valuable proxy for the actual rank since it remains largely unaffected by small singular values, to estimate the rank of model layer weights $\mathbf{W}$. The definition of stable rank is $\texttt{stable rank}(\mathbf{\Sigma})=\frac{\mathbf{1}^\top\mathbf{\Sigma}^2 \mathbf{1}}{\sigma^2_{\text{max}}(\mathbf{W})}$, where $\mathbf{1}$, $\sigma_{\text{max}}^2(\cdot)$, and $\Sigma$ represent the identity column vector, the maximum squared singular value, and the diagonal matrix that stores all singular values in descending order, \ie $\mathbf{1}^\top \mathbf{\Sigma} =\begin{bmatrix}\sigma_1, \dots, \sigma_{\text{rank}(\mathbf{W})}\end{bmatrix}$, respectively. Another advantage of using stable rank is that its calculation does not require specifying any additional hyperparameters. 

\vspace{-4mm}
\paragraph{The scaled stable rank.} Stable rank, which disregards minuscule singular values, often results in very low rank estimations. This can work well for relatively small tasks, such as CIFAR-10. However, for larger scale tasks like ImageNet, using stable rank directly leads to a non-trivial accuracy drop of 2.3\% (details can be found in the appendix). To address this issue, we propose using {\it scaled stable rank}. Scaled stable rank assumes that the estimated rank of a randomly initialized matrix, \ie $\mathbf{W}^0$ (model weight at the 0-th epoch), should be close or equal to full rank. Nevertheless, based on our experimental observations, stable rank estimation of randomly initialized weights tends not to be full rank. Therefore, we store the ratio of full rank to initial stable rank (denoted as $\xi$, \eg if $\text{rank}(\mathbf{W})=512$ and $\texttt{stable rank}(\mathbf{\Sigma}^0)=200$, then $\xi=512/200$). We scale each epoch's stable rank by: 
\vspace{-2mm}
{\small
\begin{align*}
     &\texttt{scaled stable rank}(\mathbf{\Sigma}, \xi)= \xi \cdot \texttt{stable rank}(\mathbf{\Sigma});\\ &\xi=\frac{\text{rank}(\mathbf{W}^{0})}{\texttt{stable rank}(\mathbf{\Sigma}^0)}, \forall t\in \{1,2,\dots, T\}. \\[-0.65cm]
\end{align*}
}

\vspace{-4mm}
\paragraph{\cuttlefish{} rank selection.} 
We observe that different layers tend to converge to varying stable ranks (an example is shown in Figure~\ref{fig:rr-epochs-resnet18-cifar10}, with similar trends found in other tasks, as detailed in the appendix). Middle layers generally converge to larger $\rho$s, indicating greater redundancy. As a result, it is unlikely that a fixed rank ratio is optimal, as it may either fail to eliminate all redundancy in the layer weights or be too aggressive in compressing model weights, thereby compromising final accuracy. \cuttlefish{} employs the scaled stable rank at epoch $E$ (\ie the transition point from full-rank to low-rank) to factorize the full-rank model and obtain a low-rank factorized model.
\begin{figure}[ht]
    \centering
    \vspace{-2mm}
    \includegraphics[width=0.48\textwidth]{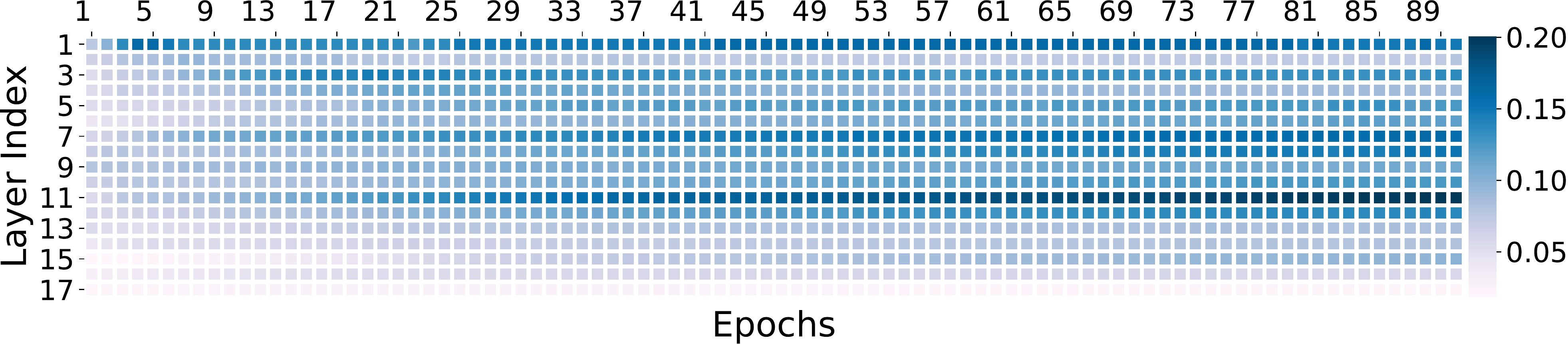}
    \vspace{-8mm}
    \caption{The rank ratios ($\rho$s) of stable ranks for ResNet-18 trained on CIFAR-10, where darker colors indicate higher $\rho$ values (results for other datasets can be found in the appendix).}
    \label{fig:rr-epochs-resnet18-cifar10}
    \vspace{-5mm}
\end{figure}

\vspace{-3mm}
\subsection{Determining full-rank training epochs $E$} 
\vspace{-1.5mm}
As discussed in Section~\ref{sec:intro}, neither too small nor too large $E$ values result in optimal accuracy. Furthermore, larger $E$ values also lead to slower training time, as full-rank models have higher computational complexity. \cuttlefish{} is inspired by the observation that the estimated ranks for all NN layers, $\mathcal{R}$, change rapidly during the early training phase but stabilize in later training epochs. A reasonable heuristic, therefore, is to switch from full-rank training to low-rank training when the estimated ranks no longer vary significantly. The question now is, {\it how can we determine if the curves of the estimated ranks have stabilized?} \cuttlefish{} tracks the sequences of stable ranks for each layer at each epoch, \ie $\varrho = \{r^0, r^1, \dots, r^t\}$. \cuttlefish{} measures the derivative of the estimated rank sequences for all layer weights ($\diff{\varrho_l}{t}$) to detect when they cease to change significantly, using a condition: $\diff{\varrho_l}{t}\leq \epsilon, \forall l \in \{K+1, \dots, L-1\}$, where $\epsilon$ is a close-to-zero rank stabilization threshold. 

\vspace{-3mm}
\subsection{Determining $K$ for hybrid architectures}\label{sec:determine-k}
\begin{figure}[ht]
    \vspace{-3mm}
    \centering
    \includegraphics[width=0.45\textwidth]{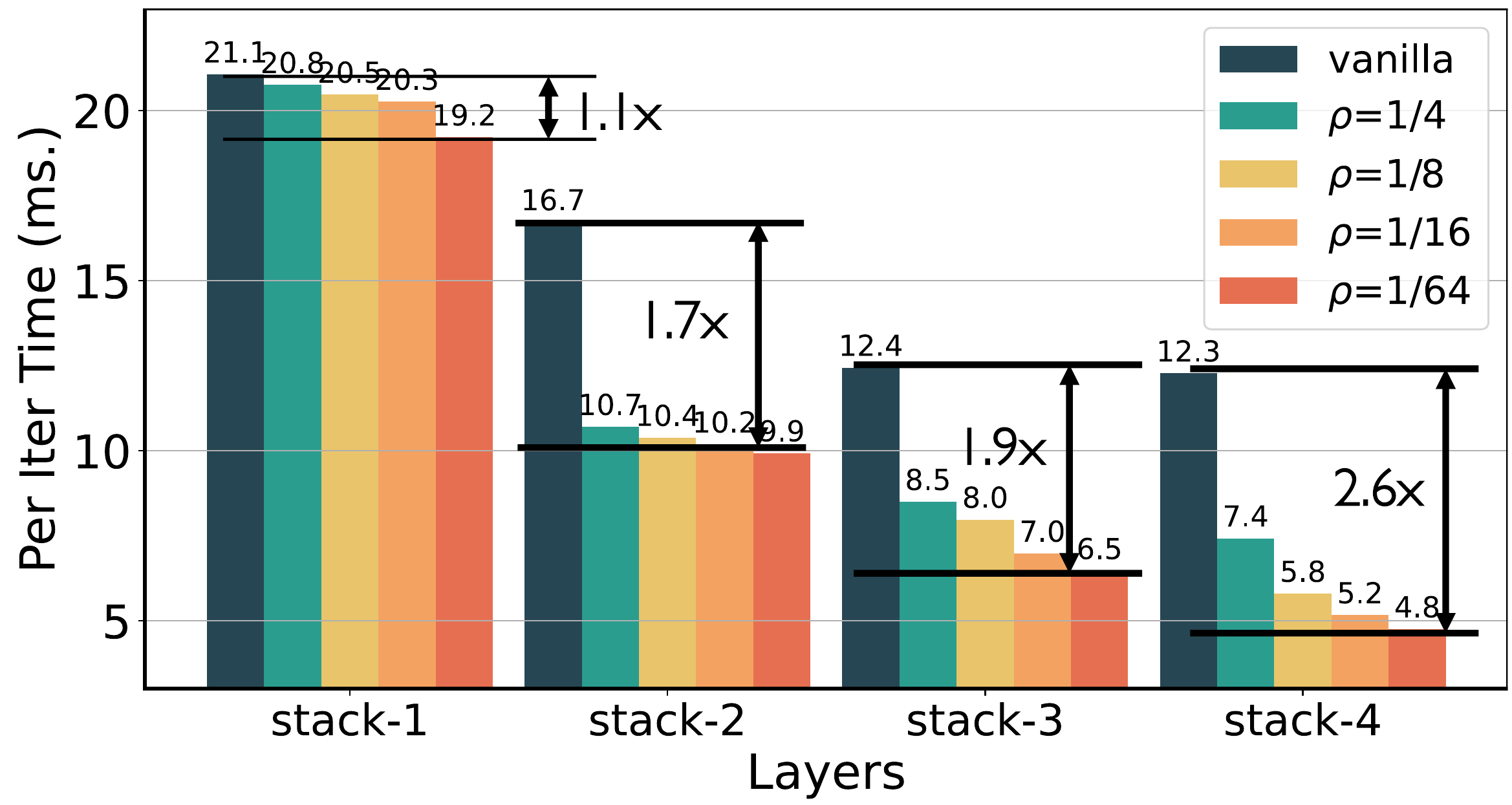}
    \vspace{-4.5mm}
    \caption{The per-iteration forward time in milliseconds, benchmarked using ResNet-18 on CIFAR-10 with a batch size of 1,024 on an EC2 p3.2xlarge instance.}
    \label{fig:k-selection-benchmark}
\vspace{-4.5mm}
\end{figure}

\vspace{-1.5mm}
$K$ balances the final accuracy and model compression rate. However, discerning the relationship between $K$ and final accuracy without fully training the model to convergence is challenging and impractical for achieving faster training speeds. For each task, \cuttlefish{} conducts lightweight profiling to measure the runtime of the low-rank NN when factorizing each layer stack, as layers within the same stack have identical weights and input sizes, and assesses whether it results in a significant speedup. \cuttlefish{} only performs factorization (with profiling rank ratio candidates: $\bar \rho$) if it leads to meaningful acceleration (determined by a threshold $\upsilon$). For example, if $\text{full-rank time}>1.5 \times \text{factorized time}$ for $\upsilon=1.5$ when $\bar \rho=\frac{1}{4}$, then \cuttlefish{} proceeds with factorization. Figure~\ref{fig:k-selection-benchmark} illustrates one example benchmark, where factorizing the first convolution stack (\ie layer $2$ to layer $5$) does not yield a substantial speedup. Consequently, \cuttlefish{} does not factorize these layers and returns $\hat K=6$.

\vspace{-4mm}
\paragraph{Why does not factorizing the initial layers result in a significant speedup?}
The reason for this can be attributed to the concept of {\it arithmetic intensity}, which is defined as the ratio of FLOPS to the bytes of data that must be accessed for a specific computation~\cite{jeffers2016intel}. When a certain layer has low arithmetic intensity, the GPU cannot operate at peak performance, and therefore, even if the FLOPs are substantially reduced, the actual speedup will not be significant. For a convolution layer, its arithmetic intensity is proportional to $\mathcal{O}(\frac{Bmnk^2HW}{mnk^2+BmHW})$ where $B, H, W$ stand for the batch size, height, and width of the input image. In convolution networks, it is generally assumed that the initial layers have small $mn$ but large $HW$, leading to $BmHW \gg mnk^2$ and $\mathcal{O}(\frac{Bmnk^2HW}{mnk^2+BmHW})\rightarrow \mathcal{O}(nk^2)$. For later layers, where $H, W$ are small and $mnk^2$ are large, and thus $mnk^2\gg BmHW$, $\mathcal{O}(\frac{Bmnk^2HW}{mnk^2+BmHW})\rightarrow \mathcal{O}(BHW)$. In the example shown in Figure~\ref{fig:k-selection-benchmark}, $nk^2=64\times9=576$ for the first layer stack, while for the last layer stack, $BHW=1024\times 8 \times 8 = 65,536 \gg 576$. Consequently, the bottom layers exhibit much higher arithmetic intensity, and as a result, factorization leads to significant speed improvements. For Transformers, where each layer has identical weight and input sizes (\ie the same arithmetic intensity), we consistently factorize all Transformer layers except for the word/image sequence embedding layers.

\vspace{-3mm}
\subsection{Putting things together}
\vspace{-1.5mm}
The main algorithm of \cuttlefish{} is outlined in Algorithm~\ref{Algo:cuttlefish}. \cuttlefish{} begins with profiling to determine $\hat K$. Following this, the training method commences with full-rank training until the stable ranks for the layers to be factorized converge, \ie at epoch $\hat E$. Subsequently, \cuttlefish{} factorizes the partially trained full-rank network using the converged scaled stable ranks $\mathcal{R}$ to obtain the factorized low-rank model. Finally, the low-rank model is trained until it reaches full convergence. In our experiments, we set $\epsilon=0.1$ and $\upsilon=1.5$.

\begin{minipage}{0.478\textwidth}
\vspace{-5.5mm}
\begin{algorithm}[H]
\small
\DontPrintSemicolon
\SetNoFillComment
\caption{\cuttlefish{}} \label{Algo:cuttlefish}
\KwIn{The dataset $\mathcal{D}$, initial full-rank neural network weights $\mathcal{W}^{0}=\{\mathbf{W}^0_1, \dots, \mathbf{W}^0_L\}$, the training algorithm $A(\cdot)$, such as SGD, Adam, etc., the total number of epochs $T$, and a rank stabilization threshold $\epsilon$.}
\KwOut{The trained low-rank factorized NN. }
Initialize the following: $\hat E=T$; $\mathcal{H}=\{\}$; $\varrho_{K+1},\dots, \varrho_{L-1}=\{\}, \dots, \{\};$\;
$\hat K = \text{Profiling}(\mathcal{D},\mathcal{W}, \tau, \bar \rho)$ (Algorithm~\ref{Algo:profiling}),
$\xi_l = \text{rank}(\mathbf{W}^0_l)/\texttt{stable rank}(\Sigma^0_l), \forall l \in \{1,\dots, L\}$.\;\\
\For{$t \in \{0, 1, 2, \dots, T-1\}$}{
    \uIf{$t \leq \hat E$}{
        $\mathcal{W}^{t+1} \gets A(\mathcal{W}^{t}, \mathcal{D})$\;\\
        \For{$\mathbf{W}_l \in \mathcal{W}^t$}{
            \uIf{$K+1 \leq l < L$}{
                $\mathbf{\tilde U}_l \mathbf{\Sigma}_l \mathbf{\tilde V}^\top_l = \texttt{SVD}(\mathbf{W}_l)$;\\
                $r_l = \texttt{stable rank}(\mathbf{\Sigma}_l)$,
                $\varrho_l = \varrho_l \cup \{r_l\}$;
            }
        }
        \uIf{$\diff{\varrho_l}{x}\leq \epsilon, \forall l \in \{K+1, \dots, L-1\}$}{
            $\hat E = t+1$;
        }
    }
    \uElseIf{$t = \hat E + 1$}{
        \For{$\mathbf{W}_l \in \mathcal{W}^t$}{
            \uIf{$K+1 \leq l < L$}{
                $\mathbf{\tilde U}_l \mathbf{\Sigma}_l \mathbf{\tilde V}^\top_l = \texttt{SVD}(\mathbf{W}_l)$;\\
                $r_l=\texttt{scaled stable rank}(\mathbf{\Sigma}_l, \xi_l)$;\\
                $\mathbf{U}_l=\mathbf{\tilde U}_l \mathbf{\Sigma}^{\frac{1}{2}}_l$, $\mathbf{V}^\top_l=\mathbf{\Sigma}^{\frac{1}{2}}_l \mathbf{\tilde V}^\top_l$;\\
                $\mathcal{H}=\mathcal{H} \cup \{\mathbf{U}_l[:,1:r_l], \mathbf{V}^\top_l[1:r_l,:]\}$ (with necessary NN weights reshaping);
            }
            \Else{
                $\mathcal{H} = \mathcal{H} \cup \{W_l\}$; 
            }
        }
        $\mathcal{H}^{t}=\mathcal{H}$;
        $\mathcal{H}^{t+1} \gets A(\mathcal{H}^{t}, \mathcal{D})$;
      }
    \Else{
        $\mathcal{H}^{t+1} \gets A(\mathcal{H}^{t}, \mathcal{D})$;
        }
    }
\end{algorithm}
\vspace{-5.5mm}
\end{minipage}

\begin{minipage}{0.478\textwidth}
\vspace{-5mm}
\begin{algorithm}[H]
\small
\DontPrintSemicolon
\SetNoFillComment
\caption{Profiling} \label{Algo:profiling}
\KwIn{The dataset $\mathcal{D}$, full-rank model weights $\mathcal{W}$, the profiling iterations $\tau$, and a profiling rank ratio candidates: $\bar \rho$.}
\KwOut{Determined $\hat K$.}
\texttt{init\_timer()}\;\\
\For{\text{ layer range }($l_{beg}$, $l_{end}$) $ \in $ \text{layer stacks}}{
        $\mathcal{H} = \texttt{factorize\_layer\_stack}(\mathcal{W}, \bar \rho,l_{beg}$, $l_{end} )$;\\
        {\it start\_time} = \texttt{timer.tic()};\\
        \For{$iter \in \{1,2,\dots, \tau\}$}{
            Train $\mathcal{H}$ for one iteration\;
        }
        {\it end\_time} = \texttt{timer.toc()};\\
        {\it avg\_low-rank\_time} = ({\it end\_time}-{\it start\_time})/$\tau$;\\
        {\it start\_time} = \texttt{timer.tic()};\\
        \For{$iter \in \{1,2,\dots, \tau\}$}{
            Train $\mathcal{W}$ for one iteration;
        }
        {\it end\_time} = \texttt{timer.toc()};\\
        {\it avg\_fullrank\_time} = ({\it end\_time}-{\it start\_time})/$\tau$;\\
        \uIf{$\text{fullrank\_time}>\upsilon \cdot \text{avg\_low-rank\_time}$}{
            $\hat K = l_{end}$\;
        }
    }
\end{algorithm}
\vspace{-7mm}
\end{minipage}
\vspace{-4mm}
\section{Experiments}\label{sec:experiment}
\vspace{-2mm}
We have developed an efficient implementation of \cuttlefish{} and conducted extensive experiments to evaluate its performance across various vision and natural language processing tasks. Our study focuses on the following aspects: (i) the sizes of factorized models \cuttlefish{} discovers and their final accuracy; (ii) the end-to-end training speedups that \cuttlefish{} achieves in comparison to full-rank training and other baseline methods; (iii) how the $\hat s$s found by \cuttlefish{} compare to manually tuned and explicitly learned ones. Our comprehensive experimental results demonstrate that \cuttlefish{} automatically selects all factorization hyperparameters during training on-the-fly, eliminating the need for multiple experimental trials for factorization hyperparameter tuning. More specifically, the experimental results reveal that \cuttlefish{} generates models up to 5.6$\times$ smaller than full-rank models, and attains up to a 1.2$\times$ faster end-to-end training process while preserving comparable accuracy. Moreover, \cuttlefish{} outperforms state-of-the-art low-rank model training methods and other prominent baselines.

\vspace{-4mm}
\subsection{Experimental setup and implementation details}
\vspace{-2mm}
\paragraph{Pre-training ML tasks.}
We conducted experiments on various computer vision pre-training tasks, including CIFAR-10, CIFAR-100~\cite{krizhevsky2009learning}, SVHN~\cite{netzer2011reading}, and ImageNet (ILSVRC2012)~\cite{deng2009imagenet}. For CIFAR-10, CIFAR-100, and SVHN~\cite{netzer2011reading}, we trained VGG-19-BN (referred to as VGG-19)~\cite{simonyan2014very} and ResNet-18~\cite{he2016deep}. In the case of the SVHN dataset, we utilized the original training images and excluded the additional images. For ImageNet, our experiments involved ResNet-50, WideResNet-50-2 (referred to as WideResNet-50), DeiT-base, and ResMLP-S36~\cite{he2016deep,zagoruyko2016wide,touvron2021training,touvron2021resmlp}. Further details about the machine learning tasks can be found in the appendix.

\vspace{-4.5mm}
\paragraph{Fine-tuning ML tasks.} We experiment on fine-tuning $\text{BERT}_{\text{BASE}}$ over the GLUE benchmark~\cite{wang2018glue}.

\vspace{-4.5mm}
\paragraph{Hyperparameters \& training schedule.} 
For VGG-19 and ResNet-18 training on CIFAR-10 and CIFAR-100, we train the NNs for 300 epochs, while on SVHN, we train the NNs for 200 epochs. We employ a batch size of 1,024 for CIFAR and SVHN tasks to achieve high arithmetic intensity. The initial learning rate is linearly scaled up from 0.1 to 0.8 within five epochs and then decayed at milestones of 50\% and 75\% of the total training epochs~\cite{goyal2017accurate}. For WideResNet-50 and ResNet-50 training on the ImageNet dataset, we follow the hyperparameter settings in~\cite{goyal2017accurate}, where the models are trained for 90 epochs with an initial learning rate of 0.1, which is decayed by a factor of 0.1 at epochs 30, 60, and 80 using a batch size of 256. In addition to~\cite{goyal2017accurate}, we apply label smoothing as described in~\cite{wang2021pufferfish}. For DeiT and ResMLP, we train them from scratch, adhering to the training schedule proposed in~\cite{touvron2021training}. For the GLUE fine-tuning benchmark, we follow the default hyperparameter setup in~\cite{devlin2018bert,jiao2020tinybert}. Since \cuttlefish{} generalizes spectral initialization (SI) and is compatible with Frobenius decay (FD), we deploy FD in conjunction with \cuttlefish{} when it contributes to better accuracy. Further details on hyperparameters and training schedules can be found in the appendix.

\vspace{-4.5mm}
\paragraph{Experimental environment.} We employ the NVIDIA NGC Docker container for software dependencies. Experiments for CIFAR, SVHN, and GLUE tasks are conducted on an EC2 p3.2xlarge instance (featuring a single V100 GPU) using FP32 precision. For BERT fine-tuning and ImageNet training of ResNet-50 and WideResNet-50, the experiments are carried out on an EC2 g4dn.metal instance (equipped with eight T4 GPUs) using FP32 precision. For ImageNet training of DeiT and ResMLP, the experiments are performed on a single p4d.24xlarge instance (housing eight A100 GPUs) with mixed-precision training enabled.

\begin{figure*}[ht]
    \vspace{-3mm}
    \centering
    \subfigure[VGG-19 on CIFAR-10]{\includegraphics[width=0.31\textwidth]{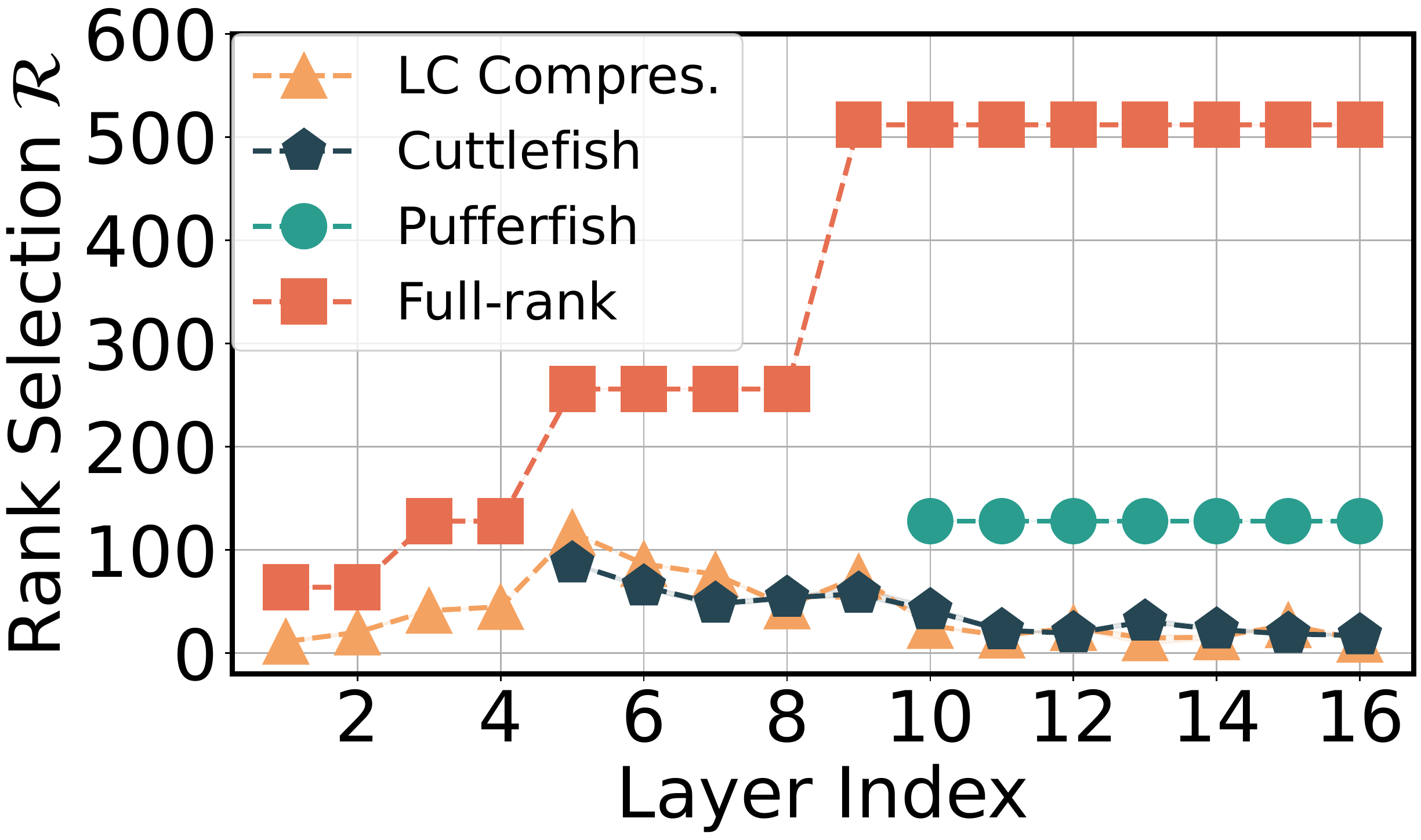}}
    \subfigure[VGG-19 on CIFAR-100]{\includegraphics[width=0.31\textwidth]{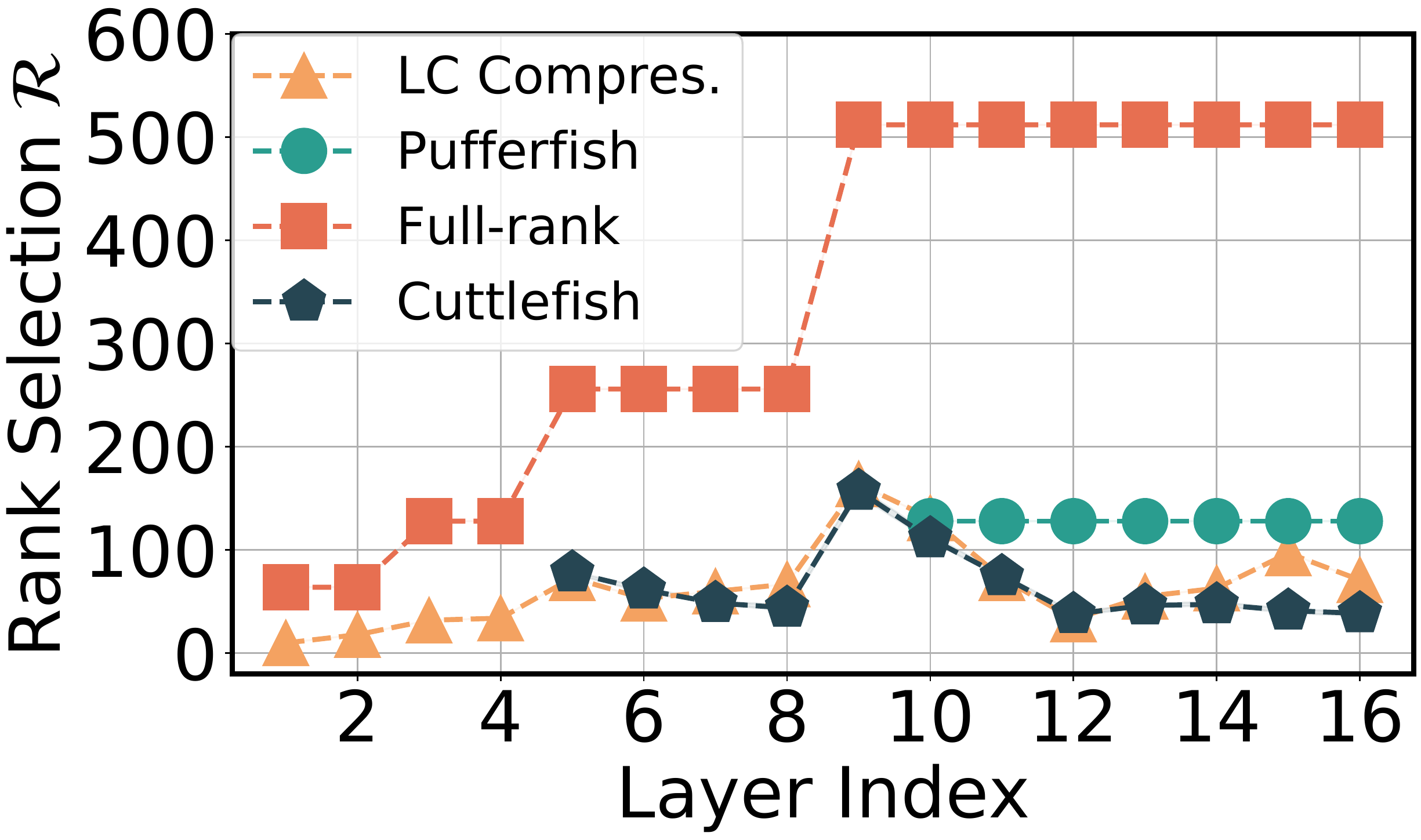}}
    \subfigure[VGG-19 on SVHN]{\includegraphics[width=0.31\textwidth]{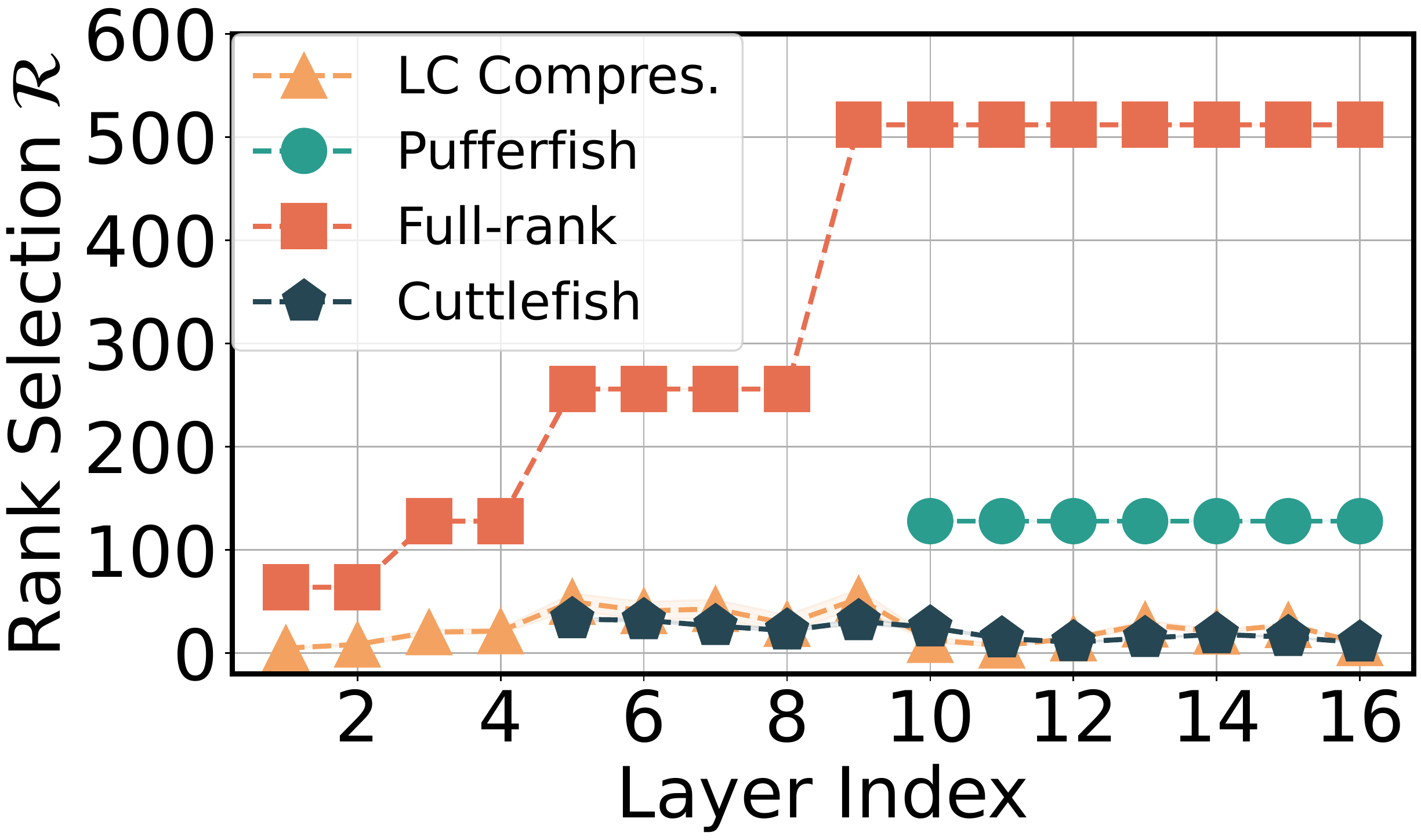}}
    \vspace{-5mm}
    \caption{Comparisons on the selected ranks $\mathcal{R}$ found by \cuttlefish{}, \pufferfish{}, LC compression, and full ranks for VGG-19 trained on CIFAR-10, CIFAR-100, and SVHN datasets.}
    \label{fig:rank-comparisons}
\vspace{-6mm}
\end{figure*}

\vspace{-4.5mm}
\paragraph{Baseline methods.}
We implement \cuttlefish{} and all considered baselines in PyTorch~\cite{paszke2019pytorch}. The baseline methods under consideration are: (i) \pufferfish{}, which employs manually tuned $s$~\cite{wang2021pufferfish}. To compare with \pufferfish{}, we use the same factorized ResNet-18, VGG-19, ResNet-50, and WideResNet-50 as reported in~\cite{wang2021pufferfish}. For DeiT and ResMLP models, not explored in the original \pufferfish{} paper, we adopt the same heuristic of using a fixed global rank ratio $\rho=\frac{1}{4}$, tuning $K$ to match the factorized model sizes found by \cuttlefish{}, and setting $E=80$ for the entire training epochs $T=300$~\cite{wang2021pufferfish}; (ii) the factorized low-rank training method with SI and FD proposed by~\cite{khodak2020initialization} (referred to as ``SI\&FD"), with $\rho$s of SI\&FD tuned to match the sizes of factorized models found by \cuttlefish{}; (iii) training time structured pruning method, or ``{\it early bird ticket}" (EB Train)~\cite{you2020drawing}; (iv) the IMP method where each pruning round follows the training length and prunes 20\% of the remaining model weights at each level, rewinding to the 6th epoch~\cite{frankle2019stabilizing}; (v) Gradient Signal Preservation (GraSP)~\cite{wang2020picking}; (vi) the learnable factorized low-rank training method, or LC model compression, where layer ranks $\mathcal{R}$ are explicitly optimized jointly with model weights $\mathcal{W}$ via an alternating optimization process~\cite{idelbayev2020low}. For GLUE fine-tuning, we compare \cuttlefish{} against DistillBERT and TinyBERT~\cite{sanh2019distilbert,jiao2020tinybert}; (vii) XNOR-Nets for training time quantization method~\cite{rastegari2016xnor}, based on the public PyTorch implementation available at \footnote{\url{https://github.com/jiecaoyu/XNOR-Net-PyTorch}}.

\vspace{-4.5mm}
\paragraph{\cuttlefish{} with FD.} To implement FD, \ie $\ell(\cdot)+\frac{\lambda}{2}\|\mathbf{U} \mathbf{V}^\top\|^2_F$ (where $\ell(\cdot)$ stands for the loss function), one has to compute the gradient on the regularization term, \ie
\vspace{-2mm}
{\small
\begin{align*}
    & \nabla_{\mathbf{U}} \frac{\lambda}{2}\|\mathbf{U} \mathbf{V}^\top\|^2_F = \lambda \mathbf{U}\mathbf{V}^\top\mathbf{V}; \nabla_{\mathbf{V}} \frac{\lambda}{2}\|\mathbf{U} \mathbf{V}^\top\|^2_F = \lambda \mathbf{U}^\top \mathbf{U}\mathbf{V}^\top    
\end{align*}\\[-0.5cm]}
where one can see there is a shared term $\mathbf{U}\mathbf{V}^\top$, which does not need to be recomputed. We optimize the implementation to only compute $\mathbf{U}\mathbf{V}^\top$ once. For the hybrid NN architectures, normal $\ell_2$ weight decay is conducted over full-rank layers when FD is enabled for factorized low-rank layers.

\vspace{-4.5mm}
\paragraph{Extra BatchNorm layers.} In experiments where FD is not enabled for \cuttlefish{}, we incorporate an extra BatchNorm (BN) layer following the $\mathbf{U}$ layer, \ie $\text{BN}_{\mathbf{V}}(\text{BN}_{\mathbf{U}}(\mathbf{x} \mathbf{U}) \mathbf{V}^\top)$, drawing inspiration from the network architecture design of MobileNets~\cite{howard2017mobilenets}. We also apply this approach to \pufferfish{}. 

\vspace{-4mm}
\subsection{Experimental results and analysis}
\vspace{-2mm}
\paragraph{How does \cuttlefish{} $s$ compare to manually tuned/learned ones?} 
A crucial question to consider is the appearance of the $s$ returned by \cuttlefish{}. We display the $\mathcal{R}$s discovered by \cuttlefish{}, \pufferfish{}, and LC compression for VGG-19 trained on CIFAR-10, CIFAR-100, and SVHN datasets (results presented in Figure~\ref{fig:rank-comparisons}). Here, it is evident that \cuttlefish{} provides a selection of $\mathcal{R}$ that closely aligns with explicitly trained rank selections, \ie LC compression, where rank selection and low-rank model weights are jointly learned during model training. This demonstrates the effectiveness of the rank selection heuristic employed by \cuttlefish{}.

\vspace{-4.5mm}
\paragraph{Parameter reduction and model accuracy.}
We thoroughly investigate the effectiveness of \cuttlefish{} and conduct extensive comparisons against the baselines, with results displayed in Tables \ref{table:cifar-main-results}, \ref{table:imagenet-main-results-2}, \ref{table:imagenet-main-results}, and \ref{table:lm-glue}. The primary observation is that \cuttlefish{} successfully reduces the number of parameters while only causing minimal loss in accuracy. Notably, for VGG-19 trained on CIFAR-10, \cuttlefish{} identifies a factorized model that is 10.8$\times$ smaller than the full-rank (vanilla) VGG-19 model, while achieving even better validation accuracy. In comparison to \pufferfish{} (shown in Table \ref{table:cifar-main-results}), \cuttlefish{} discovers a factorized low-rank model that is 4.4$\times$ smaller with similar final model accuracy for VGG-19 trained on CIFAR-10. To achieve a factorized model of the same size, SI\&FD does not always yield comparable accuracy to the full-rank model. For instance, for CIFAR-10 and CIFAR-100 trained on VGG-19, SI\&FD results in a non-trivial accuracy drop of 1.2\% and 1.8\%, respectively, because $K=1$ is always used in SI\&FD, which negatively affects the final model accuracy. On ImageNet, \cuttlefish{} attains smaller factorized ResNet-50 (0.5M fewer parameters) and WideResNet-50 (2.7M fewer parameters) with higher top-1 and top-5 validation accuracy compared to \pufferfish{}. For DeiT and ResMLP, we use a fixed rank ratio $\rho=\frac{1}{4}$ and tune the $K$s of \pufferfish{} to match the factorized low-rank model sizes of \cuttlefish{} for fair comparisons. \pufferfish{} factorized DeiT and ResMLP consistently result in inferior model accuracy compared to \cuttlefish{}. This occurs because the model weights of DeiT and ResMLP are less likely to be low rank, so using $\rho=\frac{1}{4}$ following the original \pufferfish{} heuristic leads to overly aggressive rank estimations. \cuttlefish{}, in contrast, detects this through a more appropriate rank estimation heuristic.
\begin{table*}[ht]
    \caption{The results, averaged across three independent trials with different random seeds, showcase the performance of \cuttlefish{} and other baselines on ResNet-18 and VGG-19 trained over CIFAR-10 and CIFAR-100 using a batch size of 1,024. The runtime benchmark is conducted on a single EC2 p3.2xlarge instance. $\dagger$: The SI\&FD baseline is tuned such that the model size is comparable (albeit slightly larger) to the models that \cuttlefish{} discovers. $\star$: \cuttlefish{} is tested with both FD enabled and disabled, and the results with the best accuracy are reported in the table. $\P$: XNOR-Net employs binary weights and activations; although the overall number of trainable parameters remains the same as the vanilla network, each model weight is quantized from 32-bit to 1-bit. Therefore, we report a compression rate of $3.125\%$ for XNOR-Nets. A comprehensive ablation study can be found in the appendix.} 
    \label{table:cifar-main-results}
    \vspace{-1.5mm}
    \begin{center}
    \small{
		\begin{tabular}{lllllll}
		\toprule
		& \multicolumn{3}{c}{CIFAR-10} & \multicolumn{3}{c}{CIFAR-100} \bigstrut\\
		\rowcolor{Gray} \midrule \textbf{Model:} ResNet-18 
		 & Params. ($M$) & Val. Acc. ($\%$) & Time (hrs.) & Params. ($M$) & Val. Acc. ($\%$) & Time (hrs.)
		\\
		\midrule
		\rowcolor{LightCyan} Full-rank  & $11.2$ (\textcolor{DarkGray}{100\%}) & $94.41_{\pm 0.14}$ &$0.82$ (\textcolor{DarkGray}{1$\times$})& $11.2$ (\textcolor{DarkGray}{100\%}) & $75.95_{\pm 0.23}$ &$0.82$ (\textcolor{DarkGray}{1$\times$})\bigstrut\\
		\pufferfish{} & $3.3$ (\textcolor{DarkGray}{29.9\%}) & $94.18_{\pm 0.15}$ &$0.70$ (\textcolor{DarkGray}{1.16$\times$}) & $3.4$ (\textcolor{DarkGray}{30.2\%}) & $72.43_{\pm 0.18}$& $0.70$ (\textcolor{DarkGray}{1.17$\times$}) \bigstrut\\
		\rowcolor{LightCyan} SI\&FD $^\dagger$ & $2.1$ (\textcolor{DarkGray}{18.5\%}) & $94.38_{\pm 0.03}$ & $0.59$ (\textcolor{DarkGray}{1.39$\times$}) &$2.7$ (\textcolor{DarkGray}{24.1\%}) &$75.80_{\pm0.17}$&$0.70$ (\textcolor{DarkGray}{1.16$\times$}) \bigstrut\\
		IMP & $1.9$ (\textcolor{DarkGray}{16.8\%}) &$95.04_{\pm 0.03}$&$6.55$ (\textcolor{DarkGray}{0.13$\times$}) &$2.4$ (\textcolor{DarkGray}{21.0\%}) &$75.51_{\pm 0.09}$& $5.73$ (\textcolor{DarkGray}{0.14$\times$})\bigstrut\\
		\rowcolor{LightCyan} XNOR-Net$^\P$ & $11.2$ (\textcolor{DarkGray}{3.1\%}) & $88.76_{\pm 0.14}$ &$3.61$ (\textcolor{DarkGray}{0.23$\times$}) & $11.2$ (\textcolor{DarkGray}{3.1\%}) & $57.23_{\pm 0.40}$ & $3.62$ (\textcolor{DarkGray}{0.23$\times$}) \bigstrut\\
		 \cuttlefish{}$^\star$ & $2.0$ (\textcolor{DarkGray}{17.9\%}) & $94.73_{\pm 0.08}$ &$0.70$ (\textcolor{DarkGray}{1.18$\times$}) & $2.6$ (\textcolor{DarkGray}{23.4\%}) & $75.57_{\pm 0.24}$ & $0.69$ (\textcolor{DarkGray}{1.19$\times$}) \bigstrut\\
		\rowcolor{Gray} \midrule \textbf{Model:} VGG-19
		& Params. ($M$) & Val. Acc. ($\%$) &Time (hrs.) & Params. ($M$) & Val. Acc. ($\%$) & Time (hrs.)
		\\
		\midrule
		\rowcolor{LightCyan} Full-rank  & $20.0$ (\textcolor{DarkGray}{100\%}) & $93.41_{\pm 0.15}$& $0.50$ (\textcolor{DarkGray}{1$\times$}) & $20.1$ (\textcolor{DarkGray}{100\%}) & $72.17_{\pm 0.37}$ &$0.49$ (\textcolor{DarkGray}{1$\times$})\bigstrut\\
		\pufferfish{} & $8.1$ (\textcolor{DarkGray}{40.5\%}) & $93.36_{\pm 0.09}$ & $0.46$ (\textcolor{DarkGray}{1.09$\times$}) & $8.2$ (\textcolor{DarkGray}{40.6\%}) & $72.43_{\pm 0.18}$ & $0.46$ (\textcolor{DarkGray}{1.09$\times$}) \bigstrut\\
		\rowcolor{LightCyan} SI\&FD$^\dagger$ & $2.0$ (\textcolor{DarkGray}{10.0\%}) & $92.23_{\pm 0.08}$ & $0.34$ (\textcolor{DarkGray}{1.44$\times$})&$3.3$ (\textcolor{DarkGray}{16.5\%}) &$70.42_{\pm 0.48}$&$0.39$ (\textcolor{DarkGray}{1.26$\times$})\bigstrut\\
		 LC Compress. & $1.7$ (\textcolor{DarkGray}{8.7\%}) & $93.23_{\pm 0.15}$ & $5.9$ (\textcolor{DarkGray}{0.08$\times$}) &$3.8$ (\textcolor{DarkGray}{19.0\%}) &$71.51_{\pm 0.07}$&$15.98$ (\textcolor{DarkGray}{0.03$\times$}) \bigstrut\\
		\rowcolor{LightCyan} IMP & $1.7$ (\textcolor{DarkGray}{8.6\%}) &$93.68_{\pm 0.28}$&$5.48$ (\textcolor{DarkGray}{0.09$\times$}) &$3.4$ (\textcolor{DarkGray}{16.8\%}) &$73.39_{\pm 0.32}$&$3.96$ (\textcolor{DarkGray}{0.13$\times$}) \bigstrut\\
       XNOR-Net$^\P$ & $20.0$ (\textcolor{DarkGray}{3.1\%}) & $86.61_{\pm 0.10}$ &$1.43$ (\textcolor{DarkGray}{0.35$\times$}) & $20.1$ (\textcolor{DarkGray}{3.1\%}) & $49.07_{\pm 0.28}$ & $1.43$ (\textcolor{DarkGray}{0.35$\times$}) \bigstrut\\
	\rowcolor{LightCyan} \cuttlefish{}$^\star$ & $1.9$ (\textcolor{DarkGray}{9.3\%}) & $93.54_{\pm 0.10}$& $0.42$ (\textcolor{DarkGray}{1.18$\times$}) & $3.3$ (\textcolor{DarkGray}{16.3\%}) & $72.23_{\pm 0.09}$ &$0.44$(\textcolor{DarkGray}{1.14$\times$})\bigstrut\\
    \bottomrule
    \end{tabular}}%
    \end{center}
\vspace{-6mm}
\end{table*}

\begin{table*}[ht]
    \vspace{-4mm}
    \caption{The results presented include vanilla, \pufferfish{}, and \cuttlefish{} implementations of ResNet-50 and WideResNet-50, trained on ImageNet. The FLOPs numbers represent model inference latency, measured using simulated single-batch input with dimensions of $(3, 224, 224)$. Runtime benchmarks are conducted on EC2 g4dn.metal instances.}
        \vspace{-1.5mm}
	\label{table:imagenet-main-results-2}
	\begin{center}
	  \small{
		\begin{tabular}{llllll}
		\toprule 
		\rowcolor{Gray} & \# Params.  & Val. Acc. & Val. Acc.  & FLOPs & Time
		\\
        \rowcolor{Gray} & ($M$)  &  Top-1 & Top-5  & (G) & (hrs.)
		\bigstrut\\
		\midrule
		\rowcolor{LightCyan} WideResNet-50 & $68.9$ (\textcolor{DarkGray}{100\%}) & $78.1$ & $94.0$ & $11.4$ & $147.8$ (\textcolor{DarkGray}{1$\times$}) \bigstrut\\
		\pufferfish{} & $40.0$ (\textcolor{DarkGray}{58.1\%}) & $77.86_{\pm 0.05}$ & $93.97_{\pm 0.05}$ & $10.0$ & $112.7$ (\textcolor{DarkGray}{1.31$\times$}) \bigstrut\\
		\rowcolor{LightCyan} \cuttlefish{} & $37.4$ (\textcolor{DarkGray}{54.3\%}) & $78.0_{\pm 0.06}$ & $94.04_{\pm 0.09}$ & $10.0$ & $112.7$ (\textcolor{DarkGray}{1.31$\times$}) \bigstrut\\
		\midrule
		ResNet-50 & $25.6$ (\textcolor{DarkGray}{100\%}) & $77.0$ & $93.4$ & $4.1$ & $67.0$ (\textcolor{DarkGray}{1$\times$}) \bigstrut\\
		\rowcolor{LightCyan} \pufferfish{} & $15.2$ (\textcolor{DarkGray}{59.5\%}) & $76.36_{\pm 0.03}$ & $93.21_{\pm 0.03}$ & $3.6$ & $55.6$ (\textcolor{DarkGray}{1.20$\times$}) \bigstrut\\
		\cuttlefish{} & $14.7$ (\textcolor{DarkGray}{57.4\%}) & $76.44_{\pm 0.16}$ & $93.21_{\pm 0.03}$ & $3.6$ & $56.7$ (\textcolor{DarkGray}{1.18$\times$})\bigstrut\\
		\bottomrule
		\end{tabular}}
	\end{center}
\vspace{-6mm}
\end{table*}

\begin{table}[ht]
    \vspace{-2mm}
    \caption{The results for vanilla, \pufferfish{}, and \cuttlefish{} implementations of DeiT-base and ResMLP-S36, trained on ImageNet, are presented. FLOPs numbers, which measure model inference latency, are determined using simulated single-batch input with dimensions of $(3, 224, 224)$.}
        \vspace{-1.5mm}
	\label{table:imagenet-main-results}
	\begin{center}
	  \scriptsize{
		\begin{tabular}{lllll}
		\toprule 
		\rowcolor{Gray} & \# Params.  &  Val. Acc. & Val. Acc.  & FLOPs
		\\
        \rowcolor{Gray} & ($M$)  &  Top-1 & Top-5 & (G) \bigstrut\\
		\midrule
		\rowcolor{LightCyan} DeiT-base & $86.6$ & $81.8$ & $95.6$ & $17.6$ \bigstrut\\
		\pufferfish{} & $58.3$ & $81.15_{\pm 0.04}$ & $95.58_{\pm 0.04}$ & $12.0$ \bigstrut\\
		\rowcolor{LightCyan} \cuttlefish{} & $58.3$ & $81.52_{\pm 0.03}$ & $95.59_{\pm 0.04}$ & $12.0$ \bigstrut\\
		\midrule
		ResMLP-S36 & $44.7$ & $80.1$ & $95.0$ & $8.9$ \bigstrut\\
		\rowcolor{LightCyan} \pufferfish{} & $29.3$ & $77.78_{\pm 0.20}$ & $94.00_{\pm 0.06}$ & $5.9$ \bigstrut\\
		\cuttlefish{} & $29.4$ & $78.94_{\pm 0.04}$ & $94.52_{\pm 0.05}$ & $5.8$ \bigstrut\\
		\bottomrule
		\end{tabular}}
	\end{center}
\vspace{-8mm}
\end{table}

\begin{table*}[ht]
\caption{Vanilla $\text{BERT}_{\text{BASE}}$, Distill $\text{BERT}$, Tiny $\text{BERT}_6$, as well as \cuttlefish{} $\text{BERT}_{\text{BASE}}$ are evaluated on the GLUE benchmark. F1 scores are used as the metric for QQP and MRPC, while Spearman correlations are reported for STS-B, and accuracy scores are reported for the remaining tasks.}
\label{table:lm-glue}
\vspace{-0.5mm}
    \begin{center}
		 \small{
		\begin{tabular}{ccccccccccc}
		\rowcolor{Gray} \toprule  
		 \textbf{Model} & \# Params. ($M$) & MNLI &  QNLI & QQP & RTE & SST-2 & MRPC & CoLA & STS-B & Avg.
		\bigstrut\\
		\midrule
		\rowcolor{LightCyan} $\text{BERT}_{\text{BASE}}$ & $108.3$& ${\bf 83.9}/{\bf 84.4}$ & ${\bf 90.9}$ & ${\bf 87.6}$ & $66.8$ & $92.2$ & $88.6$ & ${\bf 60.1}$ & $88.6$ & ${\bf 82.5}$\bigstrut\\
        Distill $\text{BERT}$ & $65.8$& $81.1/82.0$ & $89.1$ & $86.2$ & $57.8$ & $90.6$ & $88.6$ & $47.3$ & $83.4$ & $78.4$ \bigstrut\\
        \rowcolor{LightCyan} Tiny $\text{BERT}_6$ & $67.0$& ${\bf 83.9}/83.8$ & $90.6$ & $86.8$ & ${\bf 72.9}$ & $91.5$ & ${\bf 90.6}$ & $46.2$ & ${\bf 89.3}$ & $81.7$ \bigstrut\\
		\cuttlefish{} &${\bf 48.8}$& $83.7/{\bf 84.4}$ & $90.8$ & $86.7$ & $67.0$ & ${\bf 92.3}$ & $88.4$ & $56.8$ & $87.9$ & $82.0$ \bigstrut\\
	\bottomrule
    \end{tabular}}%
    \end{center}
\vspace{-6mm}
\end{table*}

\vspace{-4.5mm}
\paragraph{End-to-end runtime and computational complexity.} As discussed in Section~\ref{sec:determine-k}, factorized low-rank training achieves substantial speedups when arithmetic intensity is high. One way to achieve high arithmetic intensity is by using a large batch size for training. Consequently, we use a large batch size of 1,024 and measure the end-to-end training time for the experiments on CIFAR. The results, presented in Table~\ref{table:cifar-main-results}, demonstrate that \cuttlefish{} consistently leads to faster end-to-end training time (including full-rank epochs and all other overhead computations, such as profiling and stable rank computing) compared to full-rank training. For instance, \cuttlefish{} achieves 1.2$\times$ end-to-end training speedups on both ResNet-18 and VGG-19 trained on CIFAR-10. \cuttlefish{} yields comparable runtime to \pufferfish{} for ResNet-18 and faster runtime on VGG-19 because it finds smaller $K$ for VGG-19, \ie $K=4$, while \pufferfish{} uses $K=9$. SI\&FD achieves faster runtime than \cuttlefish{} due to its use of $K=1$ (and generally higher computational complexities for the initial convolution layers). However, employing such an aggressive value for $K$ inevitably results in accuracy loss, as discussed earlier. Both FC compression and IMP require heavy computation to achieve small models, which are significantly slower than full-rank training. XNOR-Nets employ binary model weights and activations, which results in a reduction of final accuracy for tasks compared to dense networks. Ideally, the use of binarized weights and activations should greatly speed up model training and substantially decrease memory consumption during the process. However, PyTorch lacks an efficient implementation of a binarized convolution operator. Consequently, our experiments utilized FP32 networks and activations to simulate binary networks, leading to a notably slower runtime compared to conventional FP32 training. This is because each layer's output necessitates binarization, and model weights must be re-binarized for every iteration. For ImageNet experiments (presented in Table~\ref{table:imagenet-main-results-2}), the memory footprints are high, limiting us to a batch size of 256. \cuttlefish{} identifies factorized ResNet-50 and WideResNet-50 models that achieve 1.2$\times$ and 1.3$\times$ end-to-end speedups for ImageNet training, respectively. Although the factorized models found by \cuttlefish{} are comparable to \pufferfish{}, it eliminates the need for extensive hyperparameter tuning for such large-scale tasks.

\vspace{-4mm}
\subsection{Computation overheads introduced by \cuttlefish{}.} 
\vspace{-2mm}
\paragraph{Computational overheads of profiling.} The profiling process in \cuttlefish{} is a lightweight operation. For instance, with ResNet-18 on the CIFAR-10 dataset, we perform profiling using $\tau=11$ iterations and exclude the running time for the first iteration for both full-rank and low-rank models (\ie running 22 iterations in total). We then average the running time figures for the remaining 10 iterations for benchmarking purposes. Averaged from three independent runs, the entire profiling stage takes 3.98 seconds, which accounts for a mere 0.16\% of the total running time of \cuttlefish{} on ResNet-18 trained on the CIFAR-10 dataset.

\vspace{-4.5mm}
\paragraph{Computational overheads of rank estimation.} It is important to emphasize that \cuttlefish{} needs to compute the singular values of the entire network weights at the end of each epoch. It is worth noting that to calculate stable ranks, only singular values are required, rather than singular vectors. This process can be accelerated by leveraging APIs, such as \texttt{scipy.linalg.svdvals}, which only compute singular values of a given matrix. Taking ResNet-18 trained on CIFAR-10 as an example, the average time taken for rank estimation using the stable rank is 0.49 seconds per epoch. For \cuttlefish{}, which requires $E=82.3$ epochs (on average) for full-rank training, the stable rank estimation takes a total of 39.97 seconds, accounting for 1.6\% of the entire end-to-end running time.

\begin{table*}[ht]
\vspace{-1mm}
\caption{In this ablation study, we evaluate the impact of extra BNs on ResNet-18 and VGG-19 trained on CIFAR-10 and CIFAR-100, as well as ResNet-50 trained on the ImageNet dataset. The end-to-end and per-iteration running time results are measured on a single p3.2xlarge EC2 instance for ResNet-18 and VGG-19, and a single g4dn.metal EC2 instance for ResNet-50. The results are averaged from three independent experiments using different random seeds.}
\label{table:cifar10-imagenet-extra-bns-ablation}
\vspace{-1.5mm}
\begin{center}
\scriptsize{
		\begin{tabular}{lllllllll}
		\toprule
		& \multicolumn{4}{c}{CIFAR-10} & \multicolumn{4}{c}{CIFAR-100} \bigstrut\\
		\rowcolor{Gray} \midrule \textbf{Model:} 
		 & Params.  & Val. Acc.  & Time End2end  & Time Iter.  & Params.  & Val. Acc. & Time End2end  & Time Iter. \\
            \rowcolor{Gray} 
		 ResNet-18  & ($M$) & ($\%$) & (hrs.) & (ms) & ($M$) & ($\%$) & (hrs.) & (ms)
		\\
		\midrule
		\rowcolor{LightCyan} w/ extra BNs  & $2.02$ & $94.36_{\pm 0.07}$ &$0.716$ & $163.42_{\pm 0.51}$ & $2.62$ & $73.64_{\pm 0.26}$ &$0.719$ & $164.72_{\pm 1.63}$ \bigstrut\\
		w/o extra BNs & $1.97$ & $94.32_{\pm 0.22}$ &$0.706$ & $158.94_{\pm 0.53}$ & $2.60$ & $73.77_{\pm 0.14}$& $0.689$ & $158.53_{\pm 1.32}$\bigstrut\\
		\rowcolor{Gray} \midrule \textbf{Model:} 
		 & Params.  & Val. Acc.  & Time End2end  & Time Iter.  & Params.  & Val. Acc. & Time End2end  & Time Iter. \\
            \rowcolor{Gray} 
		VGG-19 & ($M$) & ($\%$) & (hrs.) & (ms) & ($M$) & ($\%$) & (hrs.) & (ms)
		\\
		\midrule
		\rowcolor{LightCyan} w/ extra BNs  & $1.86$ & $93.54_{\pm 0.10}$& $0.422$ & $85.53_{\pm 0.59}$ & $3.31$ & $71.99_{\pm 0.02}$ &$0.436$ & $91.76_{\pm 0.52}$ \bigstrut\\
		w/o extra BNs & $1.86$ & $93.49_{\pm 0.08}$ & $0.419$ & $84.55_{\pm 0.26}$ & $3.31$ & $72.15_{\pm 0.24}$ & $0.432$ & $89.90_{\pm 0.18}$ \bigstrut\\
        \bottomrule
        \toprule
        \rowcolor{Gray}  ResNet-50 on ImageNet
		 & \multicolumn{2}{c}{Params. ($M$)} & \multicolumn{2}{c}{Top-1 Val. Acc. ($\%$)}  & \multicolumn{2}{c}{Time End2end (hrs.)}  & \multicolumn{2}{c}{Time Iter. (sec.)}   \bigstrut\\
        \midrule
        \rowcolor{LightCyan} w/ extra BNs
		 & \multicolumn{2}{c}{$14.7$} & \multicolumn{2}{c}{$76.44_{\pm 0.16}$}  & \multicolumn{2}{c}{$56.7$}  & \multicolumn{2}{c}{$0.43_{\pm 0.002}$}   \bigstrut\\
         w/o extra BNs
		 & \multicolumn{2}{c}{$14.7$} & \multicolumn{2}{c}{$76.23_{\pm 0.21}$}  & \multicolumn{2}{c}{$55.6$}  & \multicolumn{2}{c}{$0.42_{\pm 0.003}$}   \bigstrut\\
		\bottomrule
		\end{tabular}}%
	\end{center}
\vspace{-7mm}
\end{table*}

\vspace{-4mm}
\subsection{Ablation Study}
\vspace{-2mm}
\paragraph{Accuracy and runtime efficiency of extra BNs.} We conduct additional ablation studies to examine the influence of incorporating extra BN layers on \cuttlefish{} performance. In our primary experiments, we use FD and disable extra BN layers to ensure accurate FD gradient computation when FD leads to better accuracy. Our ablation studies involve training ResNet-18 and VGG-19 on CIFAR-10 and CIFAR-100 datasets, as well as ResNet-50 on ImageNet, and evaluating model sizes, best validation accuracy (top-1 for ImageNet), end-to-end training time, and per-iteration time on low-rank models. The hyperparameters used in the ablation studies are consistent with those used for the main results in the Experiment section. The ablation study results, shown in Table~\ref{table:cifar10-imagenet-extra-bns-ablation}, reveal that adding extra BN layers generally leads to a marginally larger model size and slower per-iteration and end-to-end runtimes. For example, when training ResNet-18 on CIFAR-10 without extra BNs, the end-to-end training time is 1.4\% faster, and the per-iteration runtime is 2.8\% faster. The impact of extra BNs on final validation accuracy varies across experiments: enabling extra BNs slightly improves accuracy for ResNet-18 and VGG-19 on CIFAR-10, but not for CIFAR-100. However, for the ImageNet experiment, adding extra BNs leads to a non-trivial increase in model accuracy by an average of 0.21\% across three independent runs with different random seeds. This improvement is significant, considering it relates to top-1 accuracy for a 1000-class classification problem. There are two potential explanations for why extra BNs help improve accuracy for ImageNet experiments more than CIFAR experiments: 1) The model capacity of ResNet-18 and VGG-19 seems sufficiently large for CIFAR datasets, allowing high compression rates (\eg, 5$\times$-10$\times$). In contrast, for ResNet-50 on ImageNet, the model capacity appears inadequate. \cuttlefish{}, in this case, does not achieve exceptionally high compression rates (\ie less than 2$\times$). Consequently, the inclusion of extra BNs appears to provide the low-rank factorized model with additional capacity to enhance its accuracy. 2) For CIFAR experiments, we used a batch size of 1024, constrained by GPU memory, while a batch size of 256 was employed for ImageNet experiments. It is possible that extra BNs offer more substantial benefits in smaller batch settings. Note that for Transformer model-based experiments, we do not enable extra BNs as LayerNorm is commonly used instead of BNs, which is beyond the scope of this ablation study.

\begin{figure}[ht]
    \vspace{-2mm}
    \centering
    \includegraphics[width=0.493\textwidth]{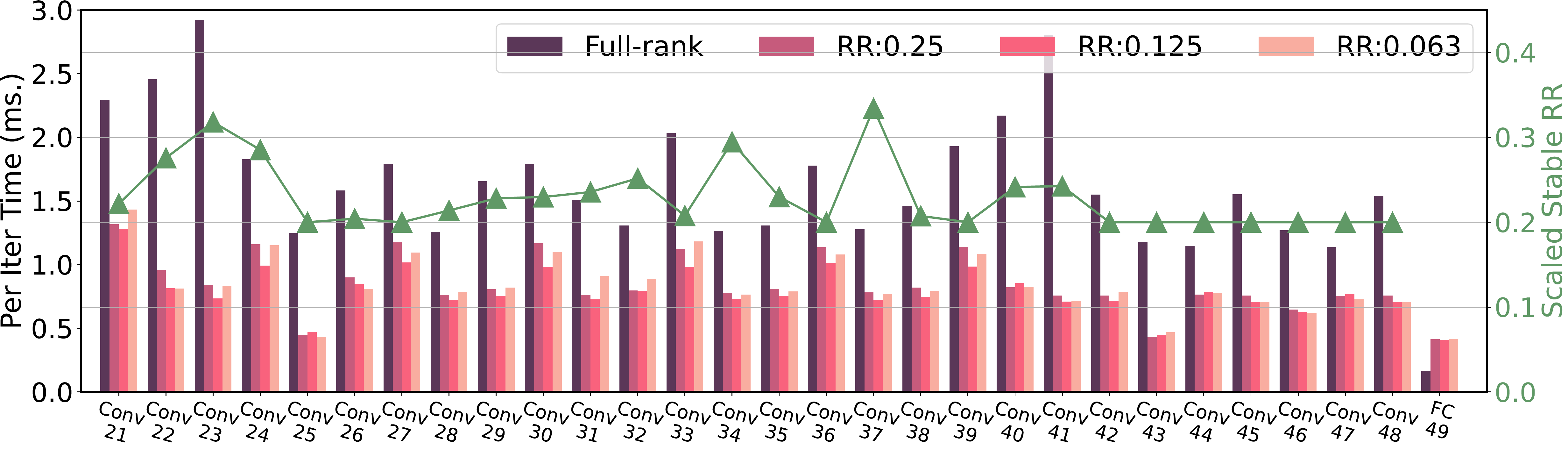}\\
    \includegraphics[width=0.493\textwidth]{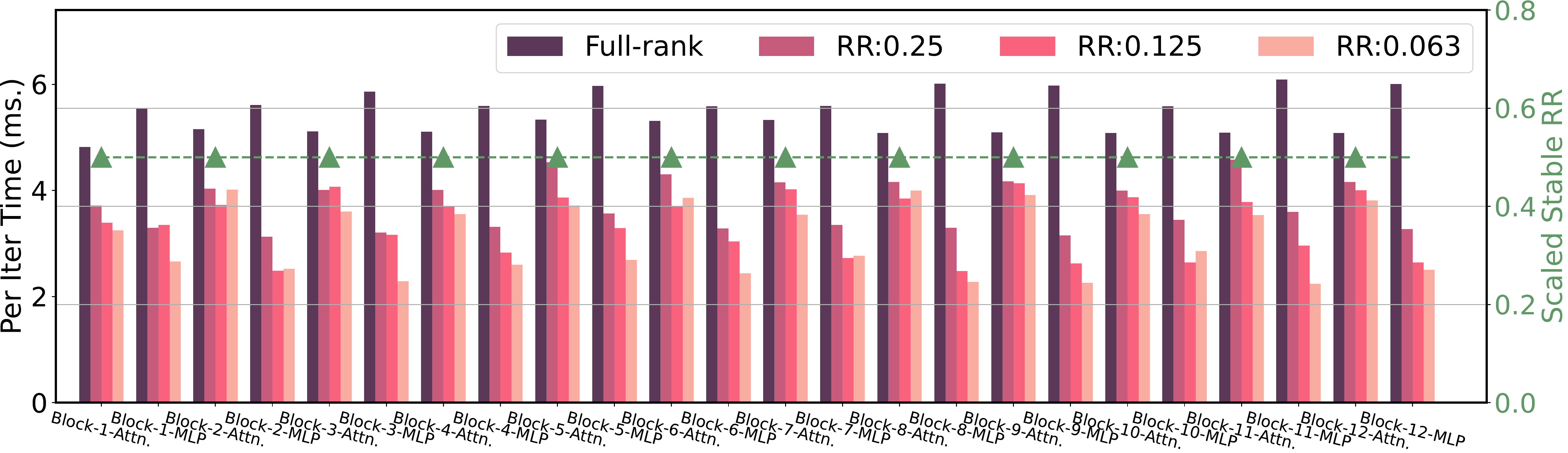}
    \vspace{-8mm}
    \caption{Ablation study on the layer-wise costs of (\textbf{Top}): ResNet-50 training on ImageNet (along with the scaled stable rank ratios selected by \cuttlefish{}); (\textbf{Bottom}): DeiT-Small training on ImageNet, batch size of both experiments are $128$, and time for both experiments are measured on an EC2 p3.2xlarge instance with a batch size of $128$. ``RR" stands for rank ratio in the figure.}
    \label{fig:layer-iter-time-per-layer}
\vspace{-1mm}
\end{figure}
\vspace{-4.5mm}
\paragraph{Effectiveness of low-rank factorization on various layer types.} In order to compare the efficiency of low-rank factorization against convolution, FC, MLP, and multi-head attention layers, we conducted an ablation study using ResNet-50 and DeiT-small on the ImageNet dataset. The per-iteration time of each layer was measured and the results are depicted in Figure~\ref{fig:layer-iter-time-per-layer} (for ResNet-50, we also illustrated the scaled stable rank ratios selected by \cuttlefish{}). We focused on forward computation time for this study, as it is well known that there is a constant factor between forward and backward computing time, and the former serves as a good proxy for per-iteration time. Due to space constraints, we only plotted the results for the 21st convolution layer onwards in the ResNet-50 experiments, although meaningful speedups were observed for the first 20 layers as well. In the case of convolution layers, our experiments revealed an average speedup of $2.1\times$ across all 49 layers when using a rank ratio of $\frac{1}{4}$. However, we observed that the last FC layer actually slowed down when factorized, regardless of the rank ratio used. This could be attributed to the small size of the FC layer, which incurs a large kernel launching overhead when split into two smaller layers, thereby nullifying any computation cost savings. Our experiments with DeiT showed that factorizing both multi-head attention and MLP layers resulted in significant speedups for all 12 Transformer encoder blocks. Additionally, factorizing the MLP layer led to greater speedup gains compared to factorizing the multi-head attention layer. Specifically, factorizing the multi-head attention layer resulted in $1.26\times$ speedups on average, while factorizing the MLP layer resulted in $1.73\times$ speedups on average for all 12 blocks at a rank ratio of $\frac{1}{4}$.

\vspace{-4mm}
\subsection{Limitations of \cuttlefish{}}\label{sec:limitation}
\vspace{-2mm}
A limitation of \cuttlefish{} is that the hyperparameters $s$ it tunes are influenced by the randomness of the training algorithm and model initialization. Consequently, different trial runs may not yield factorized models with identical sizes (although the variance is minimal). This can potentially impact exact reproducibility.
\vspace{-4mm}
\section{Conclusion}
\vspace{-2mm}
We present \cuttlefish{}, an automated low-rank training method that eliminates the need for tuning additional factorization hyperparameters, \ie $\mathcal{S}=(E, K, \mathcal{R})$. \cuttlefish{} leverages two key insights related to the emergence of stable ranks during training and the actual speedup gains achieved by factorizing different NN layers. Utilizing these insights, it designs heuristics for automatically selecting $s \in \mathcal{S}$. Our extensive experiments demonstrate that \cuttlefish{} identifies low-rank models that are not only smaller, but also yield better final accuracy in most cases when compared to state-of-the-art low-rank training methods. 

\subsection*{Acknowledgments}
We thank our shepherd, Bilge Acun, and the anonymous MLSys reviewers for their valuable insights and recommendations, which have enhanced our work. This research has been graciously funded by ONR Grant No. N00014-21-1-2806, three Sony Faculty Innovation Awards, NSF IIS1563887, NSF CCF1629559, NSF IIS1617583, NSF IIS1955532, NSF CNS2008248, NSF IIS2123952, and NSF BCS2040381, and NGA HM04762010002.

\bibliography{cf-mlsys2023}
\bibliographystyle{mlsys2023}

\newpage
\clearpage
\appendix
\section{Artifact Appendix}

\subsection{Abstract}
We have made available the necessary artifacts to replicate all results presented in the paper. Our experiments utilize Amazon EC2 computing resources, including {\it p3.2xlarge} (for ResNet-18 and VGG-19 training on CIFAR-10 and CIFAR-100 datasets, as well as BERT fine-tuning on the GLUE benchmark), {\it g4dn.metal} (for example, ResNet-50 and WideResNet-50 training on ImageNet), and {\it p4d.24xlarge} (for DeiT-base and ResMLP execution on ImageNet) instances. Additionally, we employ the NVIDIA driver and Docker to construct the software stack.

To facilitate the replication of all reported experiments, we provide scripts in our GitHub repository, accessible at \url{https://github.com/hwang595/Cuttlefish}. Running those provided scripts will set up and launch experiments to reproduce our experimental results. For ease of reproducibility, we also offer a public Amazon Machine Image (AMI) – {\it ami-05c0b3732203032b3} (in the region of {\it US West (Oregon)}) where experimental environments and the ImageNet dataset, which is time-consuming to download and set up are prepared.

\subsection{Artifact check-list (meta-information)}

In this section, we offer meta-information regarding the configuration, datasets, implementation, and other aspects of our artifacts.

{\small
\begin{itemize}
  \item {\bf Algorithm: } Our artifact encompasses the \cuttlefish{} automatic low-rank training schedule, along with the baseline methods compared in the main paper, such as \pufferfish{}, SI\&FD, XNOR-Net, GraSP, and others.
  \item {\bf Program: } N/A
  \item {\bf Compilation: } All methods and baselines are implemented in PyTorch, therefore requiring no compilation. 
  \item {\bf Transformations: } N/A
  \item {\bf Binary: } N/A
  \item {\bf Data set: }  For our main experiments, we employ CIFAR-10, CIFAR-100, SVHN, ImageNet (ILSVRC 2012), and GLUE datasets. As preparing the ImageNet dataset can be time-consuming, we provide a ready-to-use public AMI - \textit{ami-05c0b3732203032b3} in the {\it US West (Oregon)} region for convenience.
  \item {\bf Run-time environment: } N/A
  \item {\bf Hardware: } Our experiments were conducted using Amazon EC2 instances, specifically \textit{p3.2xlarge}, \textit{g4dn.metal}, and \textit{p4d.24xlarge}.
  \item {\bf Run-time state: } N/A
  \item {\bf Execution: } We provide scripts to execute and launch the experiments. Detailed descriptions and instructions can be found in the README of our GitHub repository.
  \item {\bf Metrics: } We collect metrics such as the number of parameters, validation accuracy (or similar scores for assessing model quality), wall-clock time (including end-to-end and per iteration/epoch durations), and computational complexity (measured in FLOPS).
  \item {\bf Output: } Our existing code writes checkpoints to the local disk and also prints experimental outputs/logs directly.
  \item {\bf Experiments: } N/A
  \item {\bf How much disk space required (approximately)?: } Around 1 Terabyte of disk space.
  \item {\bf How much time is needed to prepare workflow (approximately)?: } Setting up the experimental environment should take less than an hour. Downloading the ImageNet (ILSVRC 2012) dataset can take a few days. If the evaluators have access to AWS, we have also provided a public AMI - \textit{ami-05c0b3732203032b3} (in the region of {\it US West (Oregon)}) which has the datasets and dependencies pre-installed.
  \item {\bf How much time is needed to complete experiments (approximately)?: } Completing the CIFAR-10 and CIFAR-100 experiments with \cuttlefish{} typically takes less than an hour for each task (see Table~\ref{table:cifar-main-results} for details). BERT fine-tuning experiments on all GLUE datasets require approximately a few hours, while ImageNet experiments may take several days to a week to reach full convergence for each method.
  \item {\bf Publicly available?: } All our code is publicly available on the GitHub repository:  \url{https://github.com/hwang595/Cuttlefish}. For easy setup on AWS, we also provide a public AMI - with ID \emph{ami-05c0b3732203032b3} (in the region of {\it US West (Oregon)}), which can be used to launch large-scale experiments. 
  \item {\bf Code licenses (if publicly available)?: } N/A
  \item {\bf Data licenses (if publicly available)?: } We use CIFAR-10, CIFAR-100, SVHN, Imagenet (ILSVRC 2012) datasets and the GLUE benchmark which come with their own licenses. All datasets are publicly available.
  \item {\bf Workflow framework used?: } N/A
  \item {\bf Archived (provide DOI)?: } We use Zenedo to create a publicly accessible archival repository for our GitHub repository, \ie \url{https://doi.org/10.5281/zenodo.7884872}.
\end{itemize}

\subsection{Description}
We have made available the code necessary to replicate all the experiments presented in this paper through a public GitHub repository that contains comprehensive documentation, allowing users to seamlessly execute the experiments.

\subsubsection{How delivered}
Our entire codebase is available on the GitHub repository: \url{https://github.com/hwang595/Cuttlefish}. To facilitate easy setup on AWS, we offer a public AMI - identified by \emph{ami-05c0b3732203032b3} (in the region of {\it US West (Oregon)}) - which can be utilized to launch large-scale experiments.

\subsubsection{Hardware dependencies}
For all our experiments we used \textit{p3.2xlarge}, \textit{g4dn.metal}, and \textit{p4d.24xlarge} Amazon EC2 instances. To reproduce our results, one instance of each type is required.

\subsubsection{Software dependencies} We established our experimental environments using Docker, configuring them through PyTorch Docker containers from NVIDIA GPU Cloud (NGC). The experiments involving the CIFAR-10, CIFAR-100, SVHN, and ImageNet datasets were based on the NGC container {\it nvcr.io/nvidia/pytorch:20.07-py3}, while those focused on the GLUE benchmark utilized the {\it nvcr.io/nvidia/pytorch:22.01-py3} container. Since there are additional software dependencies not included in the Docker containers, we have supplied installation scripts, accompanied by instructions in the README file of our GitHub repository, to facilitate the installation of these necessary components.

\subsubsection{Data sets} For all our experiments we used CIFAR-10, CIFAR-100, SVHN, ImageNet (ILSVRC 2012), and GLUE datasets. For CIFAR-10, CIFAR-100, SVHN, and GLUE datasets, our code will automatically download them. For the ImageNet dataset, we have it ready and provided via the public AMI - with ID \emph{ami-05c0b3732203032b3} (in the region of {\it US West (Oregon)}).

\subsection{Installation}
In the GitHub README, we offer comprehensive instructions for installing dependencies and configuring the Docker environments.

\subsection{Experiment workflow}
We have provided a detailed README along with our GitHub repository which provides bash scripts to execute and launch the experiments. 
\subsection{Evaluation and expected result}
During the experiment, logs containing details such as accuracy and running time will be displayed directly. However, given the inherent variability in machine learning tasks and the diversity of hardware and system configurations, it is important to note that the exact accuracy and running time figures reported in the paper may not be replicated. Nevertheless, by using the artifacts provided, one can expect to achieve comparable accuracy and running time outcomes.

\subsection{Experiment customization}
The experiment can be customized by trying on different hardware setups. One example of this will be to run these experiments on slower GPUs (or other hardware, \eg CPUs). Another option would be to try to support more model architectures using the heuristics of \cuttlefish{} (an interesting example will be adopting \cuttlefish{} for some recently designed large language models). 

\subsection{Notes}
If the evaluator utilizes our provided AMI, the initial disk initialization will take an extended period of time during the first run.
\newpage
\clearpage
\newcommand{\blocka}[2]{\multirow{3}{*}{\(\left[\begin{array}{c}\text{3$\times$3, #1}\\[-.1em] \text{3$\times$3, #1} \end{array}\right]\)$\times$#2}
}
\newcommand{\blockb}[3]{\multirow{3}{*}{\(\left[\begin{array}{c}\text{1$\times$1, #2}\\[-.1em] \text{$3\times$3, #2}\\[-.1em] \text{1$\times$1, #1}\end{array}\right]\)$\times$#3}
}

\section{Experimental setup}\label{sec:exp_setup}
In this section, we delve into the specifics of the datasets~\ref{appdx:dataset} and model architectures~\ref{appdx:model} employed in our experiments. Additionally, we elaborate on the software environment~\ref{subset:software-details} and the implementation details of all methods included in our experiments~\ref{subset:implementation-details}. Our code can be accessed at \url{https://github.com/hwang595/Cuttlefish}.

\subsection{Dataset}\label{appdx:dataset}
We carried out experiments across multiple computer vision and NLP tasks to evaluate the performance of \cuttlefish{} and the other considered baselines. In this section, we discuss the specifics of each task in greater detail.

\paragraph{CIFAR-10 and CIFAR-100.} Both CIFAR-10 and CIFAR-100 comprise 60,000 color images with a resolution of 32$\times$32 pixels, where 50,000 images are used for training and 10,000 for validation (since there is no provided test set for CIFAR-10 and CIFAR-100, we follow the convention of other papers by conducting experiments and reporting the highest achievable accuracy on the validation datasets)~\cite{krizhevsky2009learning}. CIFAR-10 and CIFAR-100 involve 10-class and 100-class classification tasks, respectively. For data processing, we employ standard augmentation techniques: channel-wise normalization, random horizontal flipping, and random cropping. Each color channel is normalized with the following mean and standard deviation values: $\mu_r = 0.485, \mu_g = 0.456, \mu_b = 0.406$; $\sigma_r = 0.229, \sigma_g = 0.224, \sigma_b = 0.225$. The normalization of each channel pixel is achieved by subtracting the corresponding channel's mean value and dividing by the color channel's standard deviation.

\paragraph{SVHN.} The SVHN dataset comprises 73,257 training images and 26,032 validation images, all of which are colored with a resolution of 32$\times$32 pixels~\cite{netzer2011reading}. This classification dataset consists of 10 classes. As there is no clear test-validation split for the SVHN dataset, we follow the convention of other papers by conducting experiments and reporting the highest achievable accuracy on the validation datasets. There are 531,131 additional images for SVHN, but we do not include them in our experiments for this paper. For data processing, we employ the same data augmentation and normalization techniques used for CIFAR-10 and CIFAR-100, as described above.

\paragraph{ImageNet (ILSVRC 2012).} The ImageNet ILSVRC 2012 dataset consists of 1,281,167 colored training images spanning 1,000 classes and 50,000 colored validation images, also covering 1,000 classes~\cite{deng2009imagenet}. Augmentation techniques include normalization, random rotation, and random horizontal flip. The training images are randomly resized and cropped to a resolution of 224$\times$224 using the torchvision API \texttt{torchvision.transforms.RandomResizedCrop}. The validation images are first resized to a resolution of 256$\times$256 and then center cropped to a resolution of 224$\times$224. Each color channel is normalized with the following mean and standard deviation values: $\mu_r = 0.485, \mu_g = 0.456, \mu_b = 0.406$; $\sigma_r = 0.229, \sigma_g = 0.224, \sigma_b = 0.225$. Each channel pixel is normalized by subtracting the corresponding channel's mean value and then dividing by the color channel's standard deviation.

\paragraph{GLUE benchmark.} For the GLUE benchmark, we utilize the data preparation and pre-processing pipeline implemented by Hugging Face~\footnote{\url{https://github.com/huggingface/transformers/tree/main/examples/pytorch/text-classification}}. In accordance with prior work~\cite{devlin2018bert,jiao2020tinybert,hu2021lora}, we exclude the problematic WNLI downstream task.

\subsection{Model architectures}\label{appdx:model}
In this section, we provide a summary of the network architectures utilized in our experiments.

\paragraph{ResNet-18, ResNet-50, and WideResNet-50-2.} The ResNet-18 and ResNet-50 architectures are derived from the original design with minor modifications~\cite{he2016deep}. The WideResNet-50-2 adheres to the original wide residual network design~\cite{zagoruyko2016wide}. As we employed ResNet-18 for CIFAR-10 classification, we adjusted the initial convolution layer to use a $3\times 3$ convolution with padding at $1$ and stride at $1$. Our ResNet-18 implementation follows the GitHub repository~\footnote{\label{fn:kuangliu-cifar}\url{https://github.com/kuangliu/pytorch-cifar}}. For all ResNet-18, ResNet-50, and WideResNet-50-2 networks, the strides used for the four convolution layer stacks are respectively $1, 2, 2, 2$. Bias terms for all layers are deactivated (owing to the BatchNorm layers), except for the final FC layer.

\begin{table*}[ht]  
\small
\centering
\caption{The ResNet-18, ResNet-50, and WideResNet-50-2 network architectures used in our experiments. It should be noted that when using ResNet-18 for CIFAR-10 training, we make corresponding adjustments to the initial convolution layer. Each convolution layer is followed by a BatchNorm layer. In the notation used in this table, ``$7\times7, 64$" signifies that the convolution layer contains 64 convolution kernels, \ie each kernel has a dimension of $7\times 7$ and the output dimension is 64.}
\vspace{2mm}
{\footnotesize
\begin{tabular}{cccc} 
\toprule
\rowcolor{Gray} Model & ResNet-18 & ResNet-50 & WideResNet-50-2 \\
\midrule
\multirow{3}{*}{Conv 1} & 3$\times$3, 64 & 7$\times $7, 64 & 7$\times$7, 64\\
 & padding 1 & padding 3 & padding 3\\
 & stride 1 & stride 2& stride 2 \\
 & -& \multicolumn{2}{c}{Max Pool, kernel size 3, stride 2, padding 1}  \\
\midrule
\multirow{3}{*}{Layer stack 1} &  \blocka{64}{2} & \blockb{256}{64}{3} & \blockb{256}{128}{3} \\
& & &\\
& & &\\
\midrule
\multirow{3}{*}{Layer stack 2} & \blocka{128}{2} & \blockb{512}{128}{4} & \blockb{512}{256}{4} \\
& & &\\
& & &\\
\midrule
\multirow{3}{*}{Layer stack 3} & \blocka{256}{2} & \blockb{1024}{256}{6} & \blockb{1024}{512}{6} \\
& & &\\
& & &\\
\midrule
\multirow{3}{*}{Layer stack 4} & \blocka{512}{2} & \blockb{2048}{512}{3} & \blockb{2048}{1024}{3} \\
 & & & -\\
& & &\\
\midrule
\multirow{2}{*}{FC} & Avg Pool, kernel size $4$ & \multicolumn{2}{c}{Adaptive Avg Pool, output size $(1, 1)$} \\
 & $512 \times 10$ & $2048 \times 1000$ & $2048 \times 1000$\\
\bottomrule
\end{tabular}}
\vspace{-2mm}
\label{arch:resnet-typed}
\end{table*}

\paragraph{VGG-19-BN.} In our experiments, we employ the VGG-19-BN network architecture, which is a modified version of the original VGG-19~\citep{simonyan2014very}. The original VGG-19 network consists of 16 convolution layers and 3 FC layers, including the final linear classification layer. We adopt the VGG-19 architectures from~\cite{frankle2018lottery,khodak2020initialization}, which remove the first two FC layers following the last convolution layer while retaining the final linear classification layer. This results in a 17-layer architecture, but we continue to refer to it as VGG-19-BN since it stems from the original VGG-19 design. Another modification is replacing the max pooling layer after the last convolution layer (\texttt{conv16}) with an average pooling layer. The detailed architecture is displayed in Table~\ref{arch:vgg}. We follow the implementation of the \texttt{pytorch-cifar} GitHub repository mentioned above. Due to the BatchNorm layers, bias terms for all layers are deactivated, except for the final FC layer.

\begin{table*}[ht] 
    \centering
    \caption{A detailed description of the VGG-19 architecture in our experiments. After each convolution layer, a BatchNorm layer followed by a ReLU activation is included (though not shown in the table). The shapes for convolution layers are represented as $(m, n, k, k)$.}
    \vspace{2mm}
		 \small{
		 \resizebox{0.5\linewidth}{!}{
			\begin{tabular}{ccc}
				\toprule \rowcolor{Gray} \textbf{Parameter}
				& Shape &  Layer hyperparameter \bigstrut\\
				\midrule
				\textbf{layer1.conv1.weight} & $3 \times 64 \times 3 \times 3$ & stride:$1$;padding:$1$ \bigstrut\\
				\textbf{layer2.conv2.weight} & $64 \times 64 \times 3 \times 3$ & stride:$1$;padding:$1$  \bigstrut\\
				\textbf{pooling.max} & N/A & kernel size:$2$;stride:$2$  \bigstrut\\
				\textbf{layer3.conv3.weight} & $64\times 128 \times 3 \times 3$ & stride:$1$;padding:$1$ \bigstrut\\
				\textbf{layer4.conv4.weight} & $128\times 128 \times 3 \times 3$ & stride:$1$;padding:$1$ \bigstrut\\
                \textbf{pooling.max} & N/A & kernel size:$2$;stride:$2$  \bigstrut\\
				\textbf{layer5.conv5.weight} & $128 \times 256 \times 3 \times 3$ & stride:$1$;padding:$1$  \bigstrut\\
				\textbf{layer6.conv6.weight} & $256\times 256 \times 3 \times 3$ & stride:$1$;padding:$1$  \bigstrut\\
				\textbf{layer7.conv7.weight} & $256 \times 256 \times 3 \times 3$ & stride:$1$;padding:$1$  \bigstrut\\
				\textbf{layer8.conv8.weight} & $256 \times 256 \times 3 \times 3$ & stride:$1$;padding:$1$  \bigstrut\\
                \textbf{pooling.max} & N/A & kernel size:$2$;stride:$2$  \bigstrut\\
				\textbf{layer9.conv9.weight} & $256 \times 512 \times 3 \times 3$ & stride:$1$;padding:$1$  \bigstrut\\
				\textbf{layer10.conv10.weight} & $512 \times 512 \times 3 \times 3$ & stride:$1$;padding:$1$  \bigstrut\\
				\textbf{layer11.conv11.weight} & $512 \times 512 \times 3 \times 3$ & stride:$1$;padding:$1$  \bigstrut\\
				\textbf{layer12.conv12.weight} & $512 \times 512 \times 3 \times 3$ & stride:$1$;padding:$1$  \bigstrut\\
				\textbf{pooling.max} & N/A & kernel size:$2$;stride:$2$  \bigstrut\\
				\textbf{layer13.conv13.weight} & $512 \times 512 \times 3 \times 3$ & stride:$1$;padding:$1$  \bigstrut\\
				\textbf{layer14.conv14.weight} & $512 \times 512 \times 3 \times 3$ & stride:$1$;padding:$1$  \bigstrut\\
				\textbf{layer15.conv15.weight} & $512 \times 512 \times 3 \times 3$ & stride:$1$;padding:$1$  \bigstrut\\
				\textbf{layer16.conv16.weight} & $512 \times 512 \times 3 \times 3$ & stride:$1$;padding:$1$  \bigstrut\\
				\textbf{pooling.avg} & N/A & kernel size:$2$  \bigstrut\\
				\textbf{classifier.weight} & $512 \times 10$ & N/A  \bigstrut\\
				\textbf{classifier.bias} & $10$ & N/A  \bigstrut\\
				\bottomrule
			\end{tabular}}%
			}
	\label{arch:vgg}
\vspace{-2mm}
\end{table*}

\paragraph{DeiT and ResMLP.} Our implementations of DeiT-base and ResMLP-S36 are sourced directly from the model implementations provided by the Pytorch Image Models (\ie \texttt{timm}) library~\footnote{\url{https://github.com/rwightman/pytorch-image-models}}. For DeiT-base, we do not use the scaled ImageNet resolution version and we deactivate the distillation options. More specifically, we initiate the training of a \texttt{deit\_base\_patch16\_224} model from scratch, as provided by the \texttt{timm} library. For training, we employ the training method and hyperparameters specified in the original GitHub repository~\footnote{\url{https://github.com/facebookresearch/deit}}. For ResMLP-S36, we adhere to the same training methodology used for DeiT-base, utilizing the \texttt{resmlp\_36\_224} provided by the \texttt{timm} library.

\paragraph{$\text{BERT}_{\text{BASE}}$, DistillBERT, and TinyBERT.} The implementations of $\text{BERT}{\text{BASE}}$, DistillBERT, and $\text{TinyBERT}_6$ are directly provided by Hugging Face. For $\text{BERT}{\text{BASE}}$, we use the model named \texttt{bert-base-cased}. For DistillBERT, we employ the model named \texttt{distilbert-base-cased}. For $\text{BERT}_{\text{BASE}}$, we use the model named \texttt{bert-base-cased} again. For $\text{TinyBERT}_6$, we utilize the model named \texttt{huawei-noah/TinyBERT\_General\_6L\_768D}. All model names are supplied through the API of \texttt{--model\_name\_or\_path} in Hugging Face.

\subsection{Software details}\label{subset:software-details}
For the experiments on CIFAR-10, CIFAR-100, and SVHN, which include \cuttlefish{} and all considered baseline methods, our software setup is built on the NVIDIA NGC Docker container for PyTorch. We use the docker image \texttt{nvcr.io/nvidia/pytorch:20.07-py3} to set up the experiment environment on \texttt{p3.2xlarge} EC2 instances. The CUDA version we used is 11.6. For the $\text{BERT}_{\text{BASE}}$ fine-tuning experiment on the GLUE benchmark, we employ the docker image, \texttt{nvcr.io/nvidia/pytorch:22.01-py3}. We install Hugging Face with version \texttt{4.17.0.dev0}. 

\subsection{Implementation details}\label{subset:implementation-details}
For all our experiments, we set $$\texttt{torch.backends.cudnn.benchmark = True}$$ and $$\texttt{torch.backends.cudnn.deterministic = False}$$ to optimize the running speed of the experiments, as the cuDNN benchmark searches for the fastest low-level implementations. However, perfect reproducibility cannot be guaranteed under this setup. To measure runtime, we employ \texttt{torch.cuda.Event(enable\_timing=True)} to determine the elapsed time between two CUDA Event records. We fine-tune the \texttt{num\_workers} and enable \texttt{pin\_memory} for all experiments to achieve faster end-to-end runtimes. For the DeiT and ResMLP experiments on ImageNet, we enable mixed-precision training using PyTorch AMP.

\paragraph{Examples of factorized low-rank layers.} As discussed, a full-rank layer $\mathbf{W}$ can be factorized to obtain $\mathbf{U}$ and $\mathbf{V}^\top$. For fully connected (or linear) layers in ResMLP, BERT, and DeiT models, the dimensions of $\mathbf{U}$ and $\mathbf{V}^\top$ are straightforward. For instance, if $\mathbf{W}$ is a $(784, 784)$ linear projection implemented using \texttt{nn.Linear} in PyTorch, then $\mathbf{U}$ and $\mathbf{V}^\top$ can be implemented using $(784, r)$ and $(r, 784)$ as input dimensions for \texttt{nn.Linear} in PyTorch. For convolution layers, we use $1\times 1$ convolution for $\mathbf{V}^\top$, following the suggestion of ~\cite{wang2021pufferfish}. As a concrete example, when factorizing the 16th convolution layer in the VGG-19 architecture we used with $r=32$, our factorization results in $\mathbf{U}$ as a $512\times 32 \times 3 \times 3$ dimensional \texttt{nn.Conv2d} layer in PyTorch and $\mathbf{V}^\top$ as a $32\times 512 \times 1 \times 1$ dimensional \texttt{nn.Conv2d} layer in PyTorch. Other factorized low-rank convolution networks in our implementations follow the same approach.

\section{Details on hyperparameters.}\label{sec:add-hyper-params}
In this section, we discuss general-purpose hyperparameters, such as learning rate and training schedule, used in our experiments for each task. Additionally, we discuss the hyperparameter setup of \cuttlefish{} and the details on the final $\hat s \in \mathcal{S}$ that \cuttlefish{} manages to find for each experiment.

\subsection{General purpose hyperparameters}
\paragraph{CIFAR-10 and CIFAR-100.} For the CIFAR-10 and CIFAR-100 tasks, training on ResNet-18 and VGG-19, we train for $T=300$ epochs in total using the SGD optimizer with momentum (Nesterov momentum disabled). We use a batch size of $B=1,024$, scaling an initial learning rate from $\gamma = 0.1$ to $\gamma = 0.8$ in 5 epochs linearly, following the procedure proposed in~\cite{goyal2017accurate}. We set momentum and weight decay coefficients at $0.9$ and $1\times10^{-4}$, respectively (weight decay is disabled for BatchNorm layers). We employ a multi-step learning rate schedule, decaying the learning rate by a factor of $0.1$ at the $150$-th and $225$-th epochs (to $\gamma = 0.08$ and $\gamma = 0.008$ respectively). The methodology we follow for this multi-step learning rate decay is to decay the learning rate by a factor of $0.1$ at the points of $50\%$ and $75\%$ of the entire training epochs.

\paragraph{SVHN.} For the SVHN task, training on ResNet-18 and VGG-19, we train for $T=200$ epochs in total using the SGD optimizer with momentum (Nesterov momentum disabled). We train SVHN for a shorter number of epochs, as it is a relatively easier task compared to CIFAR-10 and CIFAR-100. We use a batch size of $B=1,024$, scaling an initial learning rate from $\gamma = 0.1$ to $\gamma = 0.8$ in 5 epochs linearly, following the procedure proposed in~\cite{goyal2017accurate}. We set momentum and weight decay coefficients at $0.9$ and $1\times10^{-4}$, respectively (weight decay is disabled for BatchNorm layers). We employ a multi-step learning rate schedule, decaying the learning rate by a factor of $0.1$ at the $100$-th and $150$-th epochs (to $\gamma = 0.08$ and $\gamma = 0.008$ respectively).

\paragraph{ImageNet (ILSVRC 2012) on CNNs.} For the ImageNet training task on ResNet-50 and WideResNet-50-2, we train for $T=90$ epochs in total using the SGD optimizer with momentum (Nesterov momentum disabled) and a batch size of $256$. We set momentum and weight decay coefficients at $0.9$ and $1\times10^{-4}$, respectively (weight decay is disabled for BatchNorm layers). We employ a multi-step learning rate schedule, decaying the learning rate by a factor of $0.1$ at the $30$-th, $60$-th, and $80$-th epochs (to $\gamma = 0.01$, $\gamma = 0.001$, and $0.0001$ respectively), following~\cite{goyal2017accurate}. We adopt the label smoothing technique with a probability of $0.1$ to achieve better final accuracy and to compare against \pufferfish{}~\cite{wang2021pufferfish}. For the comparison against \pufferfish{}, GraSP, and EBTrain, we disable the learning rate decay at the $80$-th epoch and label smoothing to align the experiment setup.

\paragraph{ImageNet (ILSVRC 2012) on DeiT and ResMLP.} For ImageNet training on DeiT and ResMLP, we adhere to the training schedule proposed by~\cite{touvron2021training}, wherein the models are trained on ImageNet directly from scratch. Our experiments adopt the model initialization, data augmentation, and Exponential Moving Average (EMA) methods suggested by~\cite{touvron2021training}. We do not enable distillation for DeiT and ResMLP. The AdamW optimizer is used for our experiments~\cite{loshchilov2019decoupled}. More details on the hyperparameter values can be found at~\url{https://github.com/facebookresearch/deit/blob/main/README_deit.md}, where we use the default hyperparameter setup and set the batch size to $256$.

\paragraph{BERT fine-tuning on the GLUE benchmark.} We directly utilize the training script provided by Hugging Face at~\url{https://github.com/huggingface/transformers/tree/main/examples/pytorch/text-classification}. We set the maximum sequence length to $128$ and the batch size to $32$, using the AdamW optimizer~\cite{loshchilov2019decoupled} for all downstream tasks in the GLUE benchmark. For each downstream task in GLUE, we conduct a small hyperparameter sweep within the range of $\{1e-5, 2e-5, 3e-5, 4e-5\}$, employing early stopping. For the relatively small downstream task MRPC, we fine-tune for $5$ epochs, while for all other downstream tasks, we fine-tune for $3$ epochs. We disable weight decay and learning rate warm-up during the fine-tuning process.

\subsection{\cuttlefish{} hyperparameters}
\paragraph{CIFAR-10, CIFAR-100, and SVHN.} For ResNet-18 and VGG-19 models trained on CIFAR-10, CIFAR-100, and SVHN datasets, we report the details of the hyperparameters $\hat s \in \mathcal{S}$ discovered by \cuttlefish{} and the manually tuned hyperparameters used by \pufferfish{} in Table~\ref{table:appendix-hyper-params}. It is evident that \cuttlefish{} identifies larger $K$ values and smaller $E$ values (except for CIFAR-10) compared to \pufferfish{} for ResNet-18, while for VGG-19, \cuttlefish{} discovers smaller $E$ values with slightly longer $E$ values than \pufferfish{}. Additionally, the selected ranks, i.e., $\mathcal{R}$, for \cuttlefish{}, \pufferfish{}, and LC compression (only for VGG-19) methods are illustrated in Figure~\ref{fig:rank-plot-cifar-svhn}. Notably, \cuttlefish{} consistently returns lower estimated ranks for bottom layers than \pufferfish{}, as these layers contain greater redundancy. The most striking observation from Figure~\ref{fig:rank-plot-cifar-svhn} is that the $\mathcal{R}$ values returned by \cuttlefish{} closely align with the explicitly learned $\mathcal{R}$ values of LC compression, demonstrating \cuttlefish{}'s effectiveness. Another interesting finding from Figure~\ref{fig:rank-plot-cifar-svhn} is that more challenging tasks generally require higher ranks for the factorized low-rank network to achieve satisfactory accuracy. For example, both \cuttlefish{} and LC compression identify larger $\mathcal{R}$ values for CIFAR-100 and smaller $\mathcal{R}$ values for SVHN and CIFAR-10. This is because CIFAR-100, a 100-class classification task, is more challenging than CIFAR-10 and SVHN for a given model architecture.
\begin{table*}[ht]
	\caption{The hyperparameters $\hat s \in \mathcal{S}$ optimized by \cuttlefish{} for ResNet-18 and VGG-19 models trained on CIFAR-10 use a batch size of 1,024. The runtime benchmark was conducted on a single EC2 p3.2xlarge instance, equipped with one V100 GPU.}
	\label{table:appendix-hyper-params}
	\begin{center}
      \small{
		\begin{tabular}{ccccccc}
		\toprule
		& \multicolumn{2}{c}{CIFAR-10} & \multicolumn{2}{c}{CIFAR-100} & \multicolumn{2}{c}{SVHN}\bigstrut\\
		\rowcolor{Gray} \midrule \textbf{Model:} ResNet-18
		 & $E$ & $K$ & $E$ & $K$ & $E$ & $K$ 
		\bigstrut\\
		\midrule
		\cuttlefish{} & $82.3_{\pm 10.1}$ & $5$  & $55.7_{\pm 8.7}$ & $5$  & $61.0_{\pm 2.2}$ & $5$\bigstrut\\
		\rowcolor{LightCyan} \pufferfish{} & $80$ & $3$ & $80$ & $3$ & $80$ & $3$ \bigstrut\\
		SI\&FD & $0$ & $1$ & $0$ & $1$ & $0$ & $1$ \bigstrut\\
		\rowcolor{Gray} \midrule \textbf{Model:} VGG-19
		 & $E$ & $K$ & $E$ & $K$ & $E$ & $K$ 
		\bigstrut\\
		\midrule
		\cuttlefish{} & $97.3_{\pm 1.2}$ & $4$ & $86.0_{\pm 5.7}$ & $4$ & $84.0_{\pm 0.8}$ & $4$ \bigstrut\\
		\rowcolor{LightCyan} \pufferfish{} & $80$ & $9$ & $80$ & $9$ & $80$ & $9$ \bigstrut\\
		SI\&FD & $0$ & $1$ & $0$ & $1$ & $0$ & $1$ \bigstrut\\
		\bottomrule
		\end{tabular}}%
	\end{center}
\vspace{-1mm}
\end{table*}

\begin{figure*}[ht]
    \centering
    \subfigure[ResNet-18 on CIFAR-10]{\includegraphics[width=0.315\textwidth]{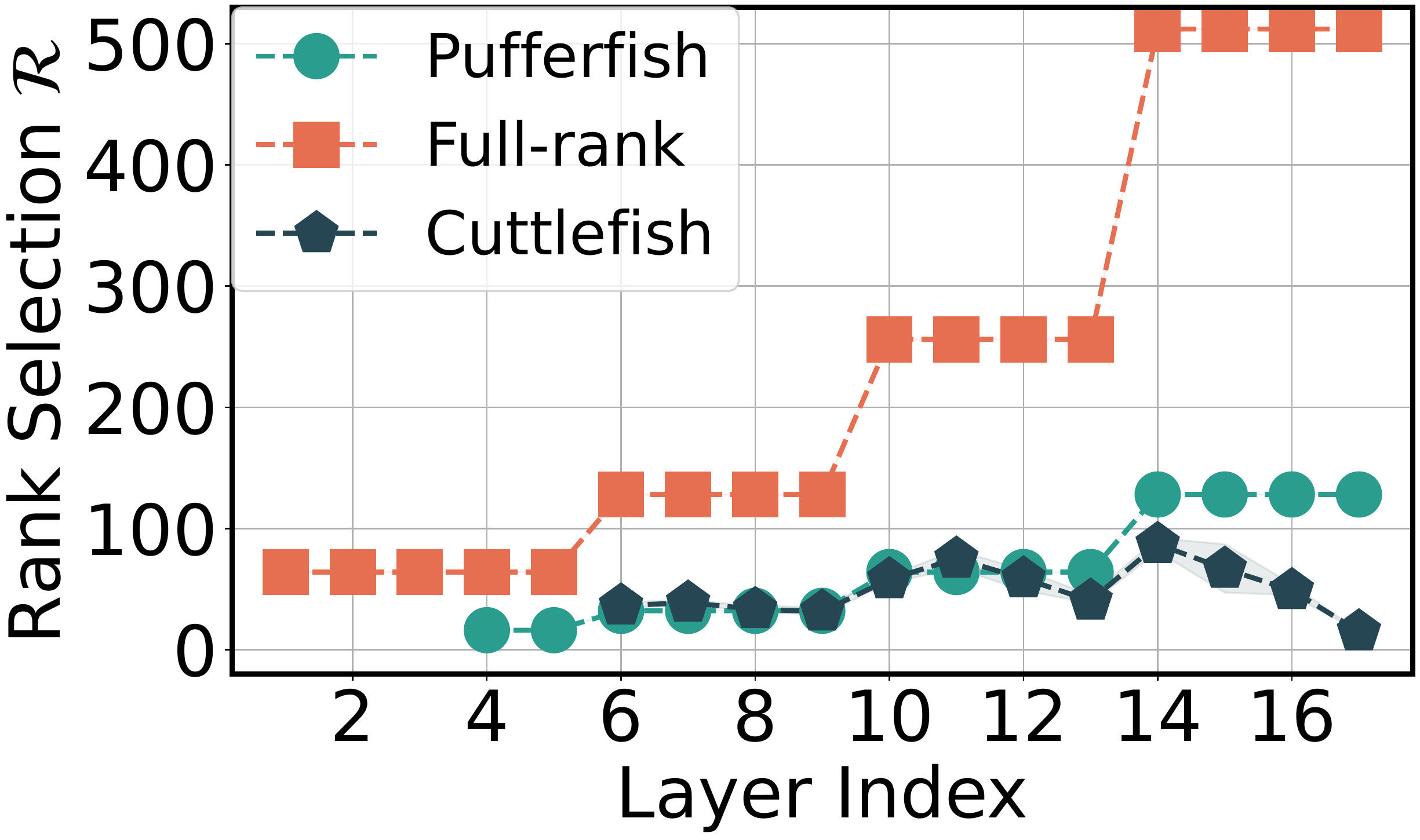}}
    \subfigure[ResNet-18 on CIFAR-100]{\includegraphics[width=0.315\textwidth]{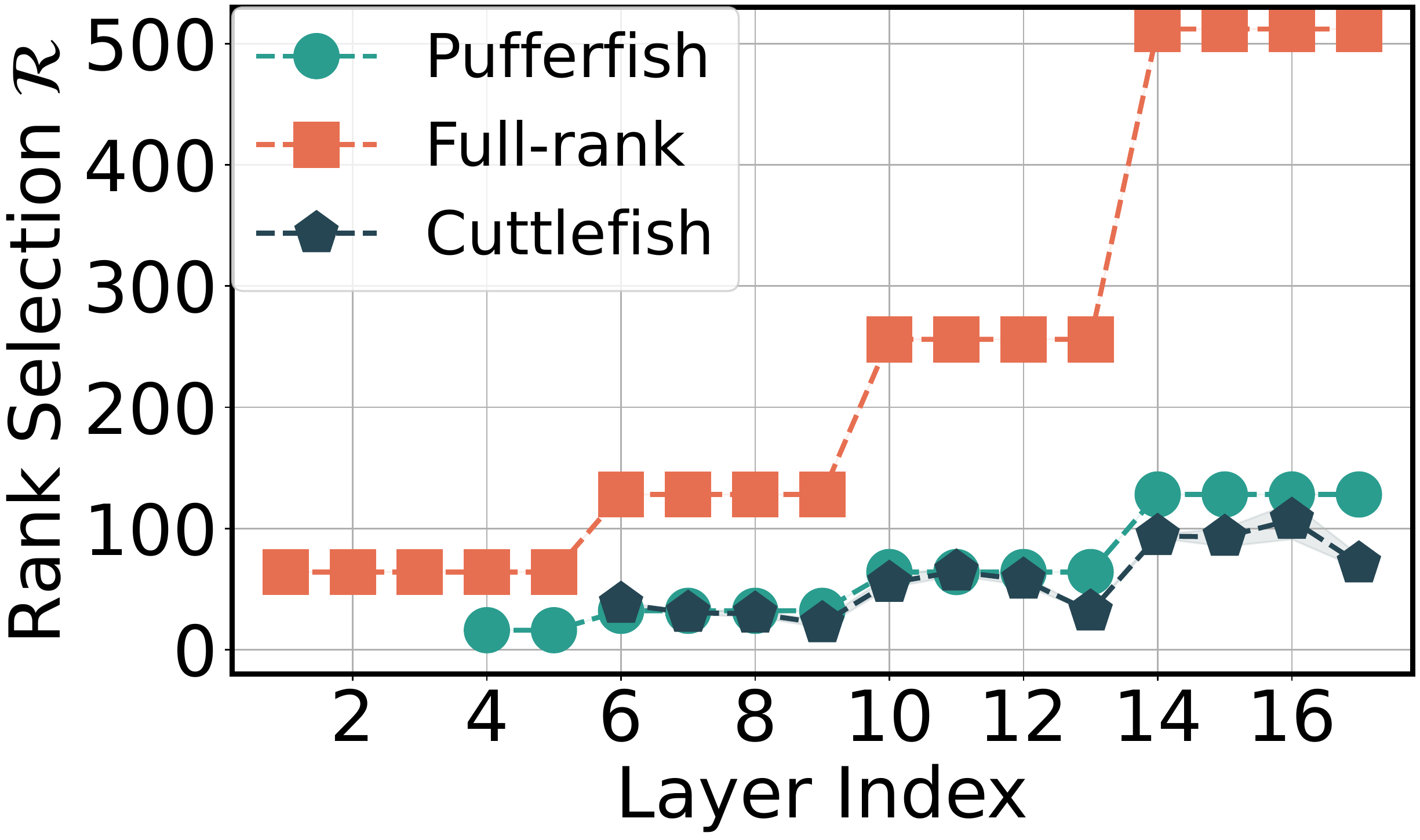}}
    \subfigure[ResNet-18 on SVHN]{\includegraphics[width=0.315\textwidth]{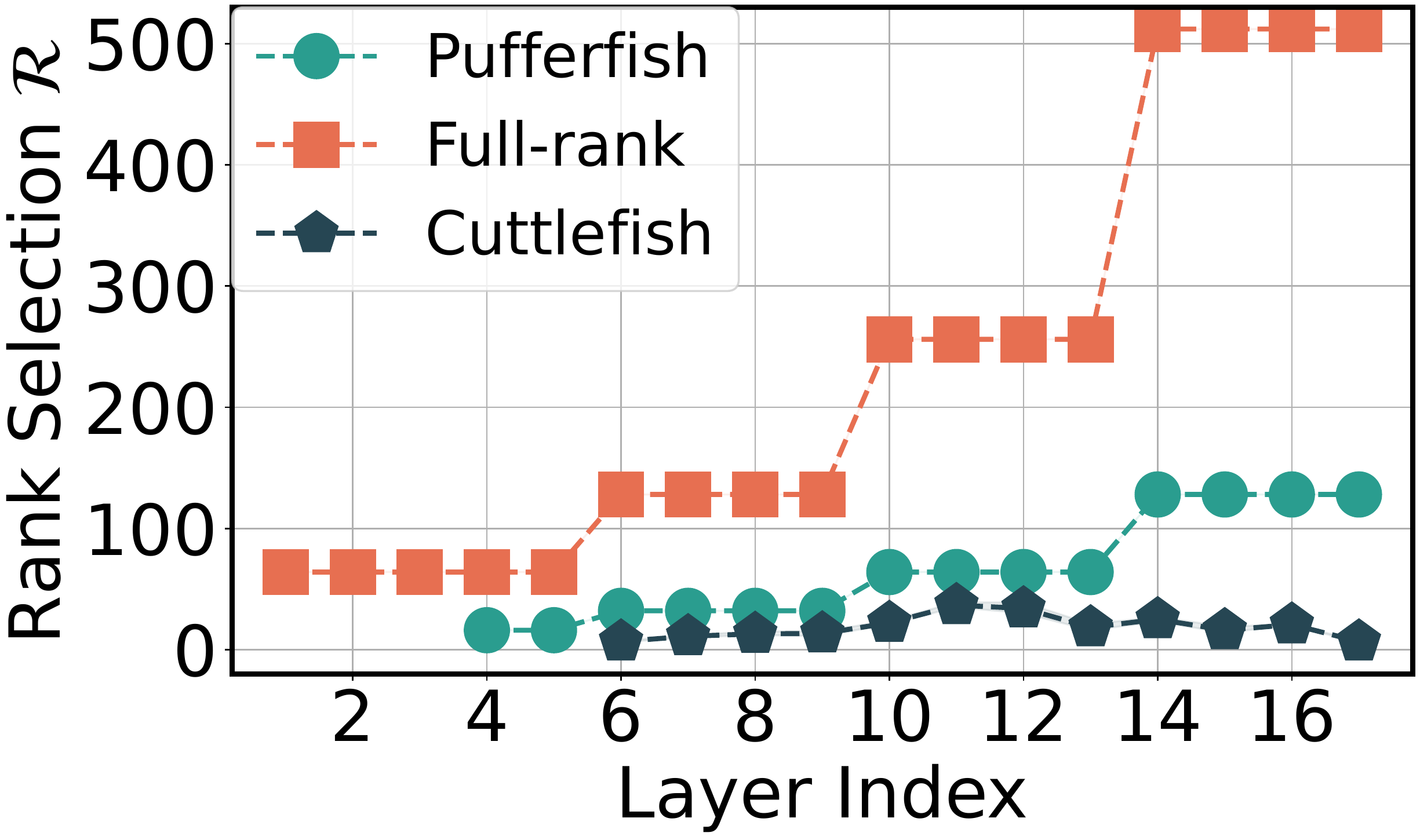}}\\
    \vspace{-2mm}
    \subfigure[VGG-19 on CIFAR-10]{\includegraphics[width=0.315\textwidth]{figures/rank_comparisons_cifar10_vgg19.pdf}}
    \subfigure[VGG-19 on CIFAR-100]{\includegraphics[width=0.315\textwidth]{figures/rank_comparisons_cifar100_vgg19.pdf}}
    \subfigure[VGG-19 on SVHN]{\includegraphics[width=0.315\textwidth]{figures/rank_comparisons_svhn_vgg19.pdf}}
    \vspace{-2mm}
    \caption{The ranks ($\mathcal{R}$) determined by \cuttlefish{}, \pufferfish{}, and LC compression (available only for VGG-19 experiments) for various layers in ResNet-18 ((a), (b), (c)) and VGG-19 ((d), (e), (f)) were trained on CIFAR-10 using a batch size of 1,024.}
    \label{fig:rank-plot-cifar-svhn}
    \vspace{-1mm}
\end{figure*}

\paragraph{ImageNet (ILSVRC 2012) on CNNs.} For ResNet-50 and WideResNet-50-2 trained on ImageNet, we report the details of the hyper-parameters $\hat s \in \mathcal{S}$ found by \cuttlefish{} and the manually tuned hyper-parameters by \pufferfish{} in Table~\ref{table:appendix-hyper-params-imagenet}. We can observe that for both ResNet-50 and WideResNet-50-2, \cuttlefish{} identifies the same $K$ and longer $E$ compared to \pufferfish{}. The ranks ($\mathcal{R}$s) chosen by \cuttlefish{} and \pufferfish{} can be found in Figure~\ref{fig:rank-plot-imagenet-cnns}, where it is evident that \cuttlefish{} employs lower ranks to factorize layers in ResNet-50 and WideResNet50-2 while using longer full-rank training epochs.

\begin{table*}[ht]
\caption{The hyperparameters $\hat s \in \mathcal{S}$ obtained by \cuttlefish{}, as well as the manually tuned $s$ from \pufferfish{}, for ResNet-50 and WideResNet-50-2 trained on ImageNet using a batch size of $256$.}
	\label{table:appendix-hyper-params-imagenet}
	\begin{center}
      \small{
		\begin{tabular}{ccc}
		\toprule
		& \multicolumn{2}{c}{ImageNet} \bigstrut\\
		\rowcolor{Gray} \midrule \textbf{Model:} ResNet-50
		 & $E$ & $K$ 
		\bigstrut\\
		\midrule
		\cuttlefish{} & $19.3_{\pm 0.5}$ & $40$ \bigstrut\\
		\rowcolor{LightCyan} \pufferfish{} & $10$ & $40$ \bigstrut\\
		\rowcolor{Gray} \midrule \textbf{Model:} WideResNet-50-2
		 & $E$ & $K$ 
		\bigstrut\\
		\midrule
		\cuttlefish{} & $21.3_{\pm 0.5}$ & $40$  \bigstrut\\
		\rowcolor{LightCyan} \pufferfish{} & $10$ & $40$ \bigstrut\\
		\bottomrule
		\end{tabular}}%
	\end{center}
 \vspace{-2mm}
\end{table*}

\begin{figure*}[ht]
    \centering
    \subfigure[ResNet-50 on ImageNet]{\includegraphics[width=0.35\textwidth]{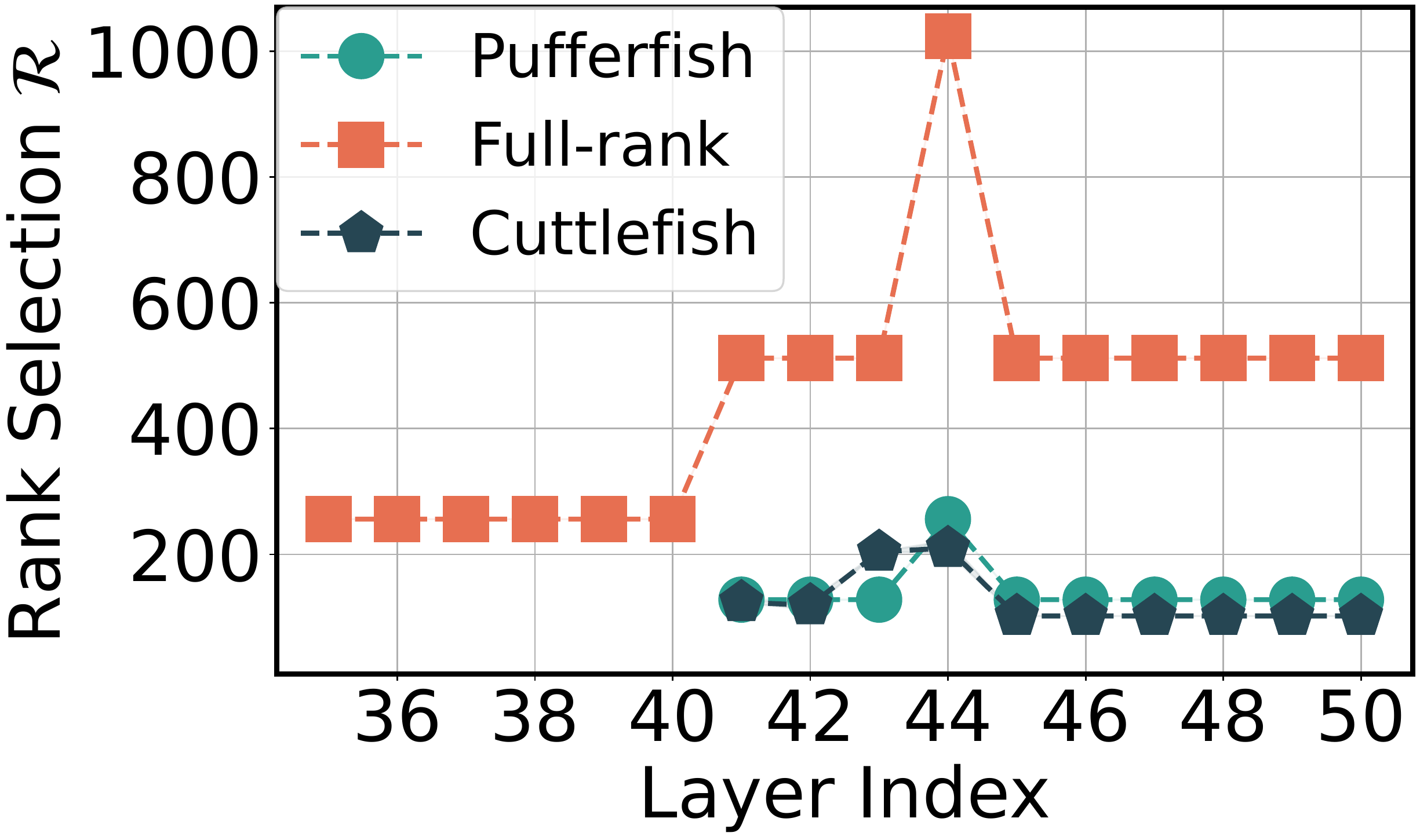}}
    \subfigure[WideResNet-50-20 on ImageNet]{\includegraphics[width=0.35\textwidth]{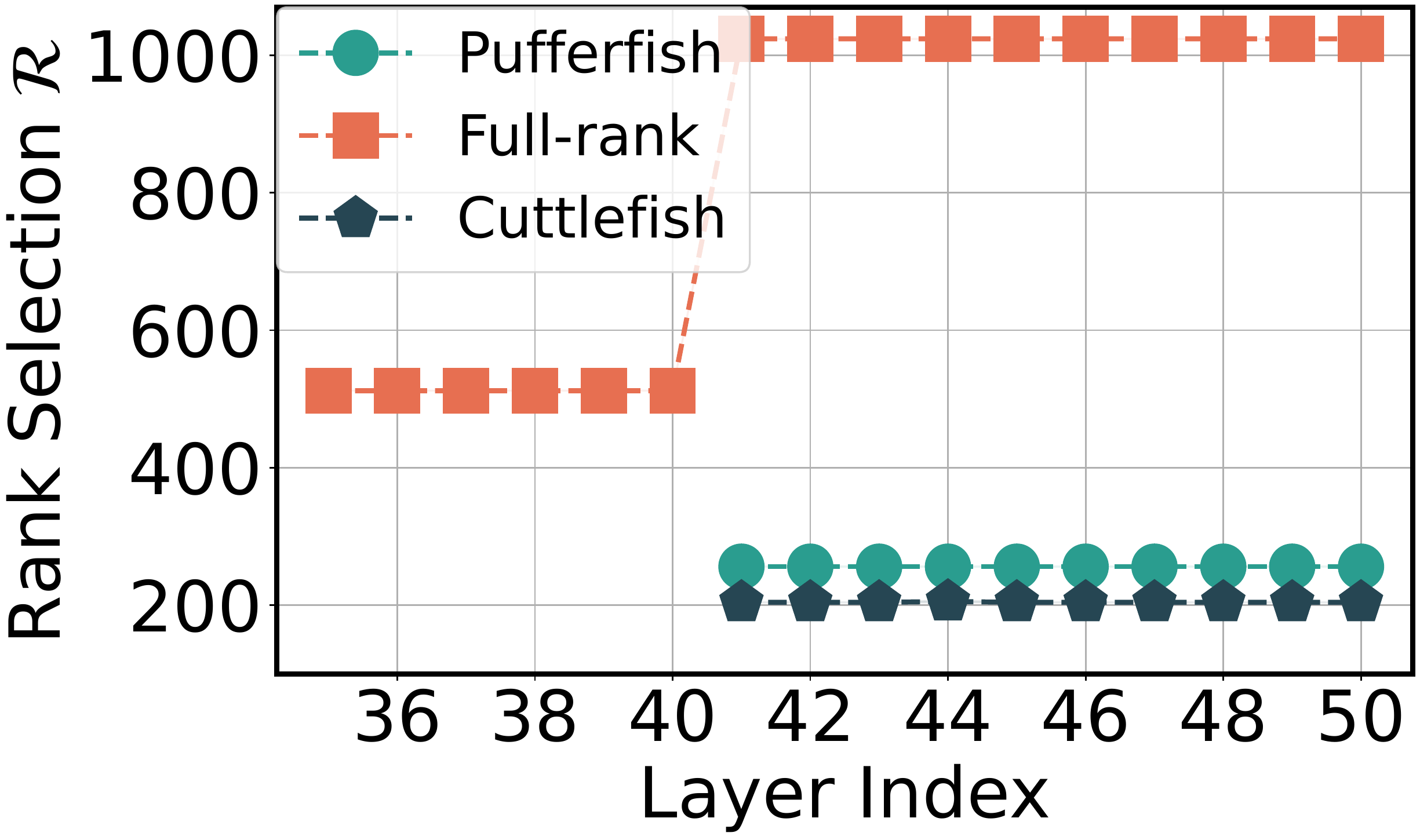}}
    \vspace{-2mm}
    \caption{The ranks $\mathcal{R}$s obtained by \cuttlefish{} and \pufferfish{} methods for different layers in ResNet-50 (a) and WideResNet-50-2 (b) trained on ImageNet with a batch size of $256$.}
    \vspace{-2mm}
    \label{fig:rank-plot-imagenet-cnns}
    \vspace{-1mm}
\end{figure*}

\paragraph{ImageNet (ILSVRC 2012) on DeiT and ResMLP.} For DeiT-base and ResMLP-S36 trained on ImageNet, we report the details of the hyperparameters $\hat s \in \mathcal{S}$ found by \cuttlefish{} and the manually tuned hyperparameters by \pufferfish{} in Table~\ref{table:appendix-hyper-params-imagenet-vit-resmlp}. We can observe that for DeiT-base and ResMLP-S36, \cuttlefish{} identifies the same $K$ and longer $E$ compared to \pufferfish{}. When selecting the ranks $\mathcal{R}$s, we found that even with scaled stable rank, the factorized low-rank models still result in a significant final accuracy drop. To understand why this occurs, we plot the cumulative distribution function (CDF) of the singular values of the Transformer encoder layers in the DeiT-base model at the full-rank to low-rank switching epoch, \ie $\hat E$ (the results are shown in Figure~\ref{fig:rank-cdf-curves-imagenet-deit}). If the curves are closer to the reference line, it indicates that the model weights are more like full-rank, \ie contain less redundancy.

From Figure~\ref{fig:rank-cdf-curves-imagenet-deit}, we can see that to approximate the major information of the weight matrix, \eg preserving $80\%$ of the singular value information, relatively higher $\mathcal{R}$s should be used, \eg $\rho = \frac{1}{2}$. Additionally, the attention weights, such as $\mathbf{W}_q$, $\mathbf{W}_k$, $\mathbf{W}_v$, as well as the projection layer after the query, key, and value layers (\ie $\mathbf{W}^o$ in our notation) tend to have lower ranks (higher redundancy) compared to the FFN layers, \ie FC1 and FC2 in Figure~\ref{fig:rank-cdf-curves-imagenet-deit}. In \cuttlefish{}, for DeiT and ResMLP models, we use a global rank ratio $\rho=\frac{1}{2}$ for all factorized layers. It is worth noting that the linear projection layer after each $\mathbf{W}_q$, $\mathbf{W}_k$, $\mathbf{W}_v$ has dimensions of $(768, 768)$. Using $\rho$ for this layer, the factorized $\mathbf{U}, \mathbf{V}^\top$ layers will have dimensions of $(768, 384), (384, 768)$, which will not result in any model size reduction or computational complexity savings. Thus, we opt not to factorize the linear projection layers in each multi-head attention layer in DeiT-base. For ResMLP-S36, we factorize all layers except for the embedding layers with a fixed global rank ratio of $\rho=\frac{1}{2}$.

A concern arises from the fact that our proposed stable rank selection heuristic for choosing $\mathcal{R}$ may not generally apply to both CNN and Transformer models, as Transformer model weights tend to have higher ranks. To address this issue, future work can adjust \cuttlefish{}'s rank selection heuristic to:
{\small
\begin{align*}
 \label{eq:adjusted-rank-estimation-metric}
    \max\{&\texttt{scaled stable rank}(\Sigma),\\
         &\texttt{accumulative rank}(\Sigma, p)\}   
\end{align*}}
Here, $\texttt{accumulative rank}(\Sigma, p)$ measures the smallest rank value $r$ such that (where $\sigma$s represent the singular values of a model weight matrix $\mathbf{W}$, and are also the elements on the diagonal of matrix $\Sigma$):
{\small
\begin{align*}
    \sum^{r}_{i=1}\sigma_i \geq  p\cdot \sum^{\text{rank}(\mathbf{W})}_{j=1} \sigma_j.
\end{align*}}
In the DeiT example mentioned earlier, we know that the $\texttt{accumulative rank}(\Sigma, 80\%)$ for most model layers is generally greater than $\frac{1}{2}\times \text{rank}(\mathbf{W})$ and the scaled stable rank for these layers is generally lower than those values. Consequently, the new metric defined above will consistently return $\frac{1}{2}\times \text{rank}(\mathbf{W})$ for all factorized layers in the DeiT-base model. Another hyperparameter tuning we performed in our experiments is decaying the base learning rate by a certain fraction after switching from full-rank to low-rank training at epoch $\hat E$. For \cuttlefish{} DeiT-base, we decay the base learning rate by $\frac{1}{3}$. For \cuttlefish{} ResMLP-S36, we decay the base learning rate by $\frac{1}{2}$.

\cuttlefish{} begins factorizing layers after the first embedding layer, which is simply a convolution layer, \ie $K=1$ for both DeiT-base and ResMLP-S36. Since we tune $K$ for \pufferfish{} such that the end model sizes match \cuttlefish{} DeiT and ResMLP, \pufferfish{} starts factorizing layers from the 7th encoder block for DeiT, \ie $K=19$ (6 blocks, 3 layers in each block), and the 18th ResMLP block, \ie $K=52$ (17 blocks, 3 layers in each block).

\begin{table*}[ht]
	\caption{The tuned hyperparameters $\hat s \in \mathcal{S}$, as determined by \cuttlefish{}, and the manually tuned $s$ from \pufferfish{} for DeiT-base and ResMLP-S36, trained on the ImageNet dataset using a batch size of 256.}
	\label{table:appendix-hyper-params-imagenet-vit-resmlp}
	\begin{center}
      \footnotesize{
		\begin{tabular}{ccc}
		\toprule
		& \multicolumn{2}{c}{ImageNet} \bigstrut\\
		\rowcolor{Gray} \midrule \textbf{Model:} DeiT-base
		 & $E$ & $K$ 
		\bigstrut\\
		\midrule
		\cuttlefish{} & $59.5_{\pm 6.5}$ & $1$ \bigstrut\\
		\rowcolor{LightCyan} \pufferfish{} & $80$ & $19$ \bigstrut\\
		\rowcolor{Gray} \midrule \textbf{Model:} ResMLP-S36
		 & $E$ & $K$ 
		\bigstrut\\
		\midrule
		\cuttlefish{} & $40.5_{\pm 1.5}$ & $1$  \bigstrut\\
		\rowcolor{LightCyan} \pufferfish{} & $80$ & $52$ \bigstrut\\
		\bottomrule
		\end{tabular}}%
	\end{center}
\end{table*}

\begin{figure*}[ht]
    \vspace{-2mm}
    \centering
    \subfigure[Encoder 0 of DeiT-base]{\includegraphics[width=0.35\textwidth]{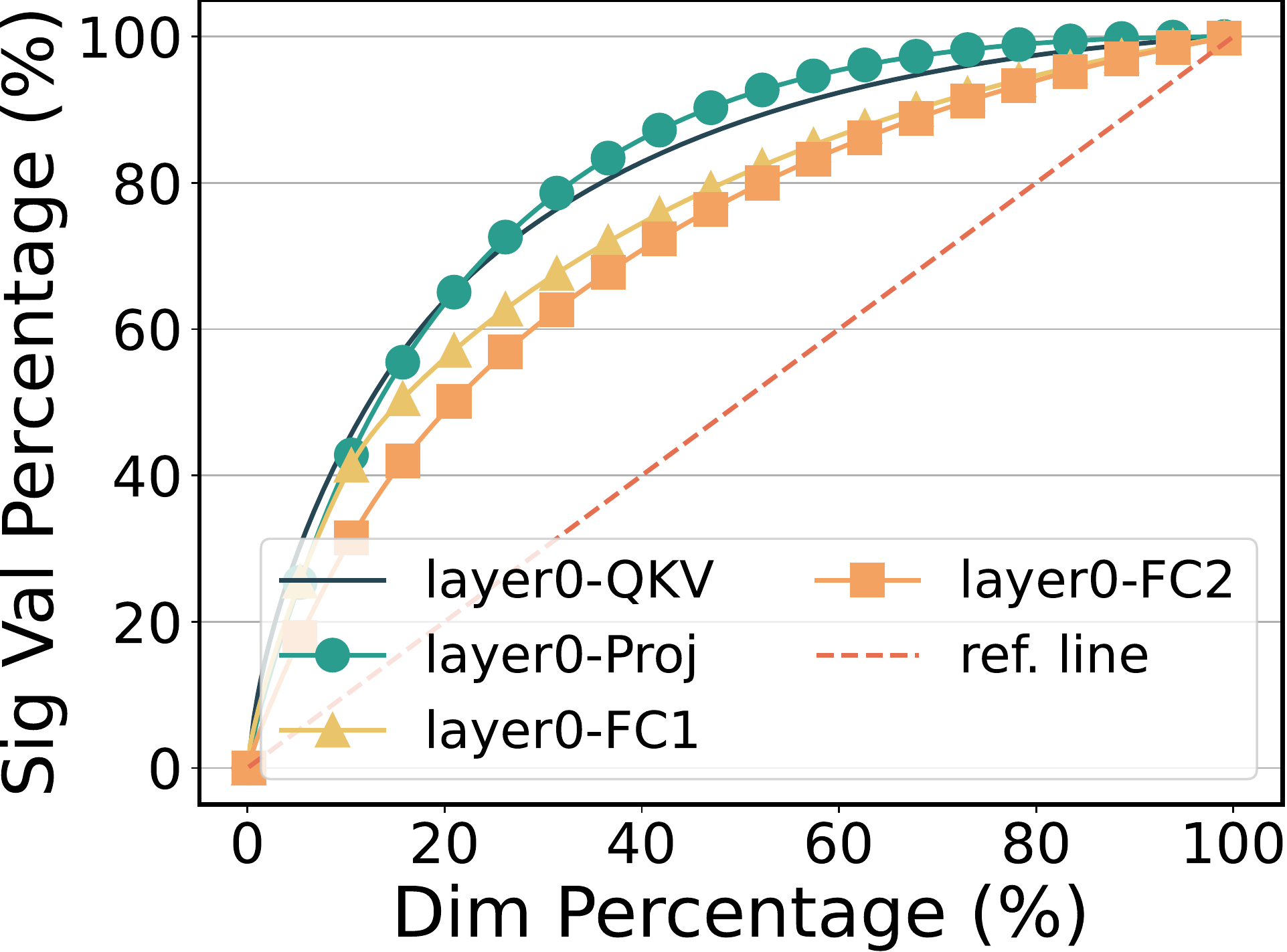}}
    \subfigure[Encoder 11 of DeiT-base]{\includegraphics[width=0.35\textwidth]{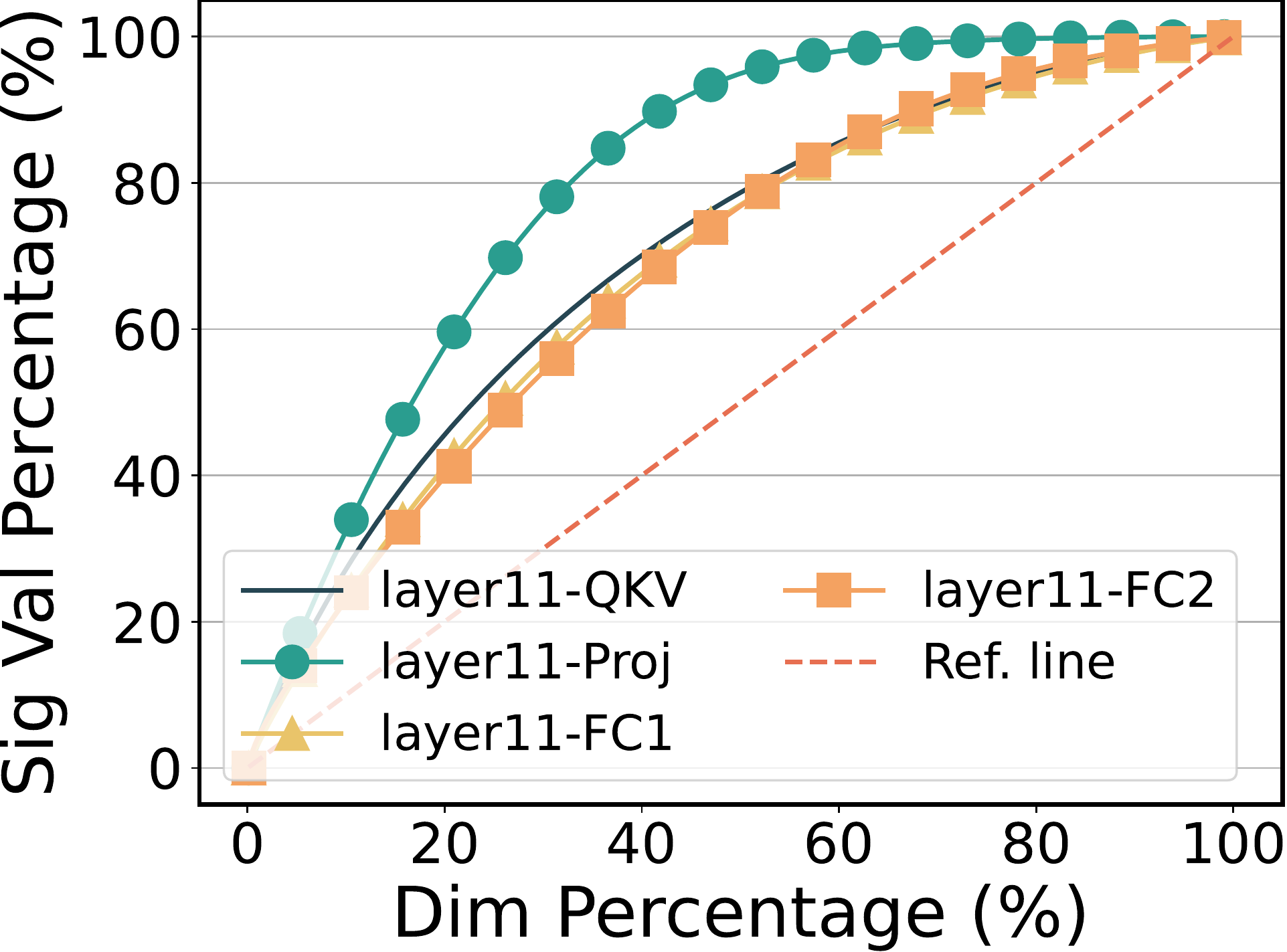}}
    \vspace{-2mm}
    \caption{The Cumulative Distribution Function (CDF) of singular values for the first Transformer encoder (\ie Encoder 0, denoted as layer0 in the figure) (a) and the last Transformer encoder (\ie Encoder 11, denoted as layer11 in the figure) (b) of DeiT-base trained on the ImageNet dataset using a batch size of 256. Other Transformer encoders exhibit similar trends.}
    \label{fig:rank-cdf-curves-imagenet-deit}
    \vspace{-2mm}
\end{figure*}

\paragraph{GLUE Benchmark on $\text{BERT}_{\text{BASE}}$.} For $\text{BERT}{\text{BASE}}$ fine-tuning on the GLUE benchmark, the fine-tuning epochs for all downstream tasks are typically small, such as $T=3, 5$. Therefore, we set $E=1$ for \cuttlefish{}. Additionally, we free the fully connected layers following each multi-head attention layer when fine-tuning the factorized $\text{BERT}_{\text{BASE}}$. As in LoRA, during the fine-tuning stage, we do not update the feed-forward network (FFN) in BERT at all; we freeze the FC1 and FC2 layers contained in the FFN~\cite{hu2021lora}. We perform learning rate sweeping over the learning rates $\gamma$s for all methods during GLUE fine-tuning. The tuned learning rates are shown in Table~\ref{table:tuned-lrs-lm-glue}. For the relatively challenging CoLA task, we fine-tune all models except for the vanilla $\text{BERT}{\text{BASE}}$ for 5 epochs instead of 3 epochs. For the RTE task, we fine-tune Distill BERT for 5 epochs.

\begin{table*}[ht]
\caption{Tuned learning rates, denoted as $\gamma$s, for vanilla $\text{BERT}_{\text{BASE}}$, Distill $\text{BERT}$, Tiny $\text{BERT}6$, and \cuttlefish{} $\text{BERT}{\text{BASE}}$ on the GLUE benchmark.}
	\label{table:tuned-lrs-lm-glue}
	\begin{center}
		 \scriptsize{
		\begin{tabular}{cccccccccc}
		\rowcolor{Gray} \toprule  
		 \textbf{Model} & \# Params. ($M$) & MNLI &  QNLI & QQP & RTE & SST-2 & MRPC & CoLA & STS-B
		\bigstrut\\
		\midrule
		\rowcolor{LightCyan} $\text{BERT}_{\text{BASE}}$ & $108.3$& 2e-5 & 2e-5 & 2e-5 & 4e-5 & 2e-5 & 2e-5 & 4e-5 & 2e-5 \bigstrut\\
        Distill $\text{BERT}$ & $65.8$& 2e-5 & 2e-5 & 2e-5 & 2e-5 & 4e-5 & 2e-5 & 2e-5 & 2e-5  \bigstrut\\
        \rowcolor{LightCyan} Tiny $\text{BERT}_6$ & $67.0$& 2e-5 & 2e-5 & 2e-5 & 2e-5 & 2e-5 & 2e-5 & 2e-5 & 2e-5  \bigstrut\\
		\cuttlefish{} &${\bf 48.8}$& 2e-5 & 2e-5 & 2e-5 & 3e-5 & 3e-5 & 2e-5 & 3e-5 & 2e-5 \bigstrut\\
		\bottomrule
		\end{tabular}}%
	\end{center}
\end{table*}

\vspace{2mm}
\subsection{Hyperparameters for other baselines.}
\vspace{2mm}
\paragraph{SI\&FD.} We adjust the fixed global rank ratios, denoted as $\rho$s, for SI\&FD so that the resulting model sizes align with the factorized low-rank models produced by \cuttlefish{}. Detailed information on the $\rho$s used in our experiments can be found in Table~\ref{table:appendix-si-fd-hyper-params}.

\begin{table*}[ht]
	\caption{The fixed rank ratios ($\rho$s) employed in SI\&FD experiments for CIFAR-10, CIFAR-100, and SVHN on ResNet-18 and VGG-19.}
        \vspace{1mm}
	\label{table:appendix-si-fd-hyper-params}
	\begin{center}
      \scriptsize{
		\begin{tabular}{cccc}
		\toprule
		\rowcolor{Gray} \textbf{Model:} ResNet-18 & CIFAR-10 & CIFAR-100 & SVHN\bigstrut\\
		\midrule
		SI\&FD & 0.08 & 0.105  & 0.032 \bigstrut\\
        \rowcolor{Gray} \midrule \textbf{Model:} VGG-19 & CIFAR-10 & CIFAR-100 & SVHN\bigstrut\\
		\midrule
		SI\&FD & 0.1 & 0.165  & 0.059 \bigstrut\\
		\bottomrule
		\end{tabular}}%
	\end{center}
\end{table*}

\paragraph{LC compression.} The implementation and hyperparameter configurations for our experiments are taken directly from the original GitHub repository\footnote{\url{https://github.com/UCMerced-ML/LC-model-compression}} associated with~\cite{idelbayev2020low}. We modified the VGG-19 model implementation in the LC compression setup to ensure consistency with the VGG-19 architecture used in our experiments.

\section{Additional experimental results}\label{sec:add_exp}
\subsection{Ablation study.} 
\paragraph{Combining \cuttlefish{} with Frobenius decay.} In this section, we present the results of an ablation study examining the combination of {\it Frobenius decay} (FD) with \cuttlefish{} across various machine learning tasks. The results can be found in Table~\ref{table:appendix-ablation-fd} and Table~\ref{table:appendix-ablation-fd-imagenet}. It is evident that applying FD to \cuttlefish{} does not consistently lead to better model accuracy. For instance, combining \cuttlefish{} with FD yields a 1.8\% higher accuracy for ResNet-18 training on CIFAR-100. However, for other tasks, incorporating FD with ResNet-18 either results in worse final model accuracy or only marginal accuracy improvements. This observation aligns with the findings of~\cite{vodrahalli2022algorithms}, indicating that FD does not always enhance the accuracy of factorized low-rank models.
\begin{table*}[ht]
    \vspace{-2mm}
    \caption{The ablation study results, averaged across three independent trials with different random seeds, showcase the performance of \cuttlefish{} combined with Frobenius decay (FD) on ResNet-18 and VGG-19 trained on CIFAR-10 using a batch size of 1,024.}
	\label{table:appendix-ablation-fd}
	\begin{center}
      \scriptsize{
		\begin{tabular}{ccccccc}
		\toprule
		& \multicolumn{2}{c}{CIFAR-10} & \multicolumn{2}{c}{CIFAR-100} & \multicolumn{2}{c}{SVHN}\bigstrut\\
		\rowcolor{Gray} \midrule \textbf{Model:} 
		 & \# Params. & Val. Acc. & \# Params. & Val. Acc. & \# Params. & Val. Acc. 
		\\
		\rowcolor{Gray} ResNet-18 & ($M$) & ($\%$)  & ($M$) & ($\%$)  & ($M$) & ($\%$) 
		\bigstrut\\
		\midrule
		\cuttlefish{} wo. FD & $2.0$ & $94.52_{\pm 0.01}$  & $2.6$ & $73.75_{\pm 0.24}$  & $0.96$ & ${\bf 96.47}_{\pm 0.02}$\bigstrut\\
		\rowcolor{LightCyan} \cuttlefish{} w. FD & $2.0$ & ${\bf 94.62}_{\pm 0.09}$ & $2.6$ & ${\bf 75.54}_{\pm 0.18}$ & $0.94$ & $96.34_{\pm 0.08}$ \bigstrut\\
		\rowcolor{Gray} \midrule \textbf{Model:} 
		& \# Params. & Val. Acc. & \# Params. & Val. Acc. &  \# Params. & Val. Acc. 
		\\
		\rowcolor{Gray} VGG-19 
		& ($M$) & ($\%$)  & ($M$) & ($\%$)  & ($M$) & ($\%$) 
		\bigstrut\\
		\midrule
		\cuttlefish{} wo. FD & $1.9$ & ${\bf 93.49}_{\pm 0.18}$ & $3.3$ & ${\bf 72.27}_{\pm 0.25}$ & $1.2$ & ${\bf 96.33}_{\pm 0.04}$ \bigstrut\\
		\rowcolor{LightCyan} \cuttlefish{} w. FD & $1.9$ & $93.42_{\pm 0.25}$ & $3.3$ & $72.15_{\pm 0.27}$ & $1.2$ & ${\bf 96.33}_{\pm 0.02}$ \bigstrut\\
		\bottomrule
		\end{tabular}}%
	\end{center}
\end{table*}

\begin{table*}[ht]
    \vspace{-2mm}
    \caption{The ablation study results, averaged across three independent trials, demonstrate the performance of \cuttlefish{} combined with Frobenius decay (FD) on ResNet-50 trained on ImageNet using a batch size of 256.}
	\label{table:appendix-ablation-fd-imagenet}
	\begin{center}
      \scriptsize{
		\begin{tabular}{cccc}
		\toprule
		\rowcolor{Gray} \textbf{Model:} ResNet-50
		 & \# Params. ($M$) & Top-1 Val. Acc. ($\%$) & Top-5 Val. Acc. ($\%$)
		\bigstrut\\
		\midrule
		\cuttlefish{} wo. FD & $14.7$ & $76.16_{\pm 0.04}$ & $92.97_{\pm 0.06}$  \bigstrut\\
		\rowcolor{LightCyan} \cuttlefish{} w. FD & $14.7$ & ${\bf 76.44}_{\pm 0.16}$ & ${\bf 93.21}_{\pm 0.03}$ \bigstrut\\
		\bottomrule
		\end{tabular}}%
	\end{center}
\end{table*}

\paragraph{The impact of scaled stable rank.} As mentioned in the main paper, the use of {\it stable rank} can lead to overly aggressive low rank estimations, potentially harming the final accuracy of factorized low-rank models. To address this issue, \cuttlefish{} employs scaled stable rank. The ablation study results are presented in Table~\ref{table:appendix-ablation-scaled-stable-rank}. We find that for CIFAR-10 and CIFAR-100 datasets, utilizing scaled stable rank is crucial for achieving satisfactory final accuracy in factorized low-rank networks. For SVHN, which is a comparatively simpler task, even the vanilla stable rank is sufficient to attain good accuracy. Thus, in our main paper's reported experiment, we use vanilla stable rank for SVHN and scaled stable rank for CIFAR-10 and CIFAR-100. For larger scale tasks on ImageNet, it is evident that adopting scaled stable rank is essential; otherwise, the model accuracy will suffer a significant drop, as shown in Table~\ref{table:imagenet-scaled-stable-rank-ablation-study}.

\begin{table*}[ht]
    \vspace{-2mm}
    \caption{The ablation study results (averaged over $3$ independent trials with varying random seeds) for \cuttlefish{} using both scaled stable rank and vanilla stable rank on ResNet-18 and VGG-19, trained on CIFAR-10, CIFAR-100, and SVHN with a batch size of $1,024$.}
	\label{table:appendix-ablation-scaled-stable-rank}
	\begin{center}
      \scriptsize{
		\begin{tabular}{ccccccc}
		\toprule
		& \multicolumn{2}{c}{CIFAR-10} & \multicolumn{2}{c}{CIFAR-100} & \multicolumn{2}{c}{SVHN}\bigstrut\\
		\rowcolor{Gray} \midrule \textbf{Model:} 
		 & \# Params. & Val. Acc. & \# Params. & Val. Acc. & \# Params. & Val. Acc. 
		\\
		\rowcolor{Gray} ResNet-18 & ($M$) & ($\%$)  & ($M$) & ($\%$)  & ($M$) & ($\%$) 
		\bigstrut\\
		\midrule
		\cuttlefish{} vanilla stable rank & $1.2$ & $94.27_{\pm 0.05}$  & $1.6$ & $74.21_{\pm 0.20}$  & $0.94$ & $96.47_{\pm 0.02}$\bigstrut\\
		\rowcolor{LightCyan} \cuttlefish{} scaled stable rank & $2.0$ & ${\bf 94.62}_{\pm 0.09}$ & $2.6$ & ${\bf 75.54}_{\pm 0.18}$ & $1.4$ & ${\bf 96.53}_{\pm 0.08}$ \bigstrut\\
		\rowcolor{Gray} \midrule \textbf{Model:} 
		& \# Params. & Val. Acc. & \# Params. & Val. Acc. &  \# Params. & Val. Acc. 
		\\
		\rowcolor{Gray} VGG-19 
		& ($M$) & ($\%$)  & ($M$) & ($\%$)  & ($M$) & ($\%$) 
		\bigstrut\\
		\midrule
		\cuttlefish{} vanilla stable rank & $1.1$ & $93.07_{\pm 0.10}$ & $1.9$ & $70.62_{\pm 0.23}$ & $1.2$ & $96.33_{\pm 0.04}$ \bigstrut\\
		\rowcolor{LightCyan} \cuttlefish{} scaled stable rank & $1.9$ & ${\bf 93.49}_{\pm 0.18}$ & $3.3$ & ${\bf 72.27}_{\pm 0.25}$ & $2.0$ & ${\bf 96.42}_{\pm 0.07}$ \bigstrut\\
		\bottomrule
		\end{tabular}}%
	\end{center}
\end{table*}

\begin{table*}[ht]
    \vspace{-2mm}
    \caption{The ablation study results (averaged over $3$ independent trials with distinct random seeds) for \cuttlefish{} using both scaled stable rank and vanilla stable rank on DeiT-base, ResNet-50, and WideResNet-50, trained on ImageNet with a batch size of $256$.}
	\label{table:imagenet-scaled-stable-rank-ablation-study}
	\begin{center}
	  \scriptsize{
		\begin{tabular}{cccc}
		\toprule 
		\rowcolor{Gray} & \# Params. ($M$)  &  Top-1 Val. Acc. & Top-5 Val. Acc.  
		\bigstrut\\
		\midrule
		 \cuttlefish{} DeiT-base vanilla stable rank & $12.5$ & $64.80_{\pm 0.82}$ & $85.46_{\pm 0.60}$ \bigstrut\\
		\rowcolor{LightCyan} \cuttlefish{} DeiT-base scaled stable rank & $58.3$ & ${\bf 81.52}_{\pm 0.03}$ & ${\bf 95.59}_{\pm 0.04}$  \bigstrut\\
		\midrule
		 \cuttlefish{} WideResNet-50 vanilla stable rank & $29.1$ & $76.86_{\pm 0.01}$ & $93.50_{\pm 0.03}$  \bigstrut\\
		\rowcolor{LightCyan} \cuttlefish{} WideResNet-50 scaled stable rank & $37.4$ & ${\bf 78.0}_{\pm 0.06}$ & ${\bf 94.04}_{\pm 0.09}$  \bigstrut\\
		\midrule
		\cuttlefish{} ResNet-50 vanilla stable rank & $11.9$ & $74.96_{\pm 0.01}$ & $92.39_{\pm 0.07}$  \bigstrut\\
		\rowcolor{LightCyan} \cuttlefish{} ResNet-50 scaled stable rank & $14.7$ & ${\bf 76.44}_{\pm 0.16}$ & ${\bf  93.21}_{\pm 0.03}$ \bigstrut\\
		\bottomrule
		\end{tabular}}
	\end{center}
\vspace{-2mm}
\end{table*}

\subsection{Additional experimental results.} 
\paragraph{BERT pre-training using \cuttlefish{}.} We perform BERT pre-training on the Wikipedia and Bookcorpus datasets, adhering to the training schedule and codebase of the 24h BERT$_{\text{LARGE}}$ for faster training speed and due to limited computing resources~\cite{izsak2021train}. The results, shown in Table~\ref{table:bert-pretraining}, indicate that \cuttlefish{} enables pre-training a BERT model with only $72\%$ of the total model parameters while achieving the same final MLM loss.

\begin{table*}[ht]
\vspace{-3mm}
\caption{Vanilla and \cuttlefish{} BERT pre-training on Wikipedia and Bookcorpus datasets.}
    \vspace{2mm}
	\label{table:bert-pretraining}
	\begin{center}
		 \scriptsize{
		\begin{tabular}{ccc}
		\rowcolor{Gray} \toprule  
		 \textbf{Model} & \# Params. ($M$) & MLM Loss 
		\bigstrut\\
		\midrule
		\rowcolor{LightCyan} Vanilla $\text{BERT}_{\text{LARGE}}$ & $345$& $1.58$\bigstrut\\
        Cuttlefish $\text{BERT}_{\text{LARGE}}$ & $249$& $1.6$  \bigstrut\\
		\bottomrule
		\end{tabular}}%
	\end{center}
\vspace{-3mm}
\end{table*}

\paragraph{Rank varying trend of other datasets.} In our main paper, we presented the rank varying trend only for ResNet-18 on CIFAR-10. In this section, we expand our analysis and report additional experimental results on rank varying trends. Specifically, we provide results for VGG-19 trained on CIFAR-10, as well as VGG-19 and ResNet-18 on CIFAR-100 and SVHN datasets. For ResNet-50 on ImageNet, we also show the same results. Figures~\ref{fig:rank-est-resnet18-vgg19-cifar10} and \ref{fig:rank-ratio-resnet18-vgg19-cifar10} display the results for VGG-19 on CIFAR-10, while Figures~\ref{fig:rank-est-resnet18-vgg19-cifar100}, \ref{fig:rank-ratio-resnet18-vgg19-cifar100}, \ref{fig:rank-ratio-resnet18-vgg19-svhn}, \ref{fig:rank-est-resnet50-imagenet}, and \ref{fig:rank-ratio-resnet50-imagenet} illustrate the results for the remaining datasets.

Overall, our observations indicate that the stable rank of the network layers fluctuates significantly in the early stages of training but eventually converges to a constant value. This trend holds across all the datasets we analyzed. 
\begin{figure*}[ht]
    \centering
    \vspace{-2mm}
    \includegraphics[width=0.315\textwidth]{figures/epochs_vs_rank_ratio_resnet18_cifar10_0.pdf}
    \includegraphics[width=0.315\textwidth]{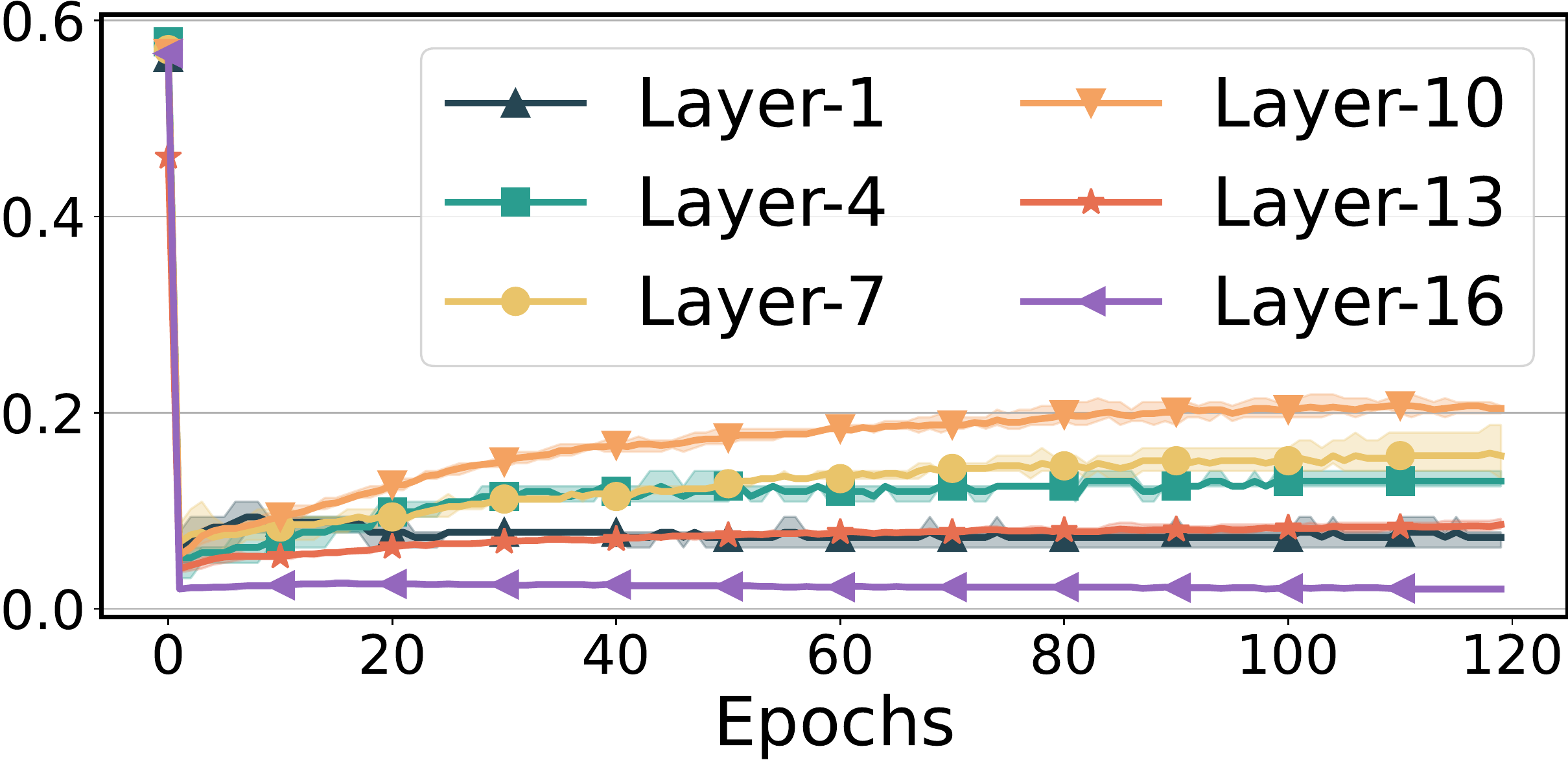}
    \includegraphics[width=0.315\textwidth]{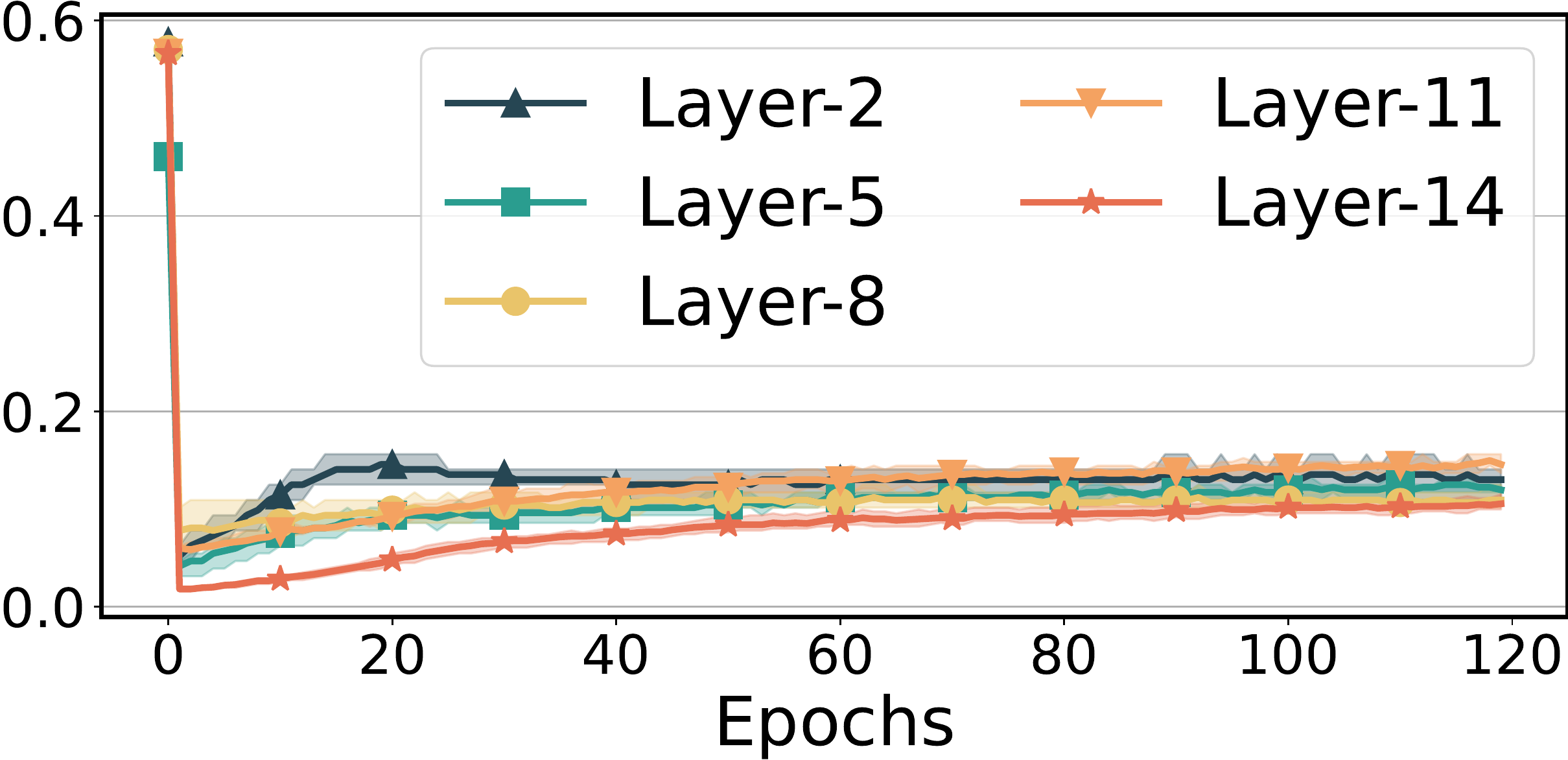}\\
    \includegraphics[width=0.315\textwidth]{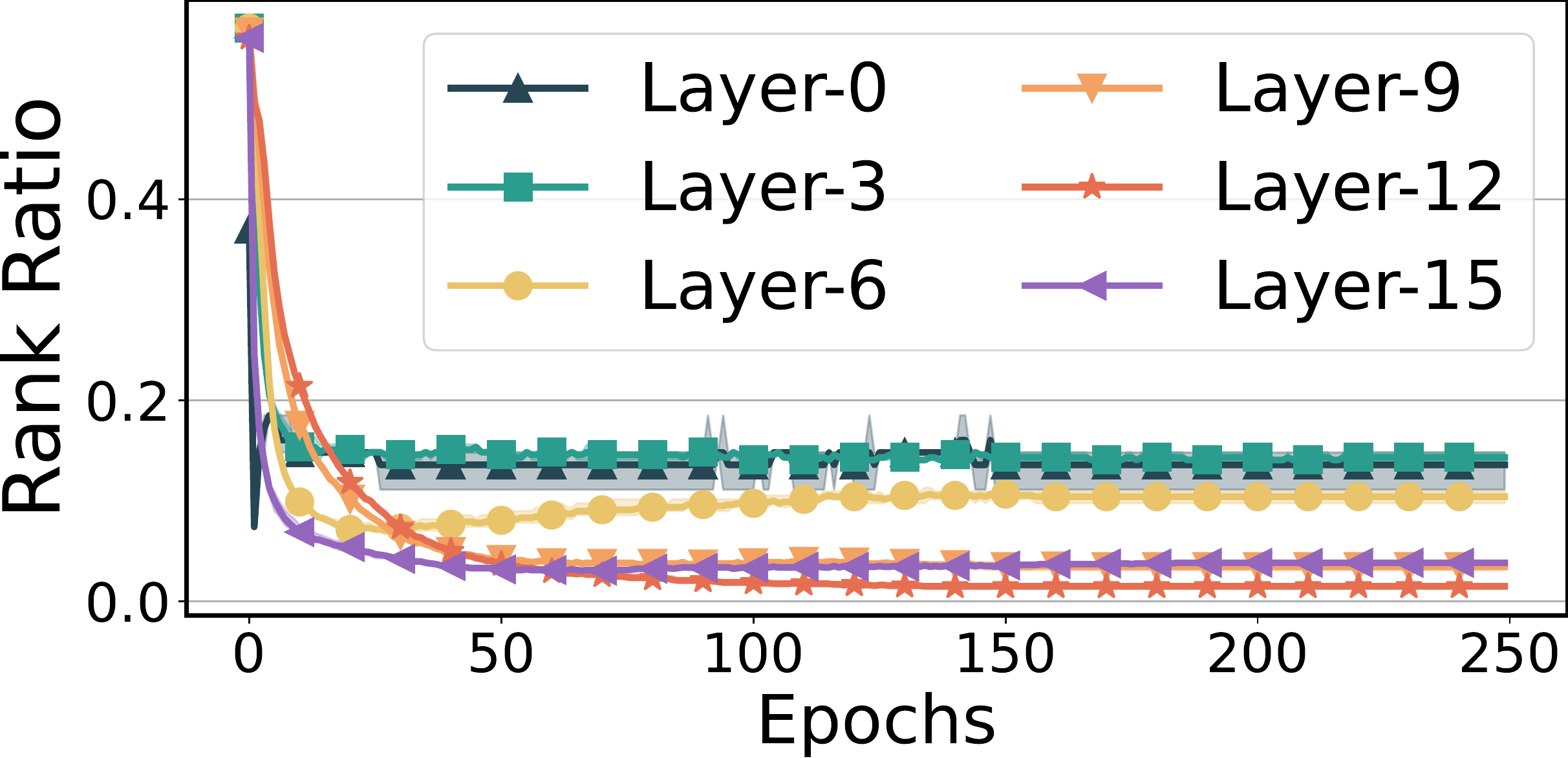}
    \includegraphics[width=0.315\textwidth]{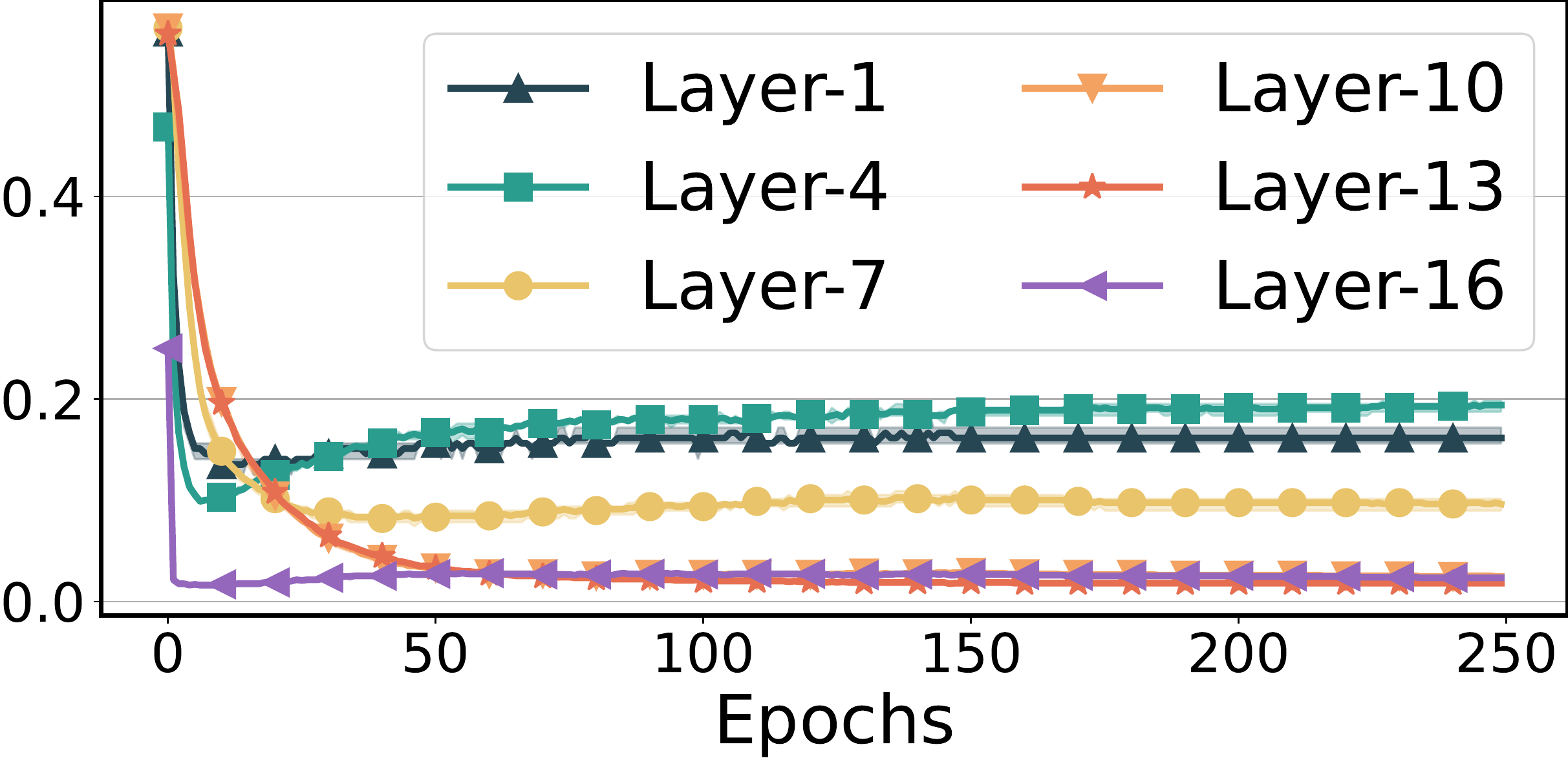}
    \includegraphics[width=0.315\textwidth]{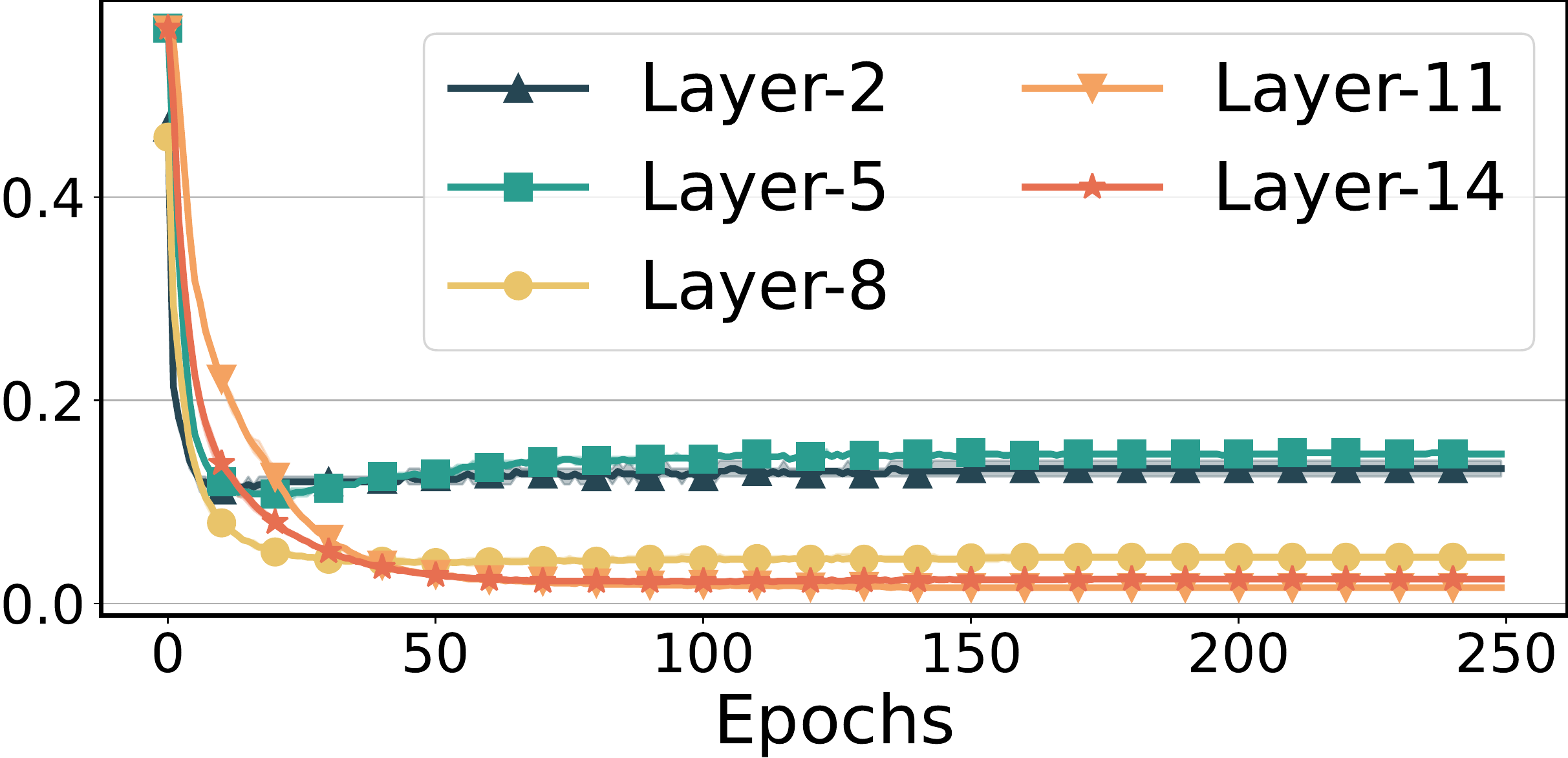}
    \vspace{-3mm}
    \caption{The stable ranks for various layers in ResNet-18 ({\bf the top row}) and VGG-19 ({\bf the bottom row}) trained on CIFAR-10 using stable rank.}
    \label{fig:rank-est-resnet18-vgg19-cifar10}
    \vspace{-2mm}
\end{figure*}

\begin{figure*}[ht]
    \vspace{-1mm}
    \centering
    \includegraphics[width=0.315\textwidth]{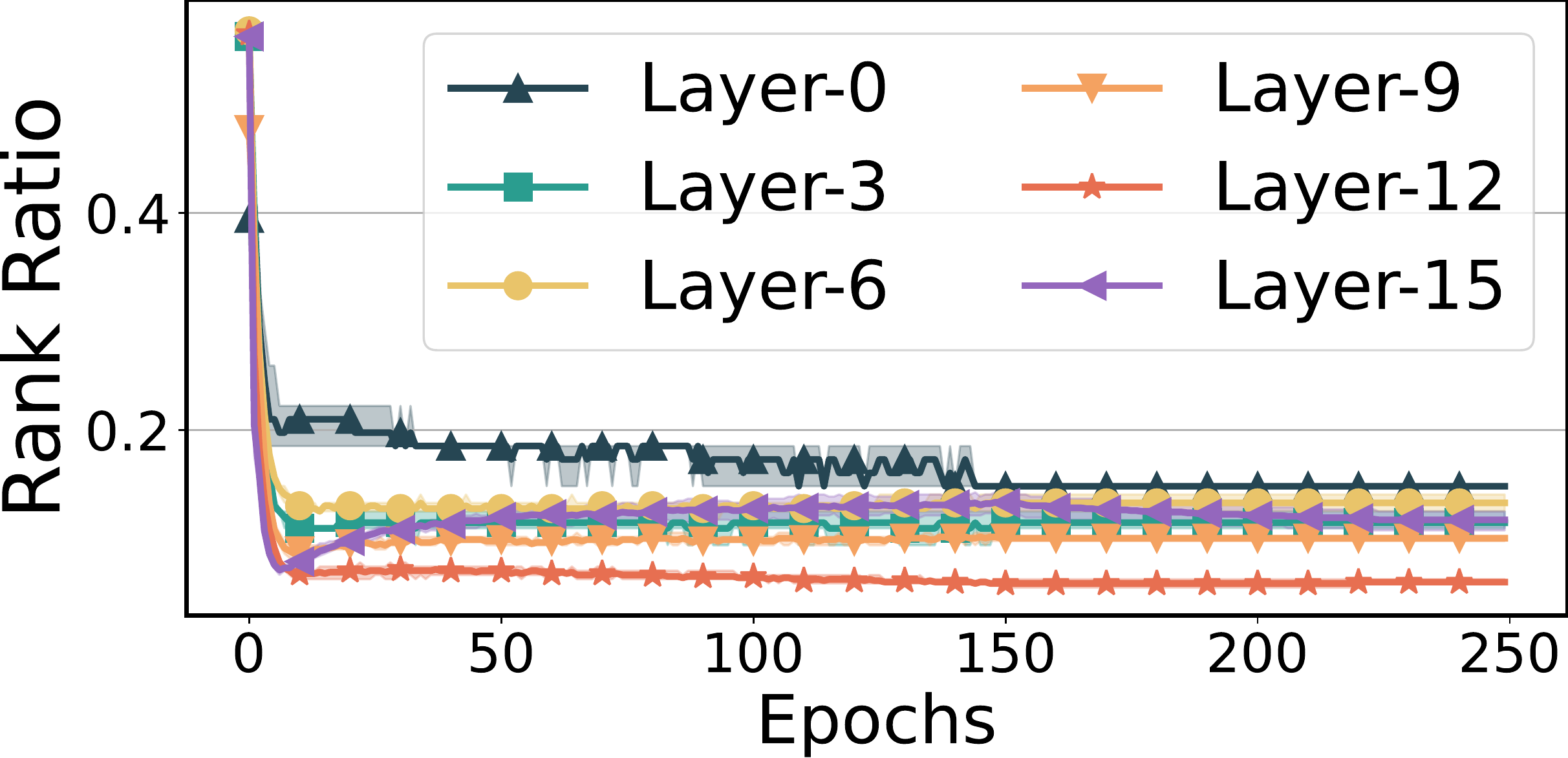}
    \includegraphics[width=0.315\textwidth]{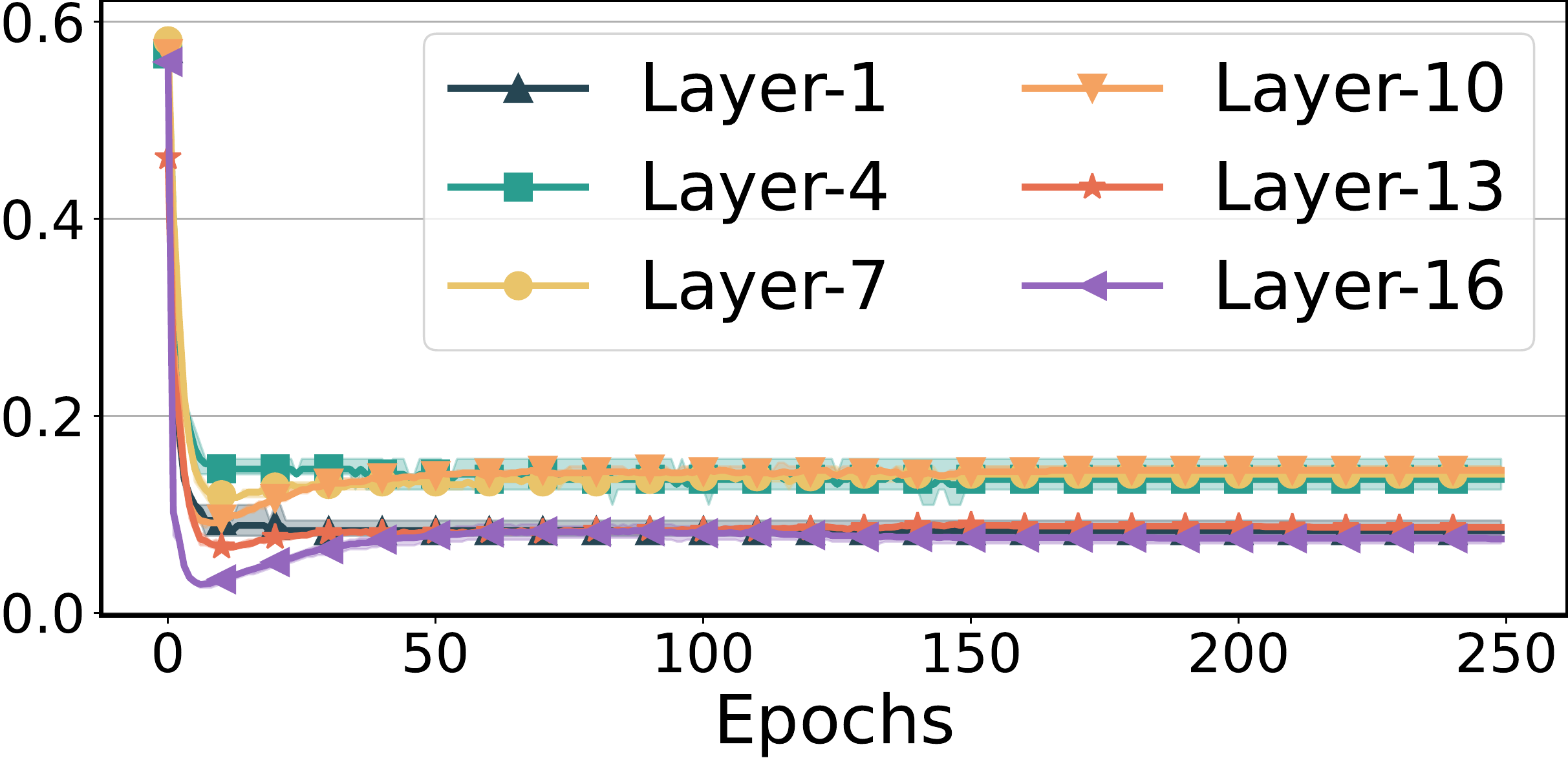}
    \includegraphics[width=0.315\textwidth]{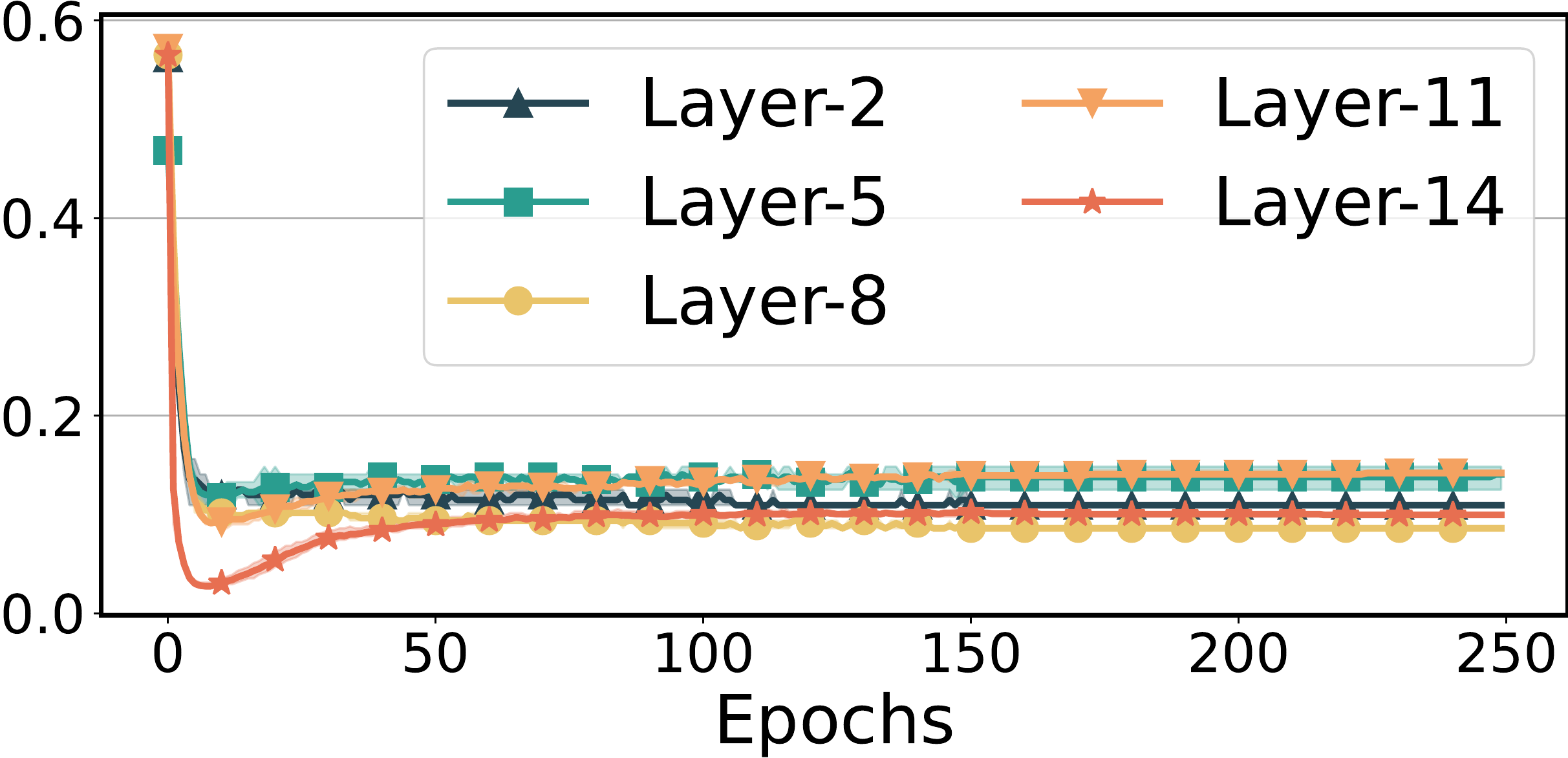}\\
    \includegraphics[width=0.315\textwidth]{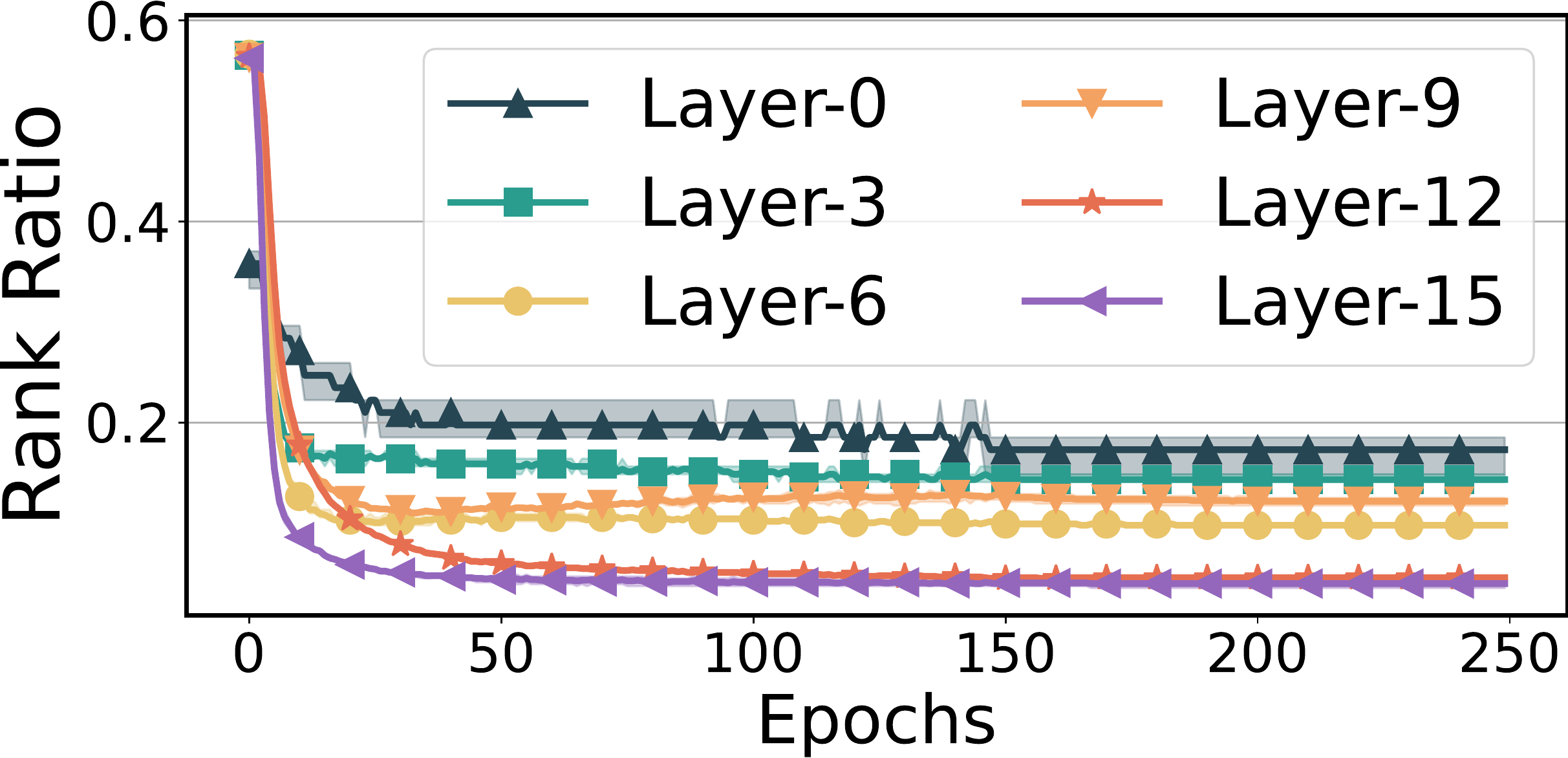}
    \includegraphics[width=0.315\textwidth]{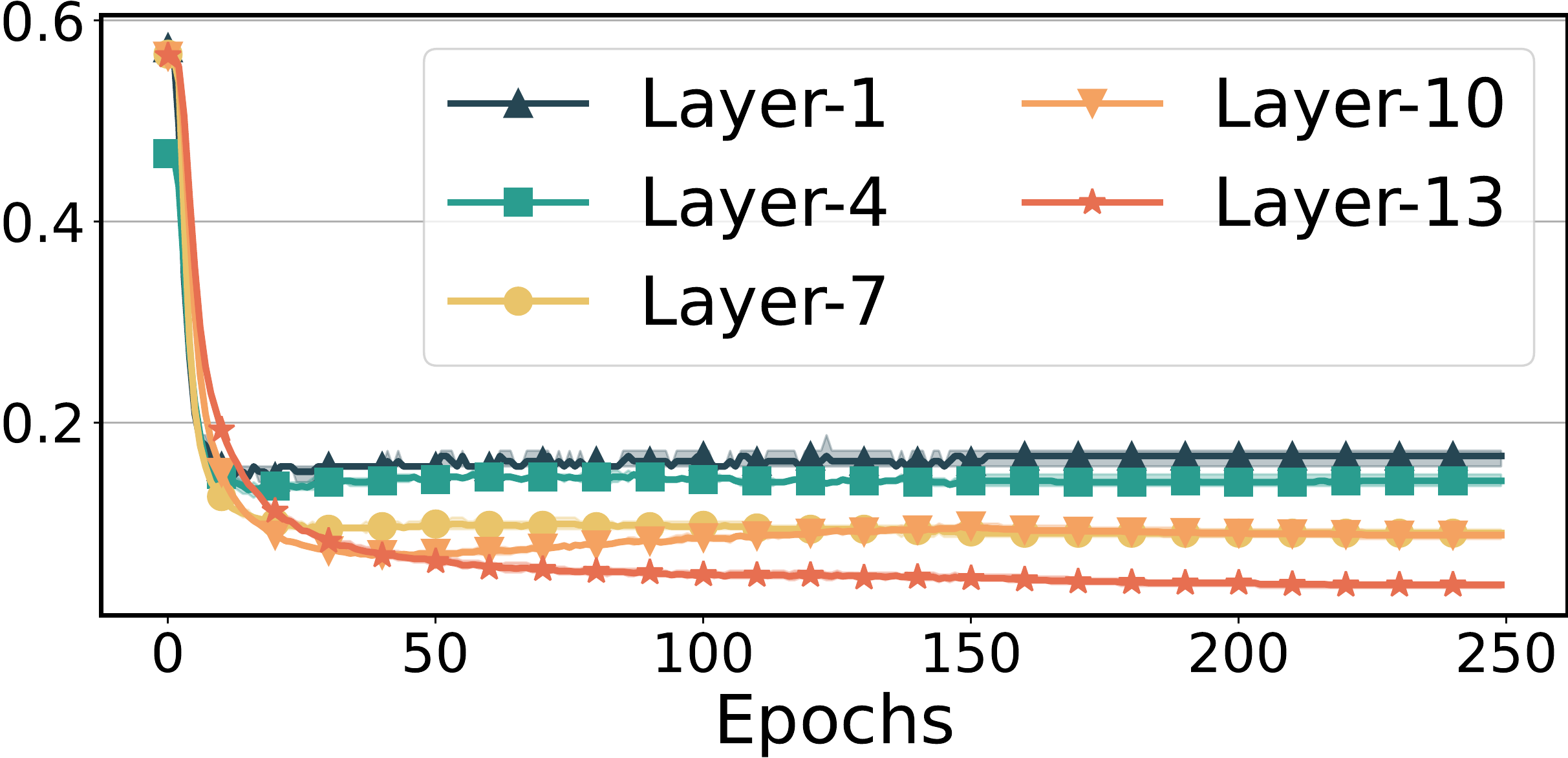}
    \includegraphics[width=0.315\textwidth]{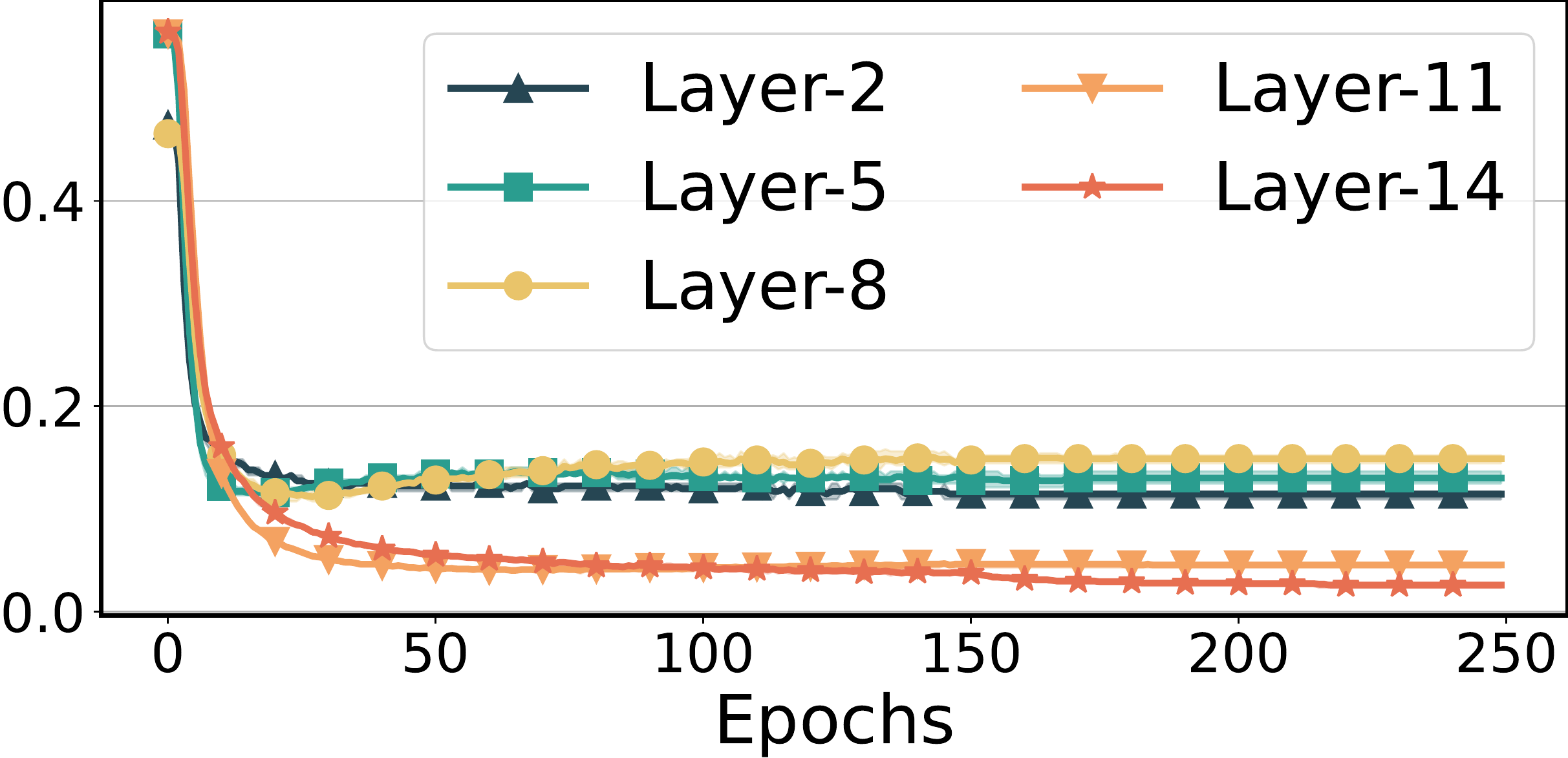}
    \vspace{-4mm}
    \caption{The stable ranks for various layers in ResNet-18 ({\bf the top row}) and VGG-19 ({\bf the bottom row}) trained on CIFAR-100 using stable rank with batch size $1,024$.}
    \vspace{-1.5mm}
    \label{fig:rank-est-resnet18-vgg19-cifar100}
\end{figure*}

\begin{figure*}[ht]
    \vspace{-1mm}
    \centering
    \includegraphics[width=0.315\textwidth]{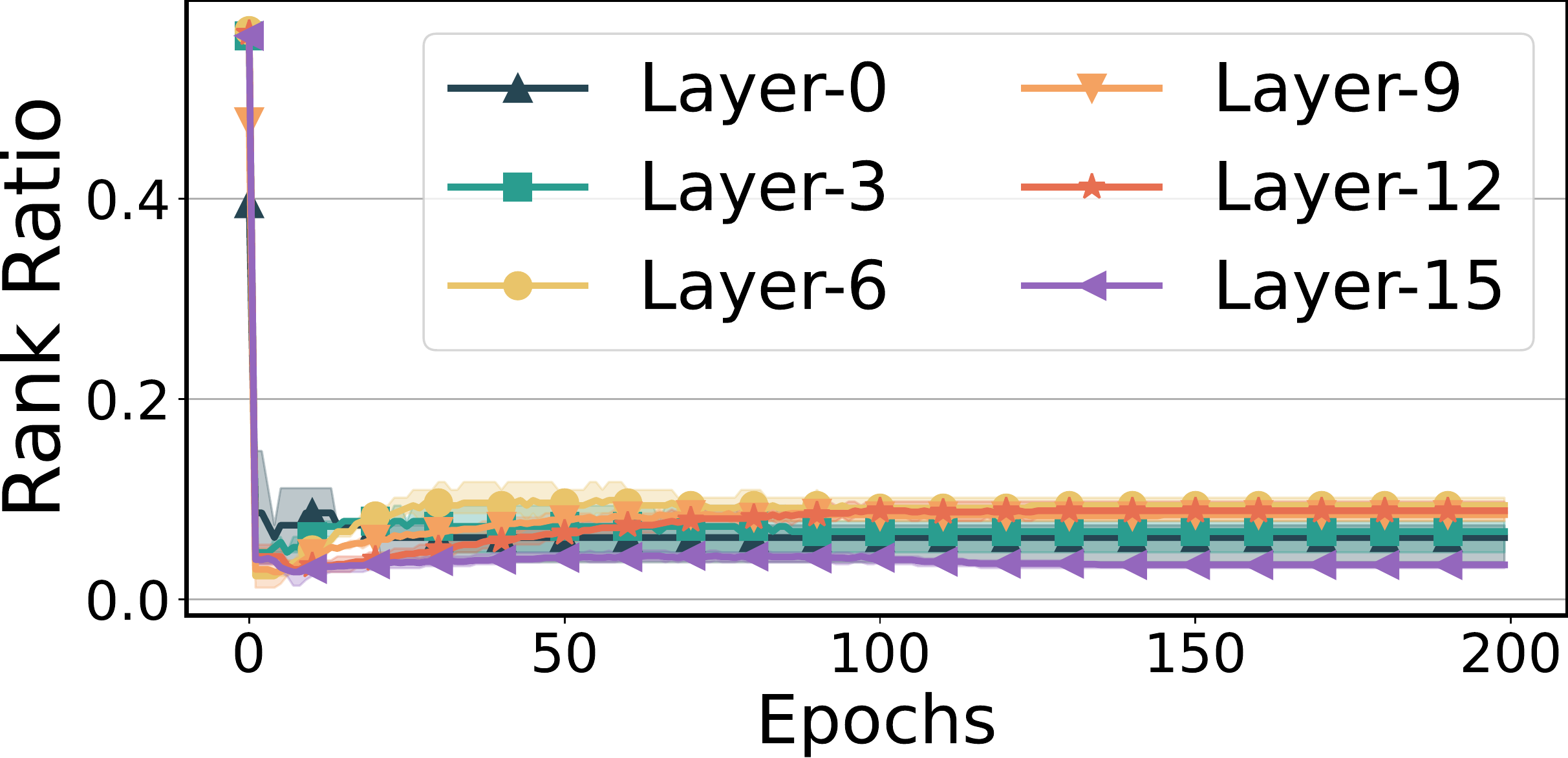}
    \includegraphics[width=0.315\textwidth]{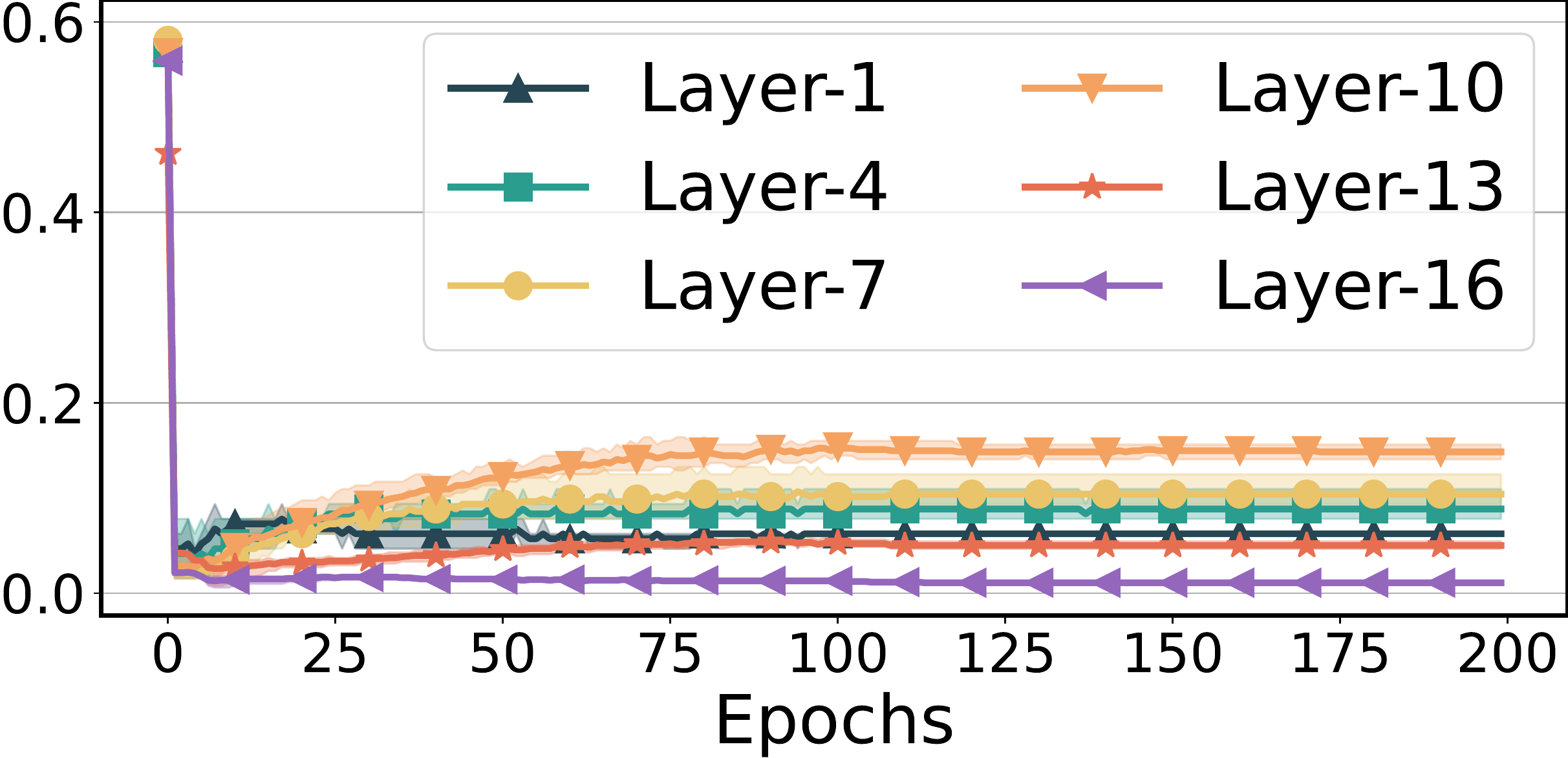}
    \includegraphics[width=0.315\textwidth]{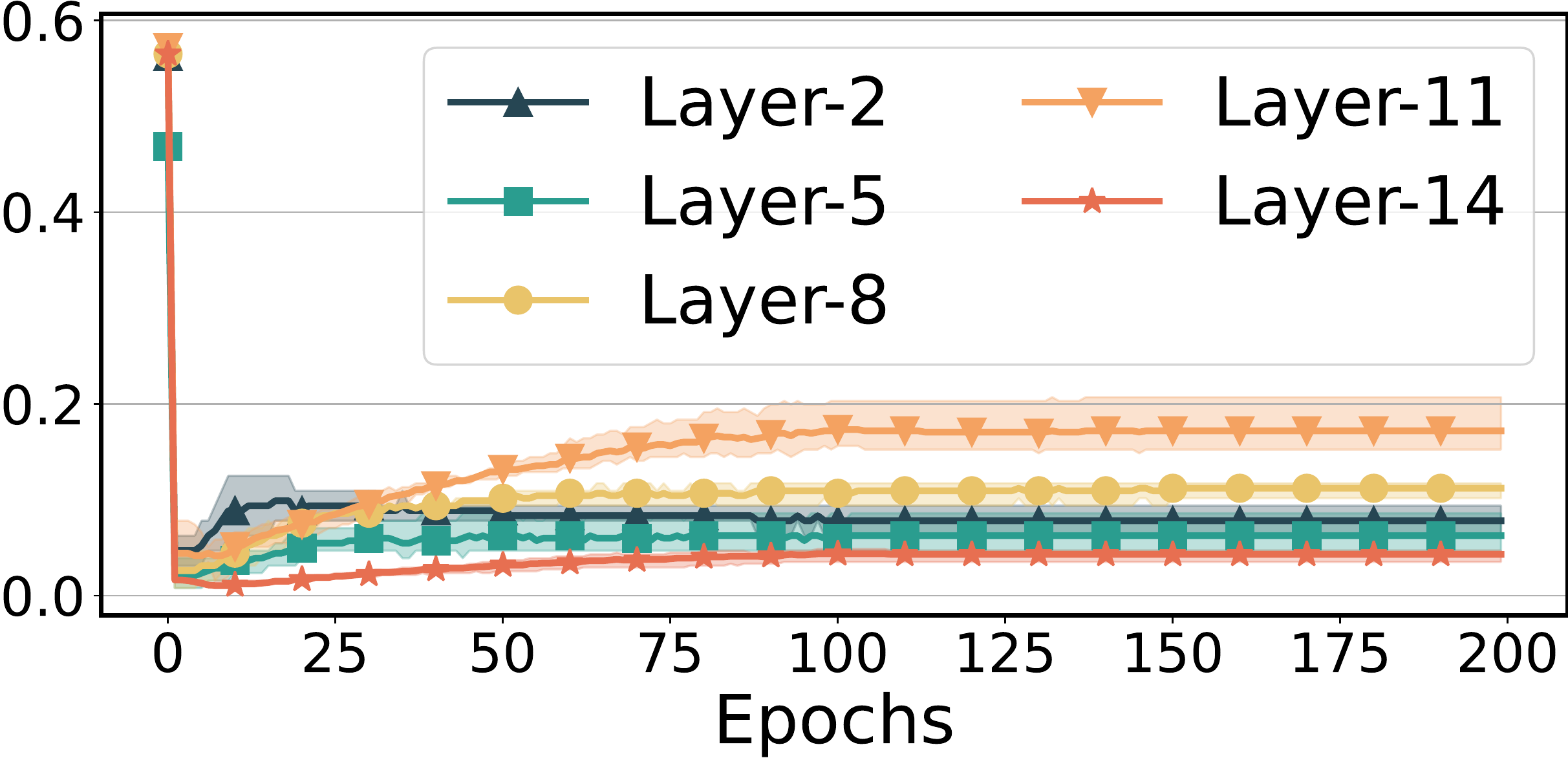}\\
    \includegraphics[width=0.315\textwidth]{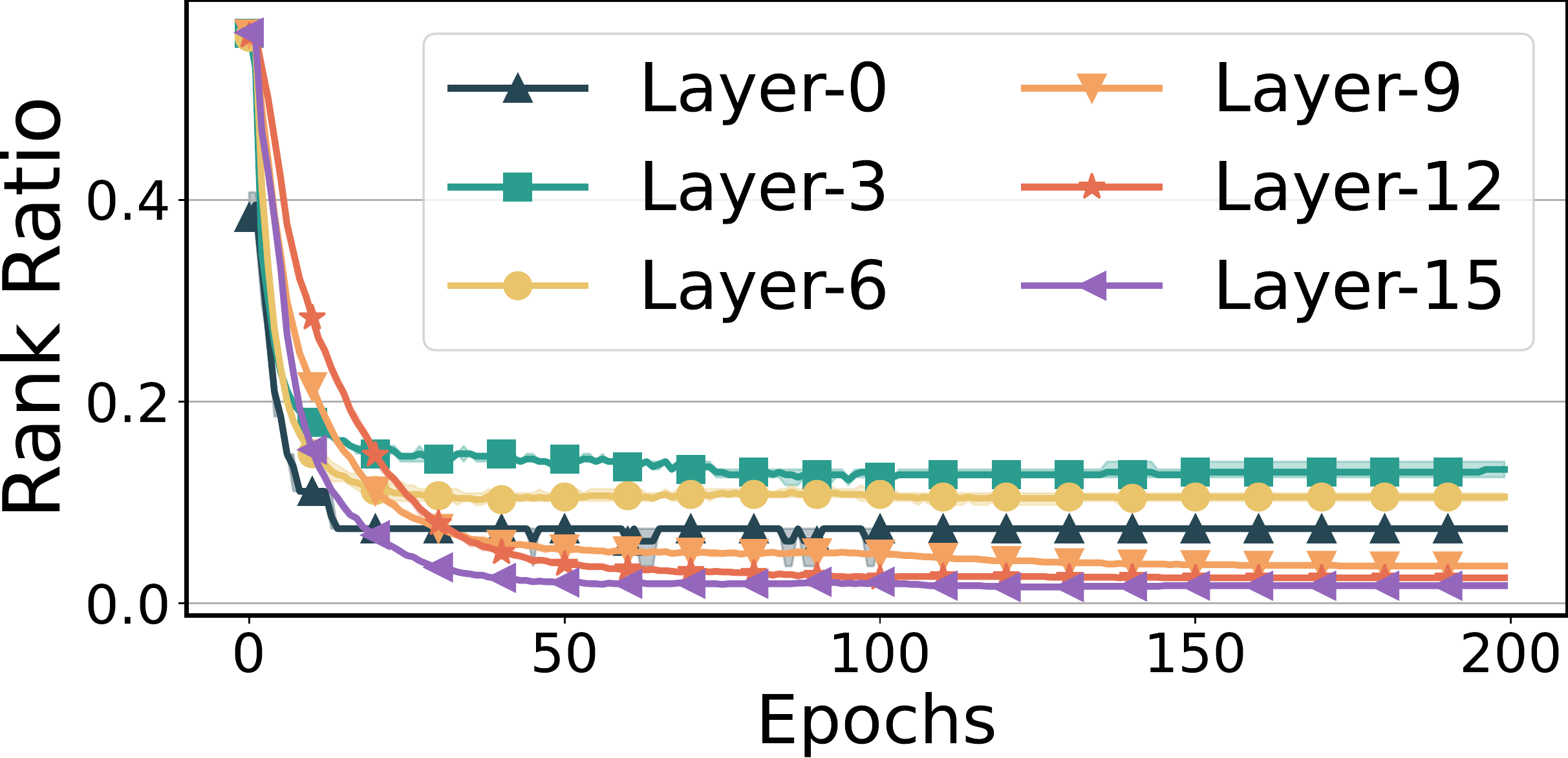}
    \includegraphics[width=0.315\textwidth]{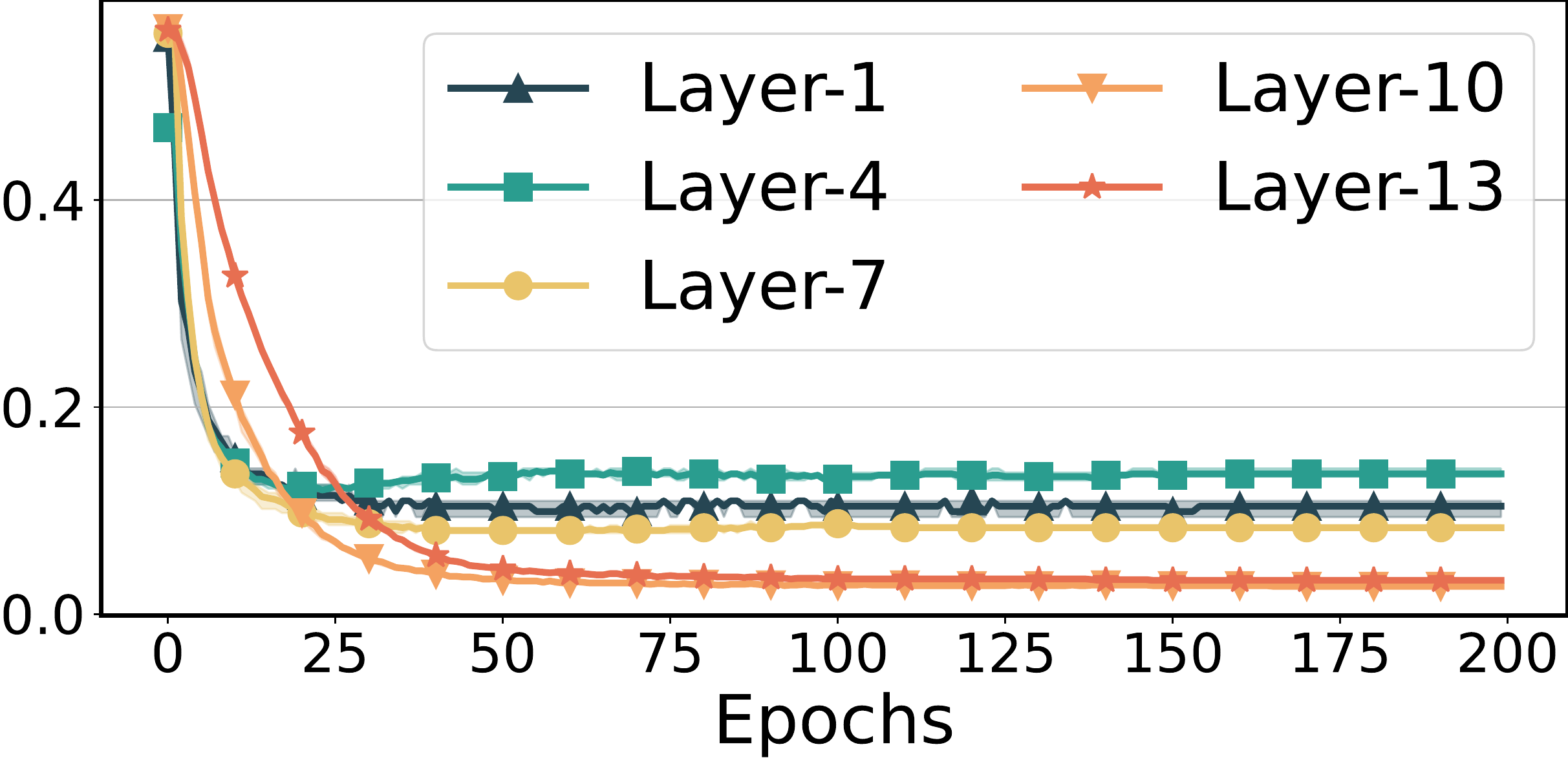}
    \includegraphics[width=0.315\textwidth]{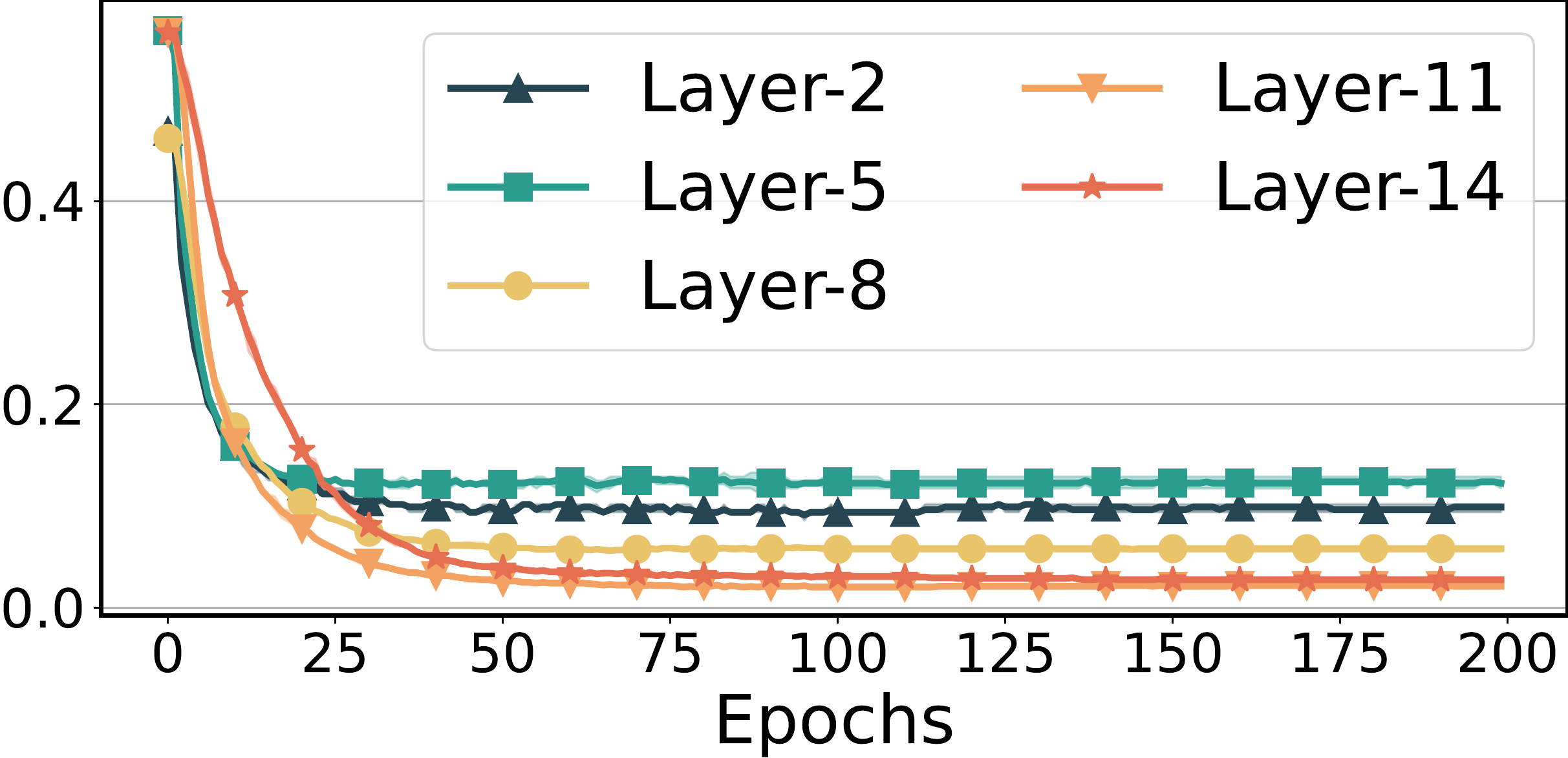}
    \vspace{-4mm}
    \caption{The stable ranks for various layers in ResNet-18 ({\bf the top row}) and VGG-19 ({\bf the bottom row}) trained on SVHN using stable rank with batch size $1,024$.}
    \vspace{-2mm}
    \label{fig:rank-est-resnet18-vgg19-svhn}
\end{figure*}

\begin{figure*}[ht]
    \centering
    \includegraphics[width=0.315\textwidth]{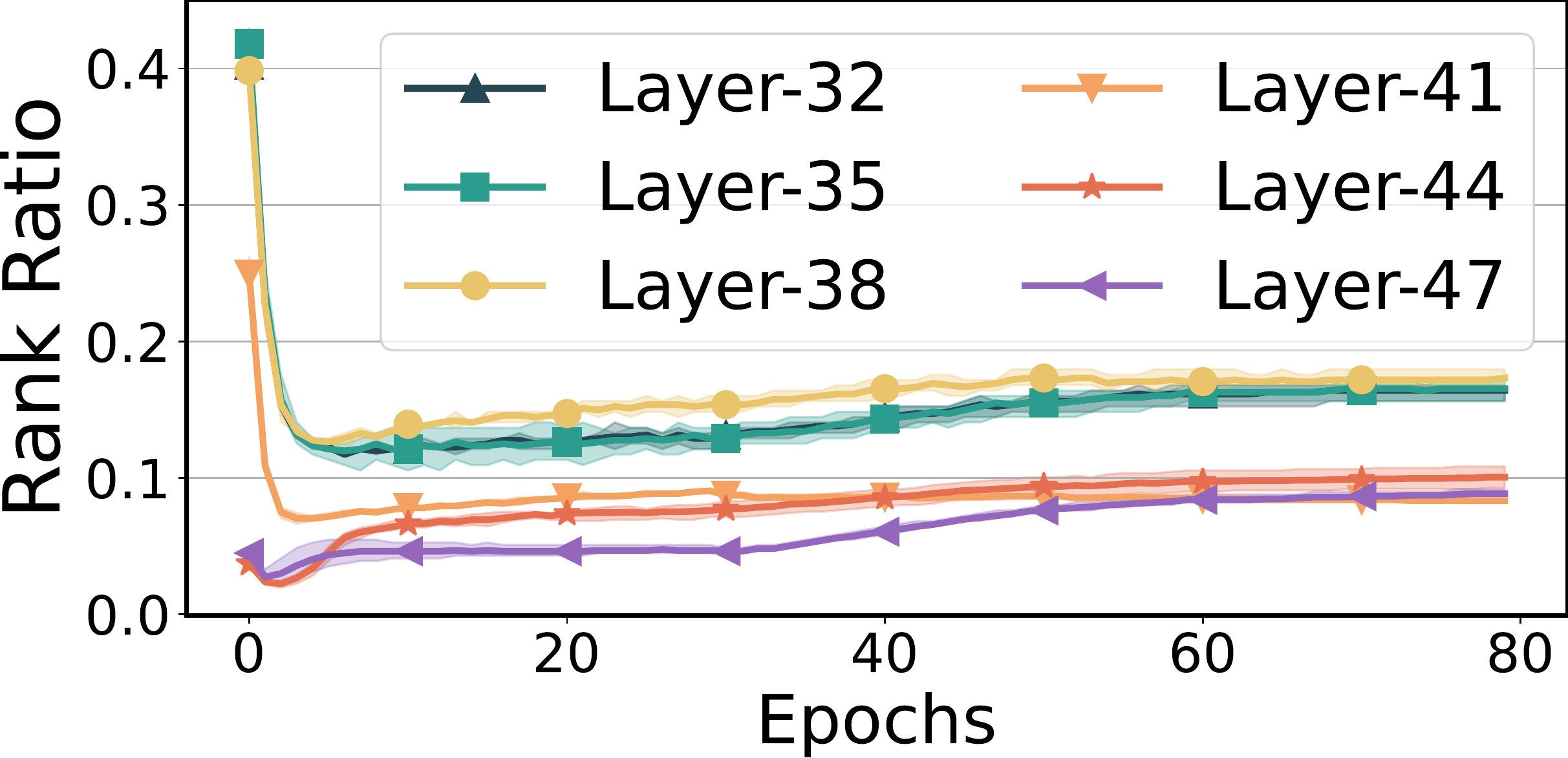}
    \includegraphics[width=0.315\textwidth]{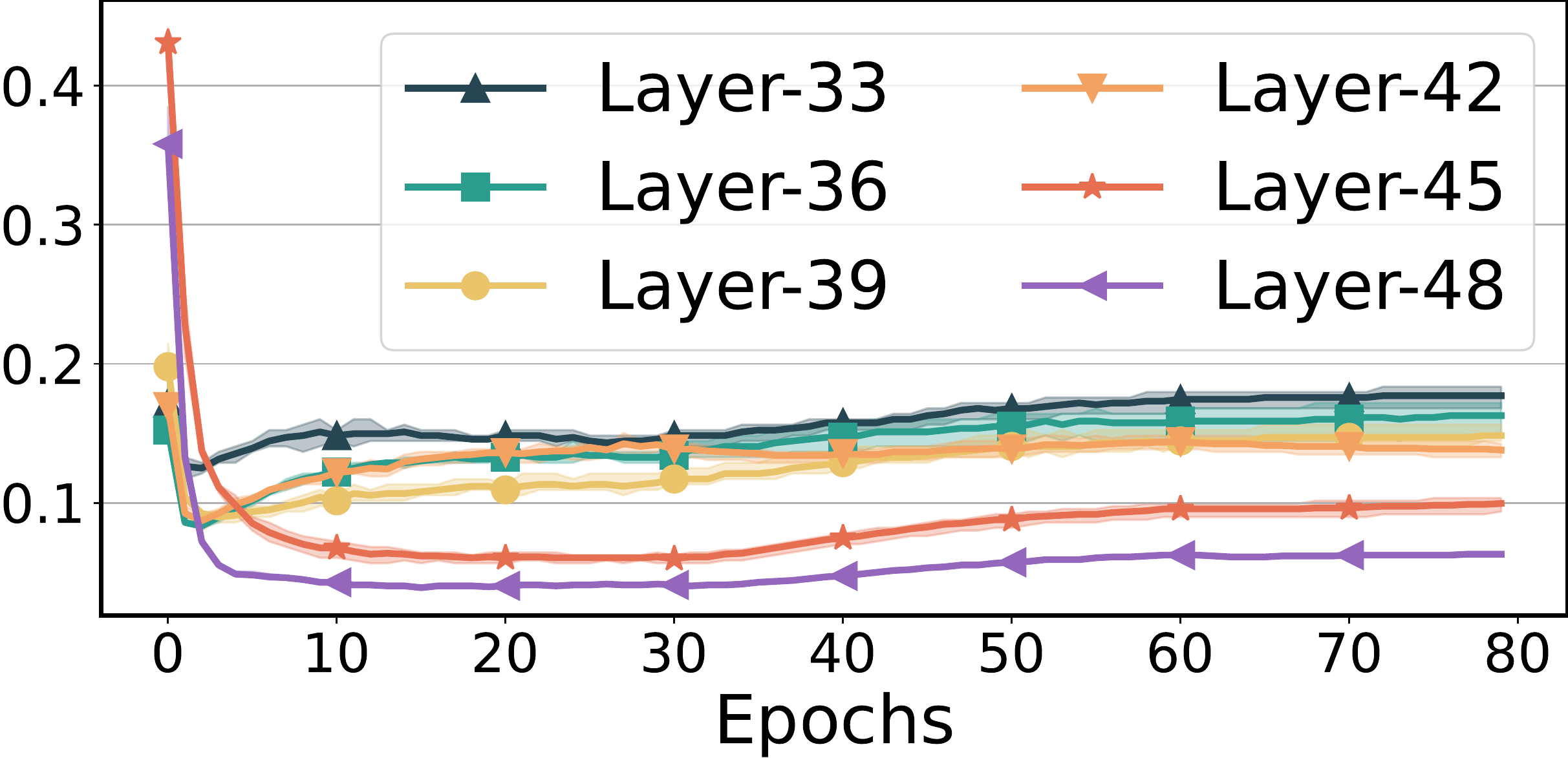}
    \includegraphics[width=0.315\textwidth]{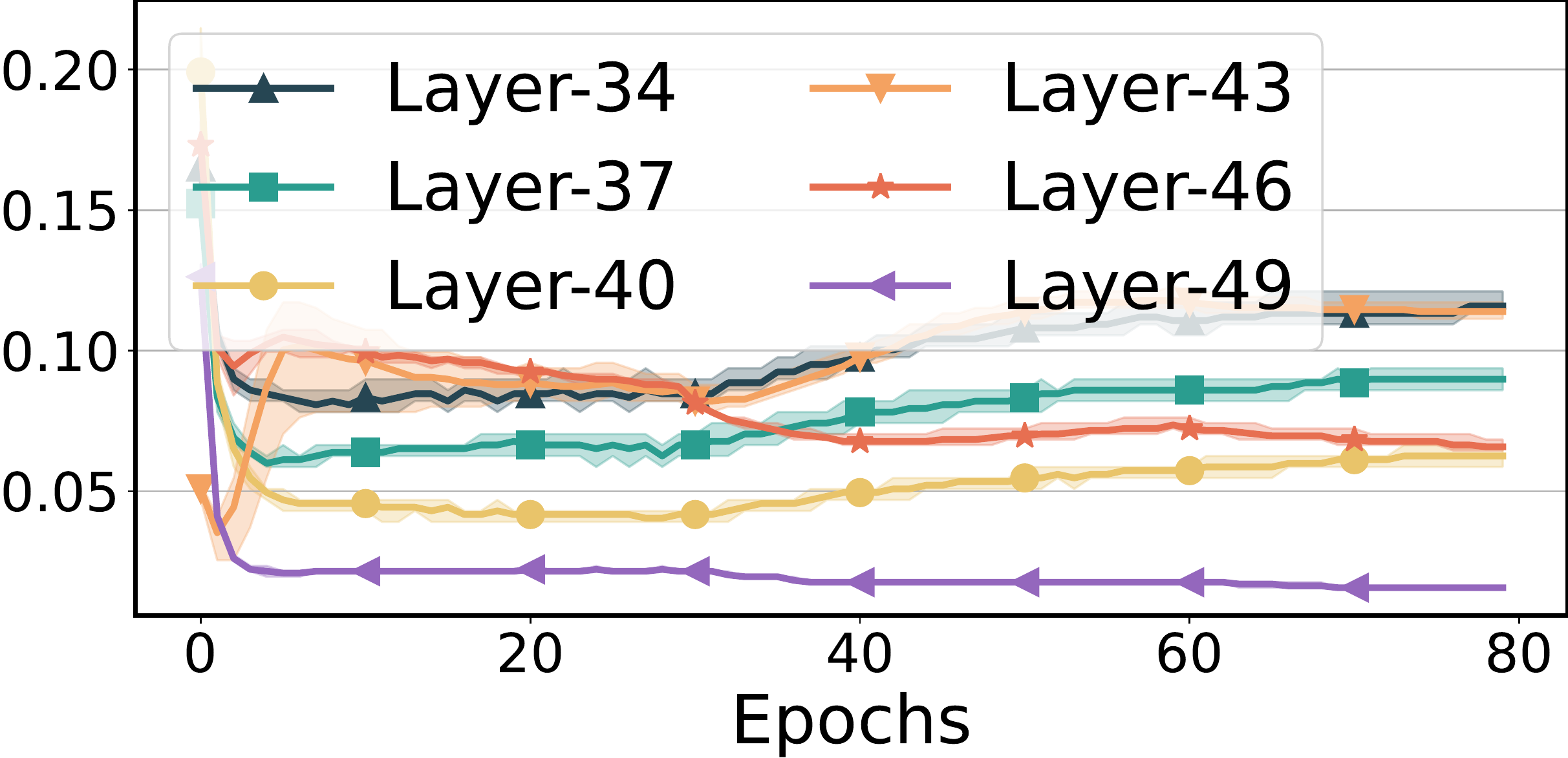}
    \vspace{-4mm}
    \caption{The stable ranks for various layers in ResNet-50 trained on ImageNet using stable rank with batch size $256$.}
    \label{fig:rank-est-resnet50-imagenet}
\vspace{-2mm}
\end{figure*}

\begin{figure*}[ht]
    \vspace{-2mm}
    \centering
    \subfigure[ResNet-18]{\includegraphics[width=0.6\textwidth]{figures/est_rank_resnet18_cifar10.pdf}}\\
    \vspace{-2mm}
    \subfigure[VGG-19]{\includegraphics[width=0.6\textwidth]{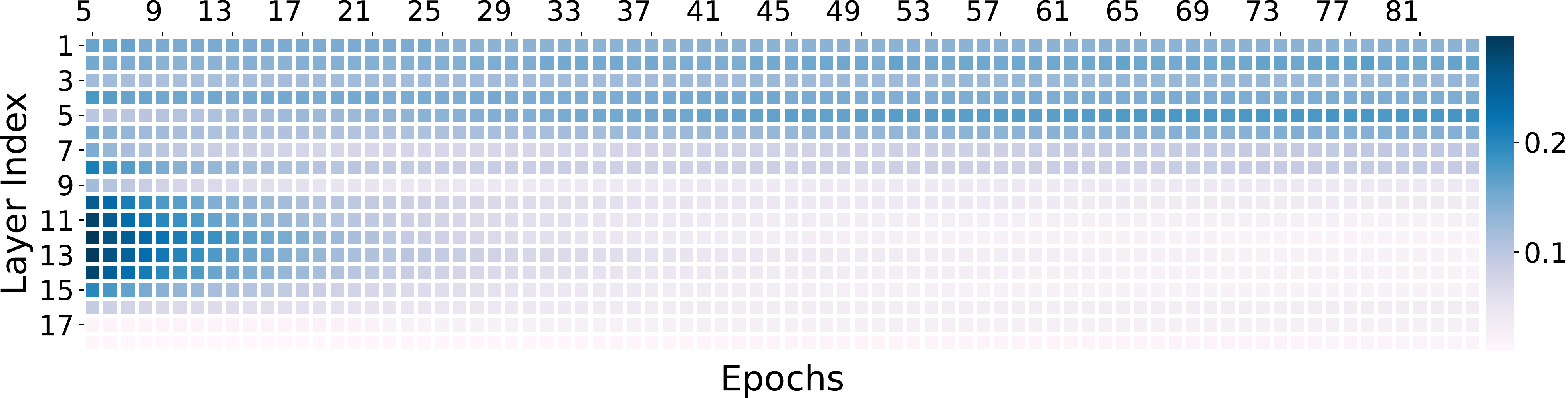}\label{fig:rank-vary-trend1}}
    \vspace{-4mm}
    \caption{The stable ranks for various layers in ResNet-18 and VGG-19 trained on CIFAR-10 using stable rank.}
    \label{fig:rank-ratio-resnet18-vgg19-cifar10}
\vspace{-2mm}
\end{figure*}

\begin{table*}[ht]
\vspace{-3mm}
\caption{ Comparison of \cuttlefish{} and other baseline methods: \pufferfish{}, EB Train ($30\%$, $50\%$) and GraSP ($30\%$, $60\%$) over the task of ResNet-50 on ImageNet.}
\label{table:comparison-eb-train-grasp}
	\begin{center}
		 \scriptsize{
		\begin{tabular}{cccc}
		\rowcolor{Gray} \toprule \textbf{Model} 
		 & \# Params. &  Val. Acc. & Val. Acc.
		\bigstrut\\
        \rowcolor{Gray} ResNet-50 
		 & ($M$) &  Top-1(\%) & Top-5(\%)
		\bigstrut\\
		\midrule
		\rowcolor{LightCyan} Full-rank & $25.6$ & $75.99$ & $92.98$ \bigstrut\\
		\pufferfish{} & $15.2$ & $75.62$ & $92.55$ \bigstrut\\
		\rowcolor{LightCyan} EB Train ($30\%$) & $16.5$ & $73.86$ & $91.52$ \bigstrut\\
		EB Train ($50\%$) & $15.1$ & $73.35$ & $91.36$ \bigstrut\\
		\rowcolor{LightCyan}GraSP ($30\%$) & $17.9$ & $74.64$ & $92.08$ \bigstrut\\
		GraSP ($60\%$) & $10.2$ & $74.02$ & $91.86$  \bigstrut\\
		\rowcolor{LightCyan} \cuttlefish{} & $14.7$ & $75.80$ & $92.70$ \bigstrut\\
		\bottomrule
		\end{tabular}}%
	\end{center}
\vspace{-3mm}
\end{table*}

\begin{figure*}[ht]
    \vspace{-2mm}
    \centering
    \subfigure[ResNet-18]{\includegraphics[width=0.6\textwidth]{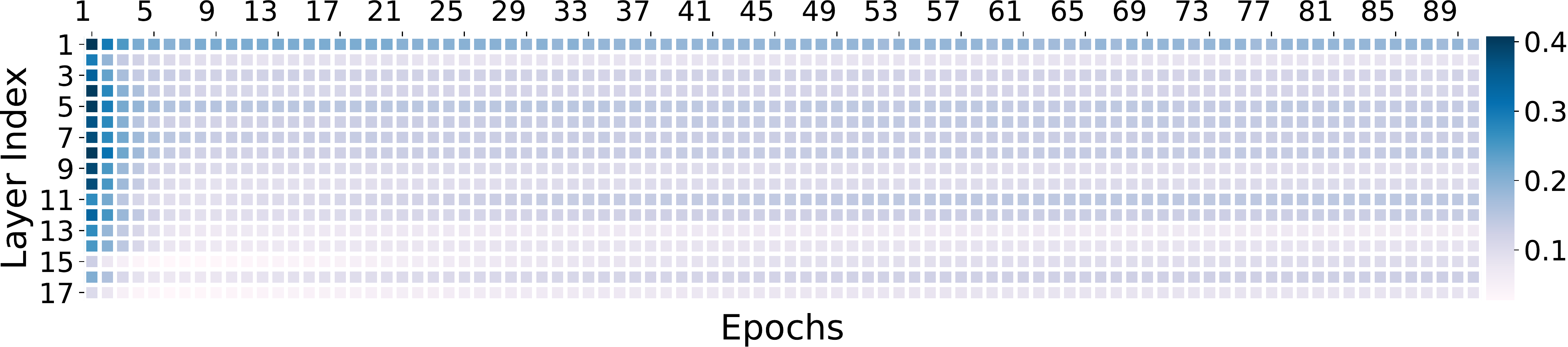}}\\
    \vspace{-2mm}
    \subfigure[VGG-19]{\includegraphics[width=0.6\textwidth]{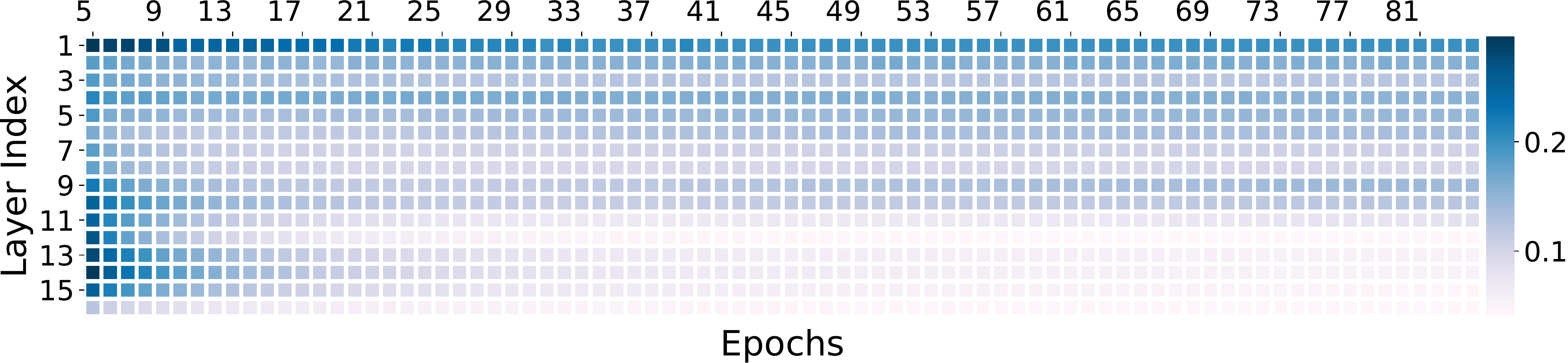}\label{fig:rank-vary-trend2}}
    \vspace{-4mm}
    \caption{The stable ranks for various layers in ResNet-18 and VGG-19 trained on CIFAR-100 using stable rank.}
    \label{fig:rank-ratio-resnet18-vgg19-cifar100}
\vspace{-2mm}
\end{figure*}

\begin{figure*}[ht]
    \vspace{-4mm}
    \centering
    \subfigure[ResNet-18]{\includegraphics[width=0.6\textwidth]{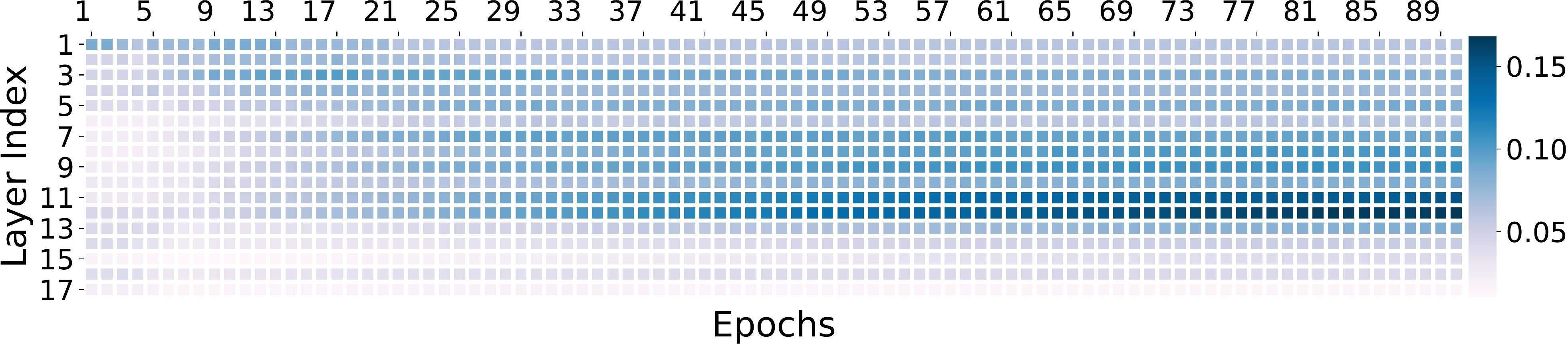}}\\
    \vspace{-2mm}
    \subfigure[VGG-19]{\includegraphics[width=0.6\textwidth]{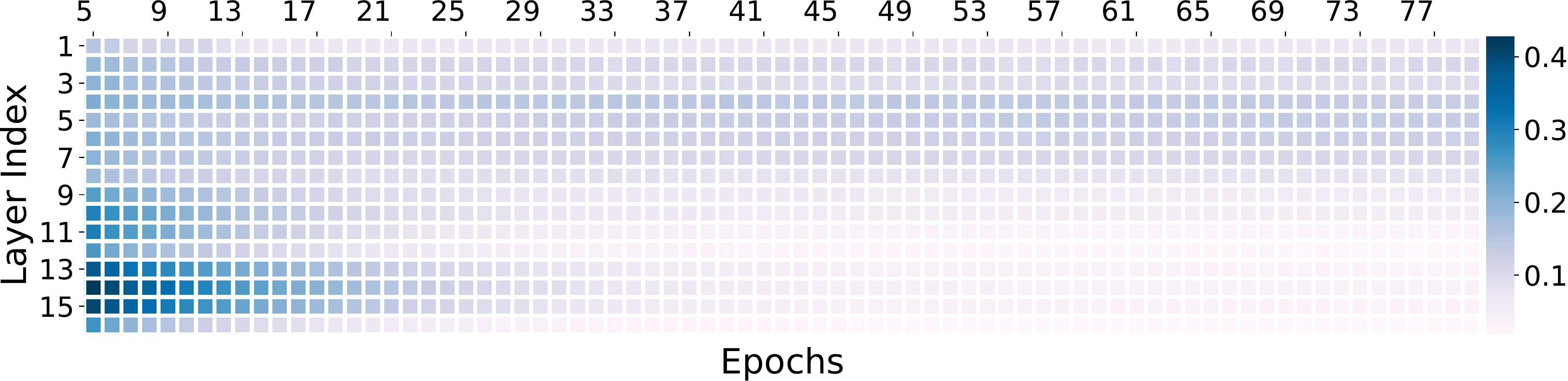}\label{fig:rank-vary-trend3}}
    \vspace{-4mm}
    \caption{The stable ranks for various layers in ResNet-18 and VGG-19 trained on SVHN using stable rank.}
    \label{fig:rank-ratio-resnet18-vgg19-svhn}
    \vspace{-4mm}
\end{figure*}

\begin{figure*}[ht]
    \vspace{-4mm}
    \centering
    \includegraphics[width=0.6\textwidth]{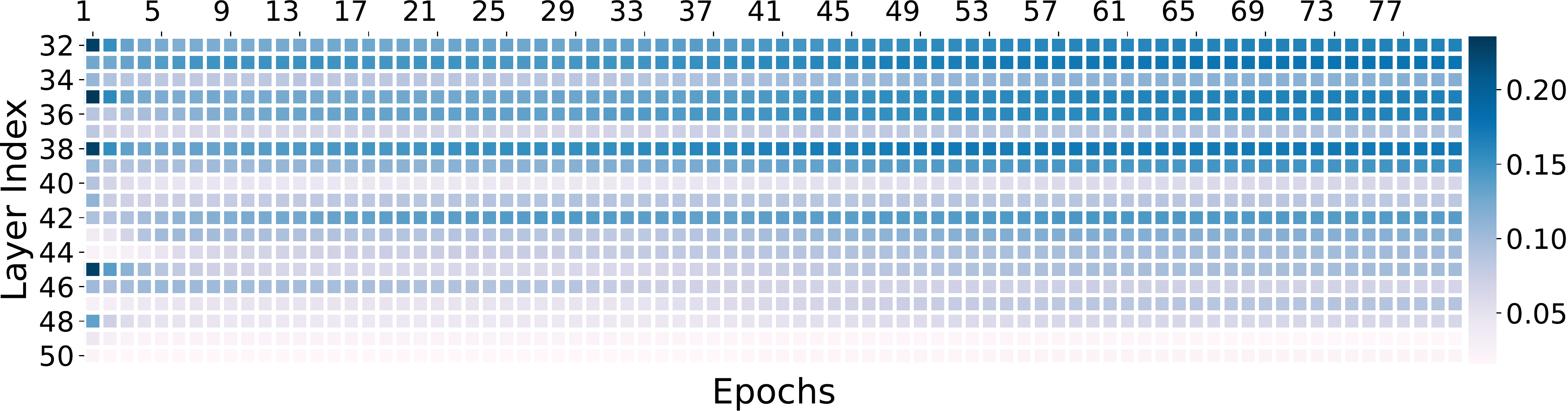}
    \vspace{-4mm}
    \caption{The stable ranks for various layers in ResNet-50 trained on ImageNet using stable rank.}
    \label{fig:rank-ratio-resnet50-imagenet}
    \vspace{-4mm}
\end{figure*}

\begin{table*}[ht]
	\caption{The results, averaged over three independent trials with different random seeds, showcase the performance of \cuttlefish{} and other baselines on ResNet-18 and VGG-19 trained on the SVHN dataset using a batch size of 1,024. Runtime benchmarks are conducted on a single EC2 p3.2xlarge instance.}
	\label{table:svhn-results}
	\begin{center}
      \scriptsize{
		\begin{tabular}{cccc}
		\toprule
		\rowcolor{Gray} \textbf{Model:} & Params. & Val. Acc. & Time
		\\
		\rowcolor{Gray} ResNet-18 & ($M$) & ($\%$) & (hrs.)
		\bigstrut\\
		\midrule
		\rowcolor{LightCyan} Full-rank & $11.2$ & $96.27_{\pm 0.08}$ & $0.81$\bigstrut\\
		\pufferfish{} & $3.3$ & $96.54_{\pm 0.06}$ & $0.71$\bigstrut\\
		\rowcolor{LightCyan} SI\&FD &$0.94$&$96.45_{\pm 0.04}$& $0.38$\bigstrut\\
		IMP & $0.96$& $96.43_{\pm 0.01}$& $9.40$\bigstrut\\
		\rowcolor{LightCyan} \cuttlefish{} & $0.96$ & $96.47_{\pm 0.02}$& $0.65$ \bigstrut\\
		\cuttlefish{}+FD & $0.94$ & $96.34_{\pm 0.08}$& $0.65$ \bigstrut\\
		\rowcolor{Gray} \midrule \textbf{Model:} 
		& Params. & Val. Acc. & Time
		\\
		\rowcolor{Gray} VGG-19 
		& ($M$) & ($\%$) & (hrs.) 
		\bigstrut\\
		\midrule
		\rowcolor{LightCyan} Full-rank & $20.0$ & $96.31_{\pm 0.05}$& $0.49$\bigstrut\\
		\pufferfish{} & $8.1$ & $96.08_{\pm 0.11}$& $0.45$\bigstrut\\
		\rowcolor{LightCyan} SI\&FD &$1.2$&$96.04_{\pm 0.16}$& $0.30$\bigstrut\\
		 FC Compress. &$1.1$& $96.42_{\pm 0.07}$& $6.68$\bigstrut\\
		 \cuttlefish{} & $1.2$ & $96.33_{\pm 0.04}$ & $0.39$\bigstrut\\
		 \rowcolor{LightCyan} \cuttlefish{}+FD & $1.2$ & $96.33_{\pm 0.02}$ & $0.39$\bigstrut\\
		\bottomrule
		\end{tabular}}%
	\end{center}
\end{table*}

\paragraph{Comparison of \cuttlefish{} to EB Train and GraSP.} Furthermore, we compare \cuttlefish{} with two other state-of-the-art approaches, namely EB Train and GraSP (as shown in Table~\ref{table:comparison-eb-train-grasp}). Notably, \cuttlefish{} outperforms both EB Train and GraSP in terms of accuracy while achieving smaller model sizes.

\paragraph{ResNet-18 and VGG-19 Experiments on SVHN.} The results of ResNet-18 and VGG-19 experiments on the SVHN dataset are presented in Table~\ref{table:svhn-results}.


\end{document}